\documentclass[letterpaper]{article} 
\usepackage[preprint]{aaai2027}
\usepackage[hyphens]{url}  
\usepackage{graphicx} 
\usepackage{natbib}  
\usepackage{caption} 
\usepackage{booktabs}

\usepackage[utf8]{inputenc} 
\usepackage[T1]{fontenc}    
\usepackage{amsfonts}       
\usepackage{microtype}      
\usepackage{xcolor}         
\usepackage{colortbl}
\usepackage{amsmath}
\usepackage{amssymb}
\usepackage{mathtools}
\usepackage{makecell}
\usepackage{multirow}
\usepackage{longtable}
\usepackage{subcaption}

\title{Can LLMs Faithfully Solve Ontology Engineering Tasks? A Mechanistic Interpretability Study}
\author {
    Bohui Zhang\textsuperscript{\rm 1},
    Albert Mero\~{n}o-Pe\~{n}uela\textsuperscript{\rm 1},
    Elena Simperl\textsuperscript{\rm 1,\rm 2}\corresponding
}
\affiliations {
    \textsuperscript{\rm 1}King's College London, United Kingdom\\
    \textsuperscript{\rm 2}Technical University of Munich, Germany\\
    bohui.zhang@kcl.ac.uk, albert.merono@kcl.ac.uk, elena.simperl@kcl.ac.uk
}

\begin{document}

\maketitle

\begin{abstract}
Ontological knowledge underpins large language model (LLM) performance on knowledge-intensive tasks. However, existing evaluations of LLMs' ontological understanding largely rely on behavioral correctness, leaving the underlying mechanisms unexplained. Existing mechanistic interpretability work has largely focused on instance-level factual knowledge, leaving ontological knowledge under-explored. To bridge this gap, we introduce MIOE, a causal-abstraction benchmark that evaluates interchange interventions for two ontology engineering tasks: entity typing and subsumption prediction. MIOE tests whether learned internal intervention targets can change an ontology-level target answer while preserving class-specific auxiliary information. Across Qwen3-8B and Qwen3-4B, we evaluate SAE-, DBM-, and DAS-based methods, including multi-task and multi-layer variants. Distributed subspace methods give the strongest results and generalize across prompt settings, but the induced changes remain only partially separated from auxiliary class-specific information. Multi-task objectives and contiguous layer-window interventions improve this balance, while class-level analyses reveal substantial variation across ontology classes. These results suggest that ontological predictions are mechanistically accessible in LLMs, although current methods do not yet isolate a cleanly disentangled causal mechanism for ontological knowledge.
\end{abstract}

\begin{links}
    \link{Code}{https://github.com/King-s-Knowledge-Graph-Lab/MIOE}
\end{links}

\section{Introduction}

Formal ontologies provide schema-level knowledge for many knowledge-intensive tasks in the AI era. They define classes, relations, constraints, and hierarchies that support structured knowledge representation, semantic reasoning, and data integration~\cite{pan-et-al-2023-tgdk}. As LLMs become increasingly capable of generating, validating, and revising knowledge, they are also being explored as assistants for ontology engineering tasks, including ontology construction, completion, alignment, and evaluation~\cite{giglou-et-al-2023-llms4ol,lo-et-al-2024-end-to-end,zhang-et-al-2025-ontochat,zhang-et-al-2025-schema}.

Despite this promise, recent empirical evidence suggests that LLMs still face substantial limitations in ontology engineering. While they can perform well on basic ontology-related tasks~\cite{he-et-al-2023-ontolama,giglou-et-al-2023-llms4ol}, their performance remains suboptimal and often unpredictable on more advanced tasks, such as ontology evaluation~\cite{tsaneva-et-al-2025}. This leaves open a critical question: whether correct ontology-oriented outputs are supported by internal computations that track the relevant ontological relation, and how these computations interact with broader subject information. This raises a concern about faithfulness, defined as the degree to which a model's output reflects the internal process that produced it~\cite{wiegreffe-pinter-2019,jacovi-goldberg-2020-faithfulness}. However, most existing evaluations of LLMs for ontology-related tasks remain behavioral, asking whether the model outputs the correct label, axiom, or judgment under a given prompt. Such evaluations are necessary, but they do not reveal the mechanism by which the answer is produced.

Mechanistic interpretability offers a way to study this problem by connecting high-level variables to causal mechanisms inside the model. Interchange interventions provide a causal abstraction test: they replace internal representations across inputs and examine whether the model's behavior changes according to the corresponding high-level variable~\cite{geiger-et-al-2021-causal,geiger-et-al-2022-iit}. Prior work has applied mechanistic interpretability to factual knowledge in LLMs, including the localization, editing, causal role of factual associations, and feature disentanglement~\cite{geva-et-al-2022-transformer,dai-et-al-2022-knowledge,meng-et-al-2022-rome,huang-et-al-2024-ravel}. However, ontological knowledge plays a different role from ordinary factual or attribute-level knowledge. While factual attributes capture local properties of a subject, ontology-level relations determine its type, taxonomic position, and the constraints or properties that are applicable to it. This makes ontological knowledge a distinctive target for mechanistic interpretability.

In this paper, we investigate mechanistic interpretability for ontological knowledge in LLMs. We move beyond output-level correctness and ask whether ontology-level variables can be connected to the model's internal computation through testable causal interventions. We introduce Mechanistic Interpretability for Ontology Engineering (MIOE), an ontology-centric benchmark for evaluating interchange interventions over two fundamental ontology engineering tasks. MIOE pairs target ontological relations with class-specific auxiliary attributes, prompt templates, and contextual inputs, enabling us to measure both intervention effects on the target relation and preservation of non-target subject information.

We ask the following research questions:
\begin{enumerate}
	\item Can learned internal intervention targets reproduce the counterfactual behavior specified by ontology-level variables in entity typing and subsumption prediction?
	\item Can learned internal interventions change the ontology-level target variable while leaving class-specific auxiliary variables invariant?
	\item How do intervention method, representation site, prompt type, and ontological class affect target-relation intervention effects and auxiliary preservation?
\end{enumerate}

Our contributions are summarized as follows:
\begin{enumerate}
    \item We formulate mechanistic interpretability for ontology engineering as a causal abstraction problem over ontology-level variables, shifting the evaluation of ontological knowledge from behavioral correctness to testable internal interventions.
    \item We introduce MIOE, a mechanistic interpretability benchmark for ontology engineering covering entity typing and subsumption prediction. MIOE provides target ontology queries, class-specific auxiliary attributes, and prompt variants for evaluating interchange interventions over ontology-level variables.
    \item We provide a systematic empirical analysis of SAE, DBM/MDBM, and DAS/MDAS on Qwen3 models. Our results show that ontology-level predictions are causally accessible but only partially disentangled, and that multi-task objectives and contiguous multi-layer interventions improve preservation and overall intervention quality.
\end{enumerate}

\section{Related Work}

\subsection{Mechanistic Interpretability}

Mechanistic interpretability aims to reverse-engineer the internal computation of black-box models by describing them in terms of transparent and human-interpretable algorithms~\cite{geiger-et-al-2025-foundation}. A central difficulty is polysemanticity, where a single component or direction can encode multiple features. This difficulty is often linked to superposition: models can represent more features than their available dimensions would otherwise allow. Sparse Autoencoders (SAEs) have therefore become a prominent approach for decomposing activation vectors into sparse latent features that are more amenable to monosemantic interpretation~\cite{bricken-et-al-2023,cunningham-et-al-2023-sae}. Subsequent work has addressed limitations of vanilla SAEs from both architectural~\cite{rajamanoharan-et-al-2024-gated-sae,gao-et-al-2025-topk-sae,bussmann-et-al-2024-batchtopk-sae} and training-strategy perspectives~\cite{ghilardi-et-al-2025-group-sae,braun-et-al-2024-end-to-end-sae}. A complementary line of mechanistic interpretability is grounded in causal abstraction, which asks whether a high-level causal model is faithfully implemented by the low-level computation of a neural network~\cite{geiger-et-al-2021-causal,geiger-et-al-2025-foundation}. Interchange interventions provide an operational test for such alignments by replacing internal representations across inputs and measuring whether the resulting behavior changes according to the corresponding high-level variable. Methods in this family include Differential Binary Masking (DBM), which searches for axis-aligned causal feature masks~\cite{de-cao-et-al-2020-dbm}, and Distributed Alignment Search (DAS), which learns distributed subspaces aligned with interpretable causal variables~\cite{geiger-et-al-2024-das,wu-et-al-2023-boundless-das}. Alongside method development, evaluation has become an increasingly important problem, since interpretability has multiple desiderata and is difficult to define with a single metric~\cite{mueller-et-al-2025-mib}. Recent benchmarks therefore evaluate different aspects of interpretability, including feature disentanglement in RAVEL~\cite{huang-et-al-2024-ravel}, known circuit recovery in InterpBench~\cite{gupta-et-al-2024-interpbench}, and SAE-specific performance across diverse evaluation dimensions in SAEBench~\cite{karvonen-et-al-2025-saebench}. Recent work has also identified failure modes and phenomena that complicate feature-level interpretation, such as feature absorption~\cite{chanin-et-al-2025-absorption}. However, existing evaluations largely focus on entity attributes, factual recall, or synthetic circuits, leaving high-level ontological knowledge under-explored.

\subsection{LLMs for Ontology Engineering}

The rise of LLMs has made them increasingly relevant to ontology engineering tasks that have traditionally relied on ontology engineers, domain experts, and rule-based tools. More broadly, the relationship between LLMs and knowledge graphs has motivated growing interest in how parametric knowledge can be used for, combined with, or evaluated against explicit knowledge representations~\cite{pan-et-al-2023-tgdk,petroni-et-al-2019-lama}. Recent work has investigated the use of LLMs across a wide range of ontology engineering tasks, including requirement elicitation~\cite{zhang-et-al-2025-ontochat}, ontology subsumption inference~\cite{he-et-al-2023-ontolama}, ontology learning~\cite{giglou-et-al-2023-llms4ol,lo-et-al-2024-end-to-end}, ontology alignment~\cite{he-et-al-2022-bertmap}, and ontology evaluation~\cite{tsaneva-et-al-2025}. These studies show that LLMs and LLM-based systems can extract and generate ontological knowledge~\cite{lo-et-al-2024-end-to-end}, improve on complex subsumption inference when a small number of annotated examples are available~\cite{he-et-al-2023-ontolama}, and memorize certain ontological knowledge while using implicit knowledge for reasoning~\cite{wu-et-al-2023-plm}. At the same time, this ability remains imperfect. Prior work reports limitations on specialized ontology tasks~\cite{he-et-al-2023-ontolama,tsaneva-et-al-2025,wu-et-al-2023-plm,zhang-et-al-2025-schema} and on long-tail factual knowledge~\cite{huang-et-al-2025-longtail}. LLMs may perform well on some ontology-related tasks while failing on others, suggesting incomplete and unevenly accessible ontological knowledge~\cite{he-et-al-2023-ontolama,wu-et-al-2023-plm}.

\section{The MIOE Benchmark}

\subsection{Task Formulation}\label{subsection:task}

MIOE is formulated as a causal abstraction benchmark for ontological knowledge. We use two ontology engineering tasks, entity typing and subsumption prediction, to define the high-level causal variables: a subject induces a target taxonomic variable, such as its entity type or superclass, together with non-target background information reflected by class-specific auxiliary attributes. 
The benchmark asks whether a mechanistic interpretability method can identify a low-level intervention target whose interchange intervention matches this high-level counterfactual: replacing the base subject's target taxonomic value with the source subject's value while preserving the base subject's auxiliary values.

Formally, let $\tau$ denote the target axiom type, with $\tau=\texttt{instance of}$ for entity typing and $\tau=\texttt{subclass of}$ for subsumption prediction. Let $\mathcal{S}_{\tau}$ be the set of subjects for axiom type $\tau$, and let $c_{\tau}(s) \subset \mathcal{C}_{\tau}$ be the ground-truth set of ontology classes for subject $s$.
For $\tau = \texttt{instance of}$, each subject $s \in \mathcal{S}_{\tau}$ is an entity, and the model predicts a class label $y_{\tau}(s)$.
For $\tau = \texttt{subclass of}$, each subject $s$ is a class or concept, and the model predicts a parent class $y_{\tau}(s)$.
In addition to the target ontological relation, each subject $s$ is associated with a set of class-specific auxiliary attributes $\mathcal{A}(s)$. For each attribute $A \in \mathcal{A}(s)$, the model predicts an attribute value $y_A(s)$, with ground-truth value $v_A(s)$. 
We write $x=p(s)$ for the instantiated model input obtained by applying prompt template $p$ to subject $s$, with $p\in\{p_{\tau},p_A,p_w\}$ denoting target-relation, auxiliary-attribute, or Wikipedia-derived contextual prompts.

The mechanistic interpretability task tests whether this high-level causal description can be supported by low-level interventions in the language model. For each target axiom type $\tau$, we ask whether there exists a low-level intervention target $F_{\tau}$ such that interchange interventions on $F_{\tau}$ produce the behavior predicted by the high-level model: the target ontological prediction transfers from the source subject, while class-specific auxiliary predictions remain those of the base subject.
To test this, we use interchange interventions between a base subject $s_{\mathrm{base}}$ and a source subject $s_{\mathrm{source}}$ such that $c_{\tau}(s_{\mathrm{base}})$ and $c_{\tau}(s_{\mathrm{source}})$ differ.
Let $\mathcal{M}$ denote the model under analysis. For a generic low-level intervention target $F$, base input $x_{\mathrm{base}}$, and source input $x_{\mathrm{source}}$, we define the interchange intervention as
\begin{equation}
\begin{aligned}
\mathrm{II}(\mathcal{M}, F, x_{\mathrm{base}}, &x_{\mathrm{source}}) := \\
&\mathcal{M}_{F \leftarrow \operatorname{GetFeature}(\mathcal{M}(x_{\mathrm{source}}),F)}(x_{\mathrm{base}})
\end{aligned}
\end{equation}
where the value of $F$ in the computation for $x_{\mathrm{base}}$ is replaced with the value induced by $x_{\mathrm{source}}$.
A successful intervention target $F_{\tau}$ should satisfy two requirements. First, it should support \emph{target-answer replacement}: when the base input queries the target relation, intervening with the source should change the prediction from the base subject's target label to the source subject's target label.
Second, it should satisfy \emph{auxiliary invariance}: when the base input queries an auxiliary attribute, the same intervention should preserve the base subject's attribute value.
For example, in \texttt{instance of}, the base class assertion may be $\mathrm{city}(\text{Amsterdam})$, while the source class assertion is $\mathrm{country}(\text{Australia})$. After intervening with the source, a target query  such as \emph{Amsterdam is an instance of} should output \emph{country}, whereas an auxiliary prompt such as \emph{``Amsterdam'' belongs to the country of} should still output the base auxiliary value \emph{Netherlands}.

\subsection{Dataset}
\label{section:dataset}

\begin{table*}[t]
  \begin{center}
  \begin{small}
  \setlength{\tabcolsep}{4.5pt}
  \begin{tabular}{lccccc}
    \toprule
    \textbf{Axiom Type} &
    \textbf{\# Classes} &
    \textbf{\# Subjects} &
    \textbf{\makecell{\# Attributes \\ (avg./class)}} &
    \textbf{\makecell{Avg. \# Attribute Values \\ (per subj. / per attr.)}} &
    \textbf{\makecell{\# Prompt Templates \\ (avg./class)}} \\
    \midrule
    \texttt{instance of} & 54 & 1,099 & 376 (7.0) & 12.0 / 2.0 & 4,288 (79.4) \\
    \texttt{subclass of} & 39 & 1,113 & 160 (4.1) & 51.8 / 17.1 & 2,095 (53.7) \\
    \bottomrule
  \end{tabular}
  \end{small}
  \end{center}
  \caption{
    Dataset statistics by axiom type, including counts of domain classes, subjects, auxiliary attributes and values, and prompt templates. For auxiliary attributes, we report the average number of values per subject and per instantiated attribute.
  }
  \label{table:dataset}
\end{table*}

We construct the MIOE dataset for evaluating interchange interventions over the two target axiom types. Each axiom-type subset is organized around subjects, ontology target classes, class-specific auxiliary attributes and values, and prompt templates. Dataset statistics are reported in Table~\ref{table:dataset}, with additional construction details provided in the supplementary materials. The dataset is sampled and refined from Wikidata~\cite{vrandecic-2012-wikidata}, WordNet~\cite{fellbaum-2005-wordnet}, and schema.org~\cite{guha-et-al-2016-schemaorg}.
To support cross-class interventions, the subject sets are organized around diverse high-level classes: 54 classes for \texttt{instance of} and 39 classes for \texttt{subclass of}. For each subject, we provide a set of class-specific attributes and values for evaluating auxiliary invariance. Detailed attribute information is provided in the supplementary materials.

The prompt template set consists of two parts, following a RAVEL-style sampling strategy~\cite{huang-et-al-2024-ravel}. The first part consists of task-specific templates, including templates for entity typing, subsumption prediction, and attribute-based queries. The second part consists of Wikipedia-based prompt samples drawn from Wikipedia corpus sentences containing the corresponding subjects~\cite{wikidump}. The two prompt sets play complementary roles in the benchmark. The ontology-oriented templates provide controlled queries for both target axioms and auxiliary attributes, while the Wikipedia-based prompts place the same subjects in more natural contexts. This separation allows us to evaluate whether the identified intervention targets remain effective beyond prompt formats and under contextualized surface realizations.

We use these prompt families to define two evaluation settings. The Entity setting evaluates held-out subjects while reusing training prompt templates, whereas the Context setting evaluates held-out prompt templates while reusing training subjects. In both settings, the source input is instantiated with a source subject and may use either an ontology-template prompt or a Wikipedia-derived contextual prompt.

\subsection{Evaluation}
\label{section:evaluation}

Because target class labels and auxiliary attribute values can admit multiple valid surface forms, we evaluate predictions against accepted ground-truth answer sets. A target relation prediction is counted as correct when $y_{\tau}(s) \in c_{\tau}(s)$, and an auxiliary prediction is counted as correct when $y_A(s) \in v_A(s)$. Following the benchmark construction, we evaluate interventions only on retained prompt instances where the unintervened model already produces an accepted answer. Base-model accuracies before this filtering step are reported in the supplementary materials.

The task formulation above defines the intervention operator and the two desired behaviors. We instantiate them as Cause and Iso metrics, following~\cite{huang-et-al-2024-ravel} and adapting them to ontological relations.
Let $\mathcal{E}^{\mathrm{cause}}_{\tau}$ denote the retained Cause examples for axiom type $\tau$. Each example $e \in \mathcal{E}^{\mathrm{cause}}_{\tau}$ contains a base subject $s_{\mathrm{base}}$, a source subject $s_{\mathrm{source}}$, a target-relation base input $x_{\mathrm{base}}=p_{\tau}(s_{\mathrm{base}})$, and a source input $x_{\mathrm{source}}$ instantiated with $s_{\mathrm{source}}$. The example-level Cause indicator is
\begin{equation}
\mathrm{Cause}(e) \coloneqq
\mathbb{I}\!\left(
\mathrm{II}(\mathcal{M}, F_{\tau}, x_{\mathrm{base}}, x_{\mathrm{source}})
\in
c_{\tau}(s_{\mathrm{source}})
\right).
\end{equation}

Similarly, let $\mathcal{E}^{\mathrm{iso}}_{\tau}$ denote the retained auxiliary-invariance examples. Each example $e \in \mathcal{E}^{\mathrm{iso}}_{\tau}$ contains a base subject $s_{\mathrm{base}}$, a source subject $s_{\mathrm{source}}$, an auxiliary attribute $A\in\mathcal{A}(s_{\mathrm{base}})$, an auxiliary base input $x_{\mathrm{base}}=p_A(s_{\mathrm{base}})$, and a source input $x_{\mathrm{source}}$ instantiated with $s_{\mathrm{source}}$. The example-level Iso indicator is
\begin{equation}
\mathrm{Iso}(e) \coloneqq
\mathbb{I}\!\left(
\mathrm{II}(\mathcal{M}, F_{\tau}, x_{\mathrm{base}}, x_{\mathrm{source}})
\in
v_A(s_{\mathrm{base}})
\right).
\end{equation}

We aggregate scores at the task level rather than directly over raw examples. Each group contains retained intervention examples sharing the same base prompt template, split, and evaluation type. We first average example-level success within each group and then macro-average across groups, preventing prompt groups with more retained examples from dominating the reported score. 
Let $\mathcal{G}^{\mathrm{cause}}_{\tau}$ and $\mathcal{G}^{\mathrm{iso}}_{\tau}$ denote the resulting sets of Cause and Iso evaluation groups, respectively. We define
\begin{equation}
\mathrm{Cause}_{\tau}
\coloneqq
\frac{1}{|\mathcal{G}^{\mathrm{cause}}_{\tau}|}
\sum_{g \in \mathcal{G}^{\mathrm{cause}}_{\tau}}
\frac{1}{|g|}
\sum_{e \in g} \mathrm{Cause}(e),
\end{equation}

\begin{equation}
\mathrm{Iso}_{\tau}
\coloneqq
\frac{1}{|\mathcal{G}^{\mathrm{iso}}_{\tau}|}
\sum_{g \in \mathcal{G}^{\mathrm{iso}}_{\tau}}
\frac{1}{|g|}
\sum_{e \in g} \mathrm{Iso}(e),
\end{equation}
and the final disentanglement score as
\begin{equation}
\mathrm{Disentangle}_{\tau}
\coloneqq
\frac{1}{2}\left(\mathrm{Cause}_{\tau} + \mathrm{Iso}_{\tau}\right).
\end{equation}

\section{Experiments}

\subsection{Mechanistic Interpretability Methods}
\label{sec:mechanistic-methods}

A range of mechanistic interpretability methods can be instantiated in MIOE. We evaluate SAE, DBM, and DAS, along with the multi-task variants MDBM and MDAS, following prior work~\cite{huang-et-al-2024-ravel}, and extend DAS/MDAS to multi-layer settings. 
Most methods operate on a single residual-stream intervention site: a fixed layer $\ell$ and the subject-token position determined by each prompt template. Multi-layer DAS and MDAS instead intervene on a small layer window, using separate subspaces at each layer.
For a single-layer intervention site, each intervention example $e$ induces a base activation $z = h^{(\ell)}(x_{\mathrm{base}}^{(e)})$ and a source activation $z' = h^{(\ell)}(x_{\mathrm{source}}^{(e)})$.
Each method defines an intervention target $F_\tau$, together with an intervention operator $\Gamma_{\vartheta}(z,z')$ that produces an intervened representation $\tilde z$. Replacing $z$ with $\tilde z$ in the forward pass instantiates the interchange intervention operator $\mathrm{II}(\mathcal{M},F_{\tau},x_{\mathrm{base}},x_{\mathrm{source}})$ defined in the task formulation.

\paragraph{SAE}
SAE provides a two-stage alternative to counterfactual training. Instead of directly optimizing an intervention subspace against cause or iso supervision, SAE first learns a sparse latent representation of activations at the intervention site, and only then identifies a subset of latent features associated with the target ontological relation. Let $a = E_{\phi}(z) \in \mathbb{R}^{d_{\mathrm{lat}}}$ be the encoder output and $\hat z = D_{\psi}(a)$ the decoder reconstruction, where $d_{\mathrm{lat}}$ is the latent dimension and the encoder includes a nonlinearity that encourages non-negative sparse activations. The autoencoder is trained on unlabeled activations by minimizing
\[
\mathcal{L}_{\mathrm{SAE}}
=
\mathbb{E}_{z}
\Big[
\|z - D_{\psi}(E_{\phi}(z))\|_2^2
\;+\;
\beta \|E_{\phi}(z)\|_1
\Big].
\]
After learning the latent space, SAE instantiates the intervention target as a latent feature set $F_{\tau} \subseteq \{1,\dots,d_{\mathrm{lat}}\}$, selected by fitting a sparse linear classifier on latent activations to predict the target ontological labels. The resulting intervention is performed in latent space:
\[
\tilde a_i
=
\begin{cases}
E_{\phi}(z')_i, & i \in F_{\tau},\\
E_{\phi}(z)_i, & i \notin F_{\tau},
\end{cases}
\qquad
\Gamma_{\mathrm{SAE}}(z,z') = D_{\psi}(\tilde a).
\]

\paragraph{DBM}
DBM instantiates $F_{\tau}$ as an axis-aligned subset of residual-stream coordinates in the original representation basis, learning a sparse selector over the $d$ coordinates of $z$.
Let $\alpha \in \mathbb{R}^d$ be mask logits, $\pi = \sigma(\alpha / T) \in [0,1]^d$ the relaxed mask at temperature $T$. We write $P_{\pi}=\mathrm{diag}(\pi)$ and let $I\in\mathbb{R}^{d\times d}$ denote the identity matrix. The DBM intervention is
\[
\Gamma_{\mathrm{DBM}}(z,z')
=
(I-P_{\pi})z + P_{\pi}z'
=
z + P_{\pi}(z' - z).
\]
The relaxed mask gives a coordinate-wise interpolation between base and source activations; in the hard-mask limit, selected coordinates are copied from source to base while the remaining coordinates are preserved. The DBM objective combines supervision on retained Cause examples with sparsity regularization on $\pi$:
\[
\begin{aligned}
&\mathcal{L}_{\mathrm{DBM}} = \\
&\mathbb{E}_{e\sim \mathcal{E}^{\mathrm{cause}}_{\tau}}
\left[
\mathrm{CE}\left(
\mathcal{M}_{z\leftarrow \Gamma_{\mathrm{DBM}}(z,z')}
(x_{\mathrm{base}}^{(e)}),
c_{\tau}(s_{\mathrm{source}}^{(e)})
\right)
\right] \\
&\qquad + \lambda\|\pi\|_1 .
\end{aligned}
\]

\paragraph{DAS}
DAS assumes that the target concept may be distributed across neurons and should therefore be modeled as a low-dimensional linear subspace. We directly parameterize this feature subspace instead of learning a full $d \times d$ rotation. Let $W \in \mathbb{R}^{k \times d}$ be a row-orthonormal matrix whose row span instantiates the intervention target $F_\tau$. The corresponding orthogonal projector is $P_W = W^\top W$, and the intervention is
\[
\Gamma_{\mathrm{DAS}}(z,z')
=
(I-P_W)z + P_W z'
=
z + W^\top W (z' - z).
\]
This exchanges only the component of the representation that lies in the learned subspace, while leaving the orthogonal complement unchanged. DAS is trained on retained Cause examples:
\[
\begin{aligned}
&\mathcal{L}_{\mathrm{DAS}} = \\
&\mathbb{E}_{e \sim \mathcal{E}^{\mathrm{cause}}_{\tau}}
\left[
\mathrm{CE}\left(
\mathcal{M}_{z\leftarrow \Gamma_{\mathrm{DAS}}(z,z')}
(x_{\mathrm{base}}^{(e)}),
c_{\tau}(s_{\mathrm{source}}^{(e)})
\right)
\right]
\end{aligned}
\]

\paragraph{Multi-task DAS \& DBM}
The objectives above train the intervention to change the target answer but do not explicitly train it to preserve auxiliary answers. We therefore use multi-task variants, MDBM and MDAS, that additionally include auxiliary-invariance examples~\cite{huang-et-al-2024-ravel}. For each training example $e$, define the desired output
\[
y^\star(e)
=
\begin{cases}
c_{\tau}(s_{\mathrm{source}}^{(e)}), & e \in \mathcal{E}^{\mathrm{cause}}_{\tau}, \\
v_{A^{(e)}}(s_{\mathrm{base}}^{(e)}), & e \in \mathcal{E}^{\mathrm{iso}}_{\tau},
\end{cases}
\]
where $A^{(e)}$ is the auxiliary attribute queried in example $e$. The generic multi-task objective is
\[
\begin{aligned}
&\mathcal{L}_{\mathrm{MT}}(\vartheta) = \\
&\mathbb{E}_{e \sim \mathcal{E}^{\mathrm{cause}}_{\tau} \cup \mathcal{E}^{\mathrm{iso}}_{\tau}}
\left[
\mathrm{CE}\left(
\mathcal{M}_{z\leftarrow \Gamma_\vartheta(z,z')}
(x_{\mathrm{base}}^{(e)}),
y^\star(e)
\right)
\right] \\
&\qquad + \Omega(\vartheta).
\end{aligned}
\]
Here $\vartheta$ denotes the method-specific parameters that instantiate the single-site intervention target $F_{\tau}$, e.g., mask logits for DBM/MDBM or a subspace matrix for DAS/MDAS. 
MDBM instantiates $\Gamma_{\vartheta}$ with $\Gamma_{\mathrm{DBM}}$ and $\Omega(\vartheta)=\lambda\|\pi\|_1$, while MDAS instantiates $\Gamma_{\vartheta}$ with $\Gamma_{\mathrm{DAS}}$ and $\Omega(\vartheta)=0$. This trains the intervention to satisfy both target-answer replacement and auxiliary invariance.

\paragraph{Multi-Layer DAS \& MDAS}
We extend DAS and MDAS to an ordered layer set $S=(\ell_1,\dots,\ell_m)$, allowing the intervention target to span adjacent residual-stream layers. For each $\ell_j\in S$, let $s_j$ denote the source representation at the prompt-specific subject-token position, and let $b_j$ denote the base representation encountered at the same layer during the intervened forward pass. The layer-specific intervention is
\[
\begin{aligned}
\Gamma_{\vartheta_j}^{(\ell_j)}(b_j,s_j)
&= b_j + P_{\vartheta_j}^{(\ell_j)}(s_j-b_j),\\
P_{\vartheta_j}^{(\ell_j)}
&= (W^{(\ell_j)})^\top W^{(\ell_j)} .
\end{aligned}
\]
The multi-layer intervention $\Gamma_{\Theta_S}^{S}$, with layer-specific parameters $\Theta_S=\{\vartheta_j\}_{j=1}^m$ and $\vartheta_j=W^{(\ell_j)}$ for DAS/MDAS, applies these layer-specific interventions across the ordered layer set. This yields a joint multi-site intervention: all layer-specific subspaces are optimized together, and lower-layer interventions can affect the later base states encountered during the same forward pass.
For multi-layer DAS, the training objective is
\[
\begin{aligned}
&\mathcal{L}_{\mathrm{ML\text{-}DAS}}(\Theta_S) = \\
&\mathbb{E}_{e\sim\mathcal{E}^{\mathrm{cause}}_{\tau}}
\left[
\mathrm{CE}\left(
\mathcal{M}_{S\leftarrow \Gamma_{\Theta_S}^{S}}
(x_{\mathrm{base}}^{(e)}),
c_\tau(s_{\mathrm{source}}^{(e)})
\right)
\right].
\end{aligned}
\]
For multi-layer MDAS, the corresponding objective is
\[
\begin{aligned}
&\mathcal{L}_{\mathrm{ML\text{-}MDAS}}(\Theta_S) = \\
&\mathbb{E}_{e \sim \mathcal{E}^{\mathrm{cause}}_{\tau} \cup \mathcal{E}^{\mathrm{iso}}_{\tau}}
\left[
\mathrm{CE}\left(
\mathcal{M}_{S\leftarrow \Gamma_{\Theta_S}^{S}}
(x_{\mathrm{base}}^{(e)}),
y^\star(e)
\right)
\right].
\end{aligned}
\]

\begin{table*}[t]
\centering
\small
\setlength{\tabcolsep}{2pt}
\renewcommand{\arraystretch}{1.05}
\begin{tabular}{l cc ccc ccc cc ccc ccc}
\toprule
\multirow{4}{*}{\textbf{Method}} &
\multicolumn{8}{c}{\textbf{\texttt{instance of}}} &
\multicolumn{8}{c}{\textbf{\texttt{subclass of}}} \\
\cmidrule(r){2-9}\cmidrule(l){10-17}
&
\multirow{2}{*}{\textbf{Dimension}} &
\multirow{2}{*}{\textbf{Layer}} &
\multicolumn{3}{c}{\textbf{Entity}} &
\multicolumn{3}{c}{\textbf{Context}} &
\multirow{2}{*}{\textbf{Dimension}} &
\multirow{2}{*}{\textbf{Layer}} &
\multicolumn{3}{c}{\textbf{Entity}} &
\multicolumn{3}{c}{\textbf{Context}} \\
\cmidrule(r){4-6}\cmidrule(r){7-9}\cmidrule(r){12-14}\cmidrule(l){15-17}
& & & \textbf{C} & \textbf{I} & \textbf{D}
& \textbf{C} & \textbf{I} & \textbf{D}
& & & \textbf{C} & \textbf{I} & \textbf{D}
& \textbf{C} & \textbf{I} & \textbf{D} \\
\midrule
SAE  & 8192  & 13 & 0.472 & 0.352 & 0.412 & 0.399 & 0.344 & 0.371 & 16384 & 12 & 0.584 & 0.210 & 0.397 & 0.624 & 0.233 & 0.429 \\
DBM  & 4096  & 10 & 0.441 & 0.409 & 0.425 & 0.385 & 0.407 & 0.396 & 4096  & 10 & 0.528 & 0.334 & 0.431 & 0.526 & 0.359 & 0.443 \\
MDBM & 4096  & 10 & 0.418 & 0.469 & 0.443 & 0.344 & 0.472 & 0.408 & 4096  & 13 & 0.491 & 0.389 & 0.440 & 0.490 & 0.409 & 0.450 \\
DAS  & 1024  & 17 & 0.589 & 0.579 & 0.584 & 0.582 & 0.512 & 0.547 & 1024  & 22 & 0.509 & 0.483 & 0.496 & 0.625 & 0.445 & 0.535 \\
MDAS & 2048  & 17 & 0.527 & 0.658 & \textbf{0.592} & 0.481 & 0.692 & \textbf{0.587} & 2048 & 18 & 0.496 & 0.530 & \textbf{0.513} & 0.525 & 0.560 & \textbf{0.542} \\
\bottomrule
\end{tabular}
\caption{Best single-layer intervention results for Qwen3-8B. C/I/D denote Cause/Iso/Disentangle. For each axiom type and method, we report the layer/configuration with the strongest Entity Disentangle score. Dimension denotes subspace rank for DAS/MDAS, ambient mask dimension for DBM/MDBM, and latent width for SAE. Qwen3-4B results are reported in the supplementary materials.}
\label{tab:single-layer-best}
\end{table*}

\subsection{Setup}

We evaluate Qwen3-4B and Qwen3-8B, two 36-layer decoder-only Transformers with hidden sizes 2,560 and 4,096, respectively~\cite{qwen3}. The single-layer experiments cover both axiom types and include SAE, DBM, MDBM, DAS, and MDAS. All methods intervene on block-output residual-stream activations at the prompt-specific subject-token position. For single-layer experiments, we evaluate each of the 36 layers as the intervention layer. The multi-layer experiments evaluate DAS and MDAS on Qwen3-8B for both axiom types, using contiguous two- and four-layer windows. For multi-layer DAS/MDAS, we use $k\times w$ to denote rank-$k$ subspaces applied at each layer of a contiguous $w$-layer window. We evaluate $256\times4$, $512\times2$, and $512\times4$, giving total intervention dimensions of 1024, 1024, and 2048, respectively, comparable to the single-layer DAS/MDAS ranks. Full configurations are provided in the supplementary materials.

\section{Results}

We first evaluate whether learned intervention targets support the expected intervention behavior for the target ontology variable in entity typing and subsumption prediction, including evaluation on both Entity and Context prompt settings. We then analyze whether this behavior remains confined to the target axiom by checking preservation of class-specific auxiliary variables. Finally, we examine how method choice, intervention dimension, representation layer, multi-layer windows, and ontology-class structure affect the balance between target-answer change and auxiliary preservation.

Table~\ref{tab:single-layer-best} summarizes the strongest Qwen3-8B single-layer results. Distributed subspace methods give the best overall performance, with MDAS achieving the highest Disentangle scores across both axiom types. Its Disentangle scores on Entity and Context are 0.592/0.587 for \texttt{instance of} and 0.513/0.542 for \texttt{subclass of}. This suggests that the identified intervention targets are not tied only to the entity-prompt split used in the main comparison. The Qwen3-4B results provided in the supplementary materials preserve the same method ordering, indicating that the result is not specific to the larger checkpoint.

The Iso scores reveal that methods without explicit auxiliary-preservation supervision, such as SAE and DBM, can produce non-trivial effects on target ontology predictions while often perturbing auxiliary attribute values. Multi-task variants directly address this issue by adding preservation supervision, and the effect is visible in Table~\ref{tab:single-layer-best}: across the Entity and Context columns, the best reported multi-task configurations improve Iso over the corresponding single-task method family by 0.081 on average, while reducing Cause by 0.052 on average. Disentangle nevertheless improves because the gain in preserving non-target information outweighs the reduction in Cause. The same Cause--Iso tension is visible along the intervention-size axis: within the reported DAS/MDAS and SAE settings, increasing the intervention dimension can improve Cause while reducing Iso in some configurations, rather than uniformly improving both.

\begin{figure}[t]
  \centering
  \includegraphics[width=0.70\linewidth]{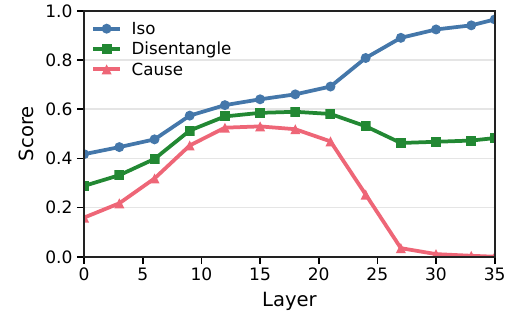}
  \includegraphics[width=0.70\linewidth]{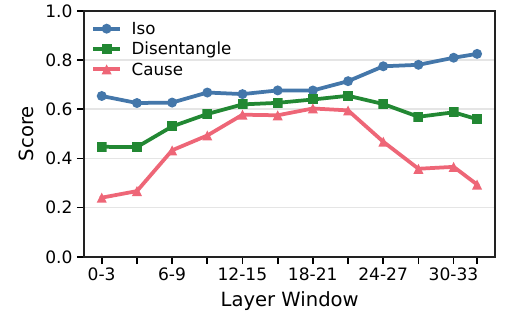}
  \caption{Representative layer profiles for Qwen3-8B on \texttt{instance of}. The top panel shows single-layer MDAS with rank 2048, and the bottom panel shows four-layer MDAS with $256\times4$ contiguous windows. Both panels report Cause, Iso, and Disentangle.}
  \label{fig:layer-profile-main}
\end{figure}

\begin{figure}[t]
  \centering
  \includegraphics[width=0.75\linewidth]{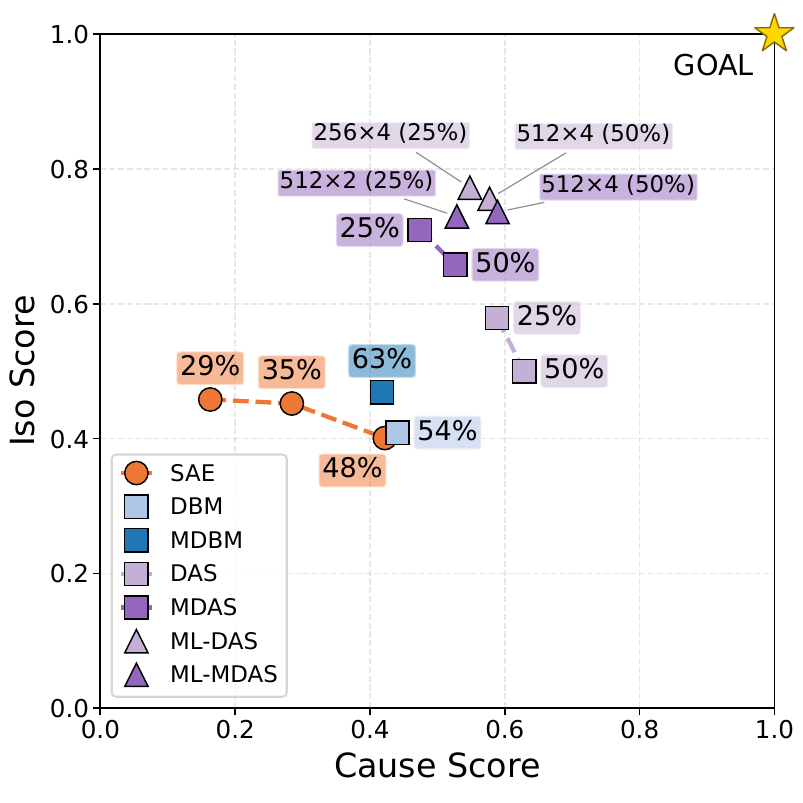}
  \caption{Cause--Iso scores for selected Qwen3-8B \texttt{instance of} configurations with different intervention dimensions. Percent labels indicate the dimensionality of $F_{\tau}$ under each method's parameterization; for multi-layer interventions, total rank is reported relative to a single-layer residual-stream width.}
  \label{fig:feature-size-main}
\end{figure}

\begin{table}[t]
\centering
\small
\setlength{\tabcolsep}{3pt}
\renewcommand{\arraystretch}{1.05}
\begin{tabular}{clcccc}
\toprule
\textbf{$k\times w$} & \textbf{Method} & \textbf{Window} & \textbf{Cause} & \textbf{Iso} & \textbf{Disentangle ($\Delta$)} \\
\midrule
\rowcolor{black!8}\multicolumn{6}{c}{\textbf{\texttt{instance of}}} \\
\midrule
\multirow{2}{*}{$512\times2$} & DAS  & 23--24 & 0.560 & 0.726 & 0.643 (+0.059) \\
& MDAS & 22--23 & 0.529 & 0.730 & 0.629 (+0.038) \\
\multirow{2}{*}{$256\times4$} & DAS  & 24--27 & 0.548 & 0.772 & 0.660 (+0.077) \\
& MDAS & 22--25 & 0.590 & 0.741 & 0.666 (+0.074) \\
\multirow{2}{*}{$512\times4$} & DAS  & 24--27 & 0.577 & 0.756 & \textbf{0.667} (\textbf{+0.102}) \\
& MDAS & 23--26 & 0.589 & 0.736 & 0.663 (+0.070) \\
\midrule
\rowcolor{black!8}\multicolumn{6}{c}{\textbf{\texttt{subclass of}}} \\
\midrule
\multirow{2}{*}{$512\times2$} & DAS  & 23--24 & 0.468 & 0.559 & 0.514 (+0.018) \\
& MDAS & 22--23 & 0.440 & 0.617 & 0.529 (+0.036) \\
\multirow{2}{*}{$256\times4$} & DAS  & 22--25 & 0.555 & 0.473 & 0.514 (+0.018) \\
& MDAS & 21--24 & 0.486 & 0.616 & 0.551 (+0.058) \\
\multirow{2}{*}{$512\times4$} & DAS  & 23--26 & 0.537 & 0.513 & 0.525 (+0.033) \\
& MDAS & 22--25 & 0.537 & 0.593 & \textbf{0.565} (+0.052) \\
\bottomrule
\end{tabular}
\caption{Multi-layer DAS/MDAS results for Qwen3-8B. The $k\times w$ column denotes a rank-$k$ subspace at each layer in a contiguous $w$-layer window, and Window gives the contiguous layer span of the reported best configuration. $\Delta$ reports the Disentangle improvement over the single-layer result with the same method and total dimension.}
\label{tab:instance-multilayer}
\end{table}

\begin{figure*}[t]
  \centering
  \includegraphics[height=0.31\linewidth]{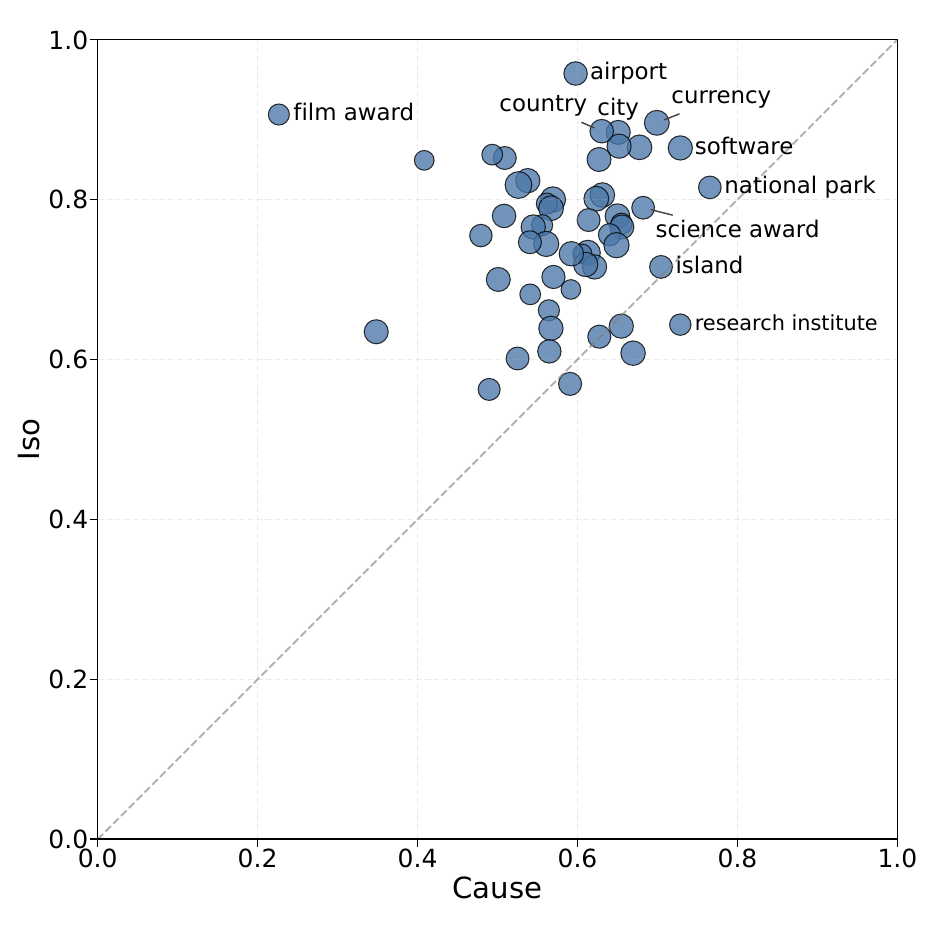}
  \hfil
  \includegraphics[height=0.31\linewidth]{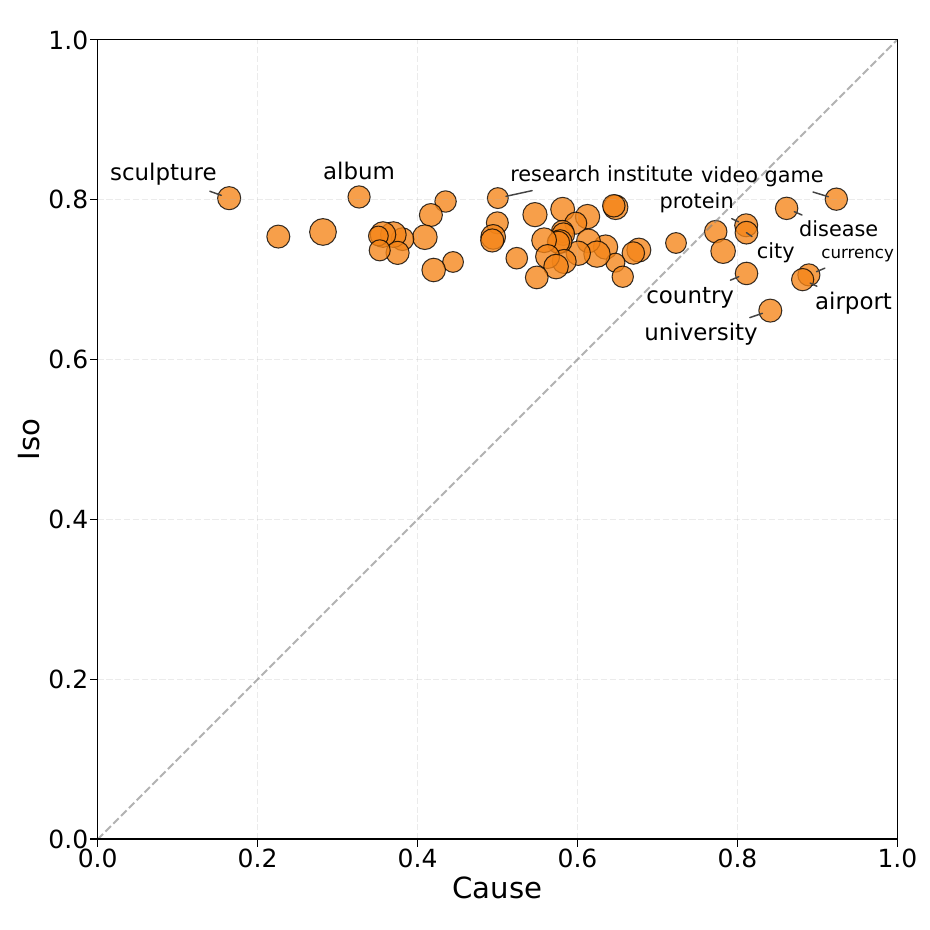}
  \hfil
  \includegraphics[height=0.308\linewidth]{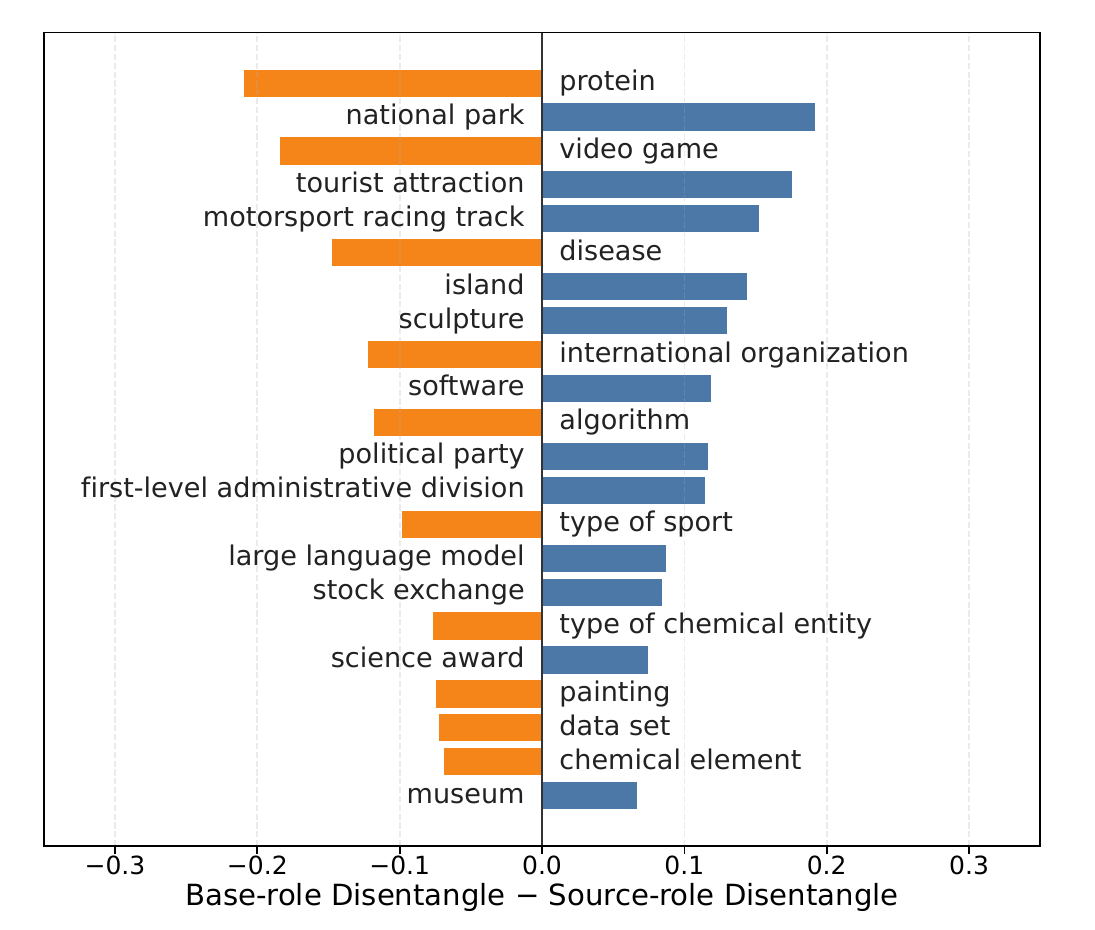}
  \caption{Class-level analysis of Qwen3-8B multi-layer MDAS for \texttt{instance of}. Left and middle: Cause--Iso scatter plots for base-role and source-role classes. Right: largest base--source role gaps, measured as base-role minus source-role Disentangle. Colors denote stronger base-role performance \raisebox{-0.15ex}{\textcolor[HTML]{4C78A8}{\rule{1.2em}{0.7em}}} or stronger source-role performance \raisebox{-0.15ex}{\textcolor[HTML]{F58518}{\rule{1.2em}{0.7em}}}.}
  \label{fig:class-level-main}
\end{figure*}

Figure~\ref{fig:layer-profile-main} illustrates that the target-answer intervention effect is highly layer-dependent. Full layer-wise profiles for SAE, single-layer interventions, and multi-layer interventions are provided in the supplementary materials. Across methods, Cause is weak in the earliest layers, rises through the middle of the model, and often reaches its strongest values in middle or middle-to-late layers. Later layers often obtain strong Iso, but Cause can simultaneously fall to very low values, suggesting that the intervention may no longer be effectively changing the model's generated output.

Contiguous layer-window interventions further improve over comparable single-layer settings, as shown in Table~\ref{tab:instance-multilayer}. The table reports both axiom types and gives $\Delta$ against the single-layer result with the same method and total dimension. Figure~\ref{fig:feature-size-main} provides the same view geometrically: the multi-layer configurations move upward in Cause--Iso space relative to the corresponding single-layer DAS/MDAS points, mainly by improving Iso while retaining substantial Cause. On Qwen3-8B \texttt{instance of}, the strongest rows are four-layer windows: DAS $512\times4$ reaches 0.667 Disentangle, while MDAS $256\times4$ and $512\times4$ reach 0.666 and 0.663. The gain is not explained by total dimensionality alone: at total dimension 1024, $256\times4$ outperforms $512\times2$ for both DAS and MDAS. These results suggest that the relevant intervention target is better captured across a short middle-to-late layer interval than at a single residual-stream site.

Furthermore, Figure~\ref{fig:class-level-main} and additional class-level analyses in the supplementary materials show that intervention success varies substantially across ontology classes. Figure~\ref{fig:class-level-main}(a,b) shows that base-role and source-role classes have different Cause--Iso profiles. Classes with concrete and regular attribute profiles, such as \texttt{airport}, \texttt{country}, \texttt{currency}, and \texttt{university}, often achieve higher class-level Disentangle, whereas broader or more heterogeneous classes, such as \texttt{body of water}, \texttt{geographical feature}, \texttt{food}, \texttt{biological process}, and \texttt{material}, show lower scores and appear harder to change at the target level without also affecting auxiliary information. Figure~\ref{fig:class-level-main}(c) further shows role asymmetry: \texttt{protein} and \texttt{video game} are stronger as sources, whereas \texttt{national park} and \texttt{tourist attraction} are stronger as bases. This indicates that intervention difficulty depends not only on the class itself, but also on whether it supplies or receives the intervened representation. The supplementary analyses further show that similar class rankings recur across Qwen3-4B and Qwen3-8B, suggesting that this variation is not merely an artifact of one checkpoint. This gives MIOE a diagnostic role beyond aggregate method comparison: it identifies where ontology-level interventions are more effective and where they remain more tightly coupled to class-specific auxiliary information. Taken together, the results suggest that ontology-oriented predictions are causally accessible through learned intervention targets, while their separation from auxiliary class-specific information remains partial and class-dependent.

\section{Conclusion}

We introduced MIOE, a causal abstraction benchmark for analyzing ontological knowledge in language models through interchange interventions. MIOE pairs target ontology variables from entity typing and subsumption prediction with class-specific auxiliary attributes, allowing us to test whether an internal intervention changes the target answer while preserving non-target information. 

Our experiments show that ontology-level predictions are mechanistically accessible but only partially disentangled. Multi-task objectives improve preservation of auxiliary answers, and contiguous layer-window interventions further improve over comparable single-layer settings, suggesting that the relevant computation is better captured across short middle-to-late layer spans than at a single residual-stream site. The class-level analysis also shows that intervention success varies substantially across ontology classes, indicating that aggregate scores hide meaningful differences in how different parts of the ontology are represented.

Overall, these findings support a qualified interpretation of LLMs for ontology engineering. Correct ontological predictions do not by themselves imply a clean or modular internal mechanism for the corresponding axiom. Instead, MIOE suggests that ontological variables are causally accessible, but not cleanly separated from broader class-specific information, either because the representations are distributed and entangled with class-specific semantics or because current intervention-search methods remain insufficiently precise. Future work should extend this analysis to richer ontology constraints, broader model families, and settings that test whether ontology-level intervention behavior remains stable across domains, prompts, and complex reasoning contexts.

\bibliography{aaai2027}

@inproceedings{huang-et-al-2024-ravel,
    title = {{RAVEL: Evaluating Interpretability Methods on Disentangling Language Model Representations}},
    author = "Huang, Jing  and
      Wu, Zhengxuan  and
      Potts, Christopher  and
      Geva, Mor  and
      Geiger, Atticus",
    editor = "Ku, Lun-Wei  and
      Martins, Andre  and
      Srikumar, Vivek",
    booktitle = "Proceedings of the 62nd Annual Meeting of the Association for Computational Linguistics (Volume 1: Long Papers)",
    month = aug,
    year = "2024",
    address = "Bangkok, Thailand",
    publisher = "Association for Computational Linguistics",
    url = "https://aclanthology.org/2024.acl-long.470",
    pages = "8669--8687",
}

@misc{karvonen-et-al-2025-saebench,
    title={{SAEBench: A Comprehensive Benchmark for Sparse Autoencoders in Language Model Interpretability}}, 
    author={Adam Karvonen and Can Rager and Johnny Lin and Curt Tigges and Joseph Bloom and David Chanin and Yeu-Tong Lau and Eoin Farrell and Callum McDougall and Kola Ayonrinde and Matthew Wearden and Arthur Conmy and Samuel Marks and Neel Nanda},
    year={2025},
    eprint={2503.09532},
    archivePrefix={arXiv},
    primaryClass={cs.LG},
    url={https://arxiv.org/abs/2503.09532}, 
}

@inproceedings{gupta-et-al-2024-interpbench,
    author = {Gupta, Rohan and Arcuschin, Iv\'{a}n and Kwa, Thomas and Garriga-Alonso, Adri\`{a}},
    booktitle = {Advances in Neural Information Processing Systems},
    doi = {10.52202/079017-2950},
    editor = {A. Globerson and L. Mackey and D. Belgrave and A. Fan and U. Paquet and J. Tomczak and C. Zhang},
    pages = {92922--92951},
    publisher = {Curran Associates, Inc.},
    title = {{InterpBench: Semi-Synthetic Transformers for Evaluating Mechanistic Interpretability Techniques}},
    url = {https://proceedings.neurips.cc/paper_files/paper/2024/file/a8f7d43ae092d9a5295775eb17f3f4f7-Paper-Datasets_and_Benchmarks_Track.pdf},
    volume = {37},
    year = {2024}
}

@inproceedings{mueller-et-al-2025-mib,
    title={{MIB: A Mechanistic Interpretability Benchmark}},
    author={Aaron Mueller and Atticus Geiger and Sarah Wiegreffe and Dana Arad and Iv{\'a}n Arcuschin and Adam Belfki and Yik Siu Chan and Jaden Fried Fiotto-Kaufman and Tal Haklay and Michael Hanna and Jing Huang and Rohan Gupta and Yaniv Nikankin and Hadas Orgad and Nikhil Prakash and Anja Reusch and Aruna Sankaranarayanan and Shun Shao and Alessandro Stolfo and Martin Tutek and Amir Zur and David Bau and Yonatan Belinkov},
    booktitle={Forty-second International Conference on Machine Learning},
    year={2025},
    url={https://openreview.net/forum?id=sSrOwve6vb}
}

@article{bricken-et-al-2023,
    title={{Towards Monosemanticity: Decomposing Language Models With Dictionary Learning}},
    author={Bricken, Trenton and Templeton, Adly and Batson, Joshua and Chen, Brian and Jermyn, Adam and Conerly, Tom and Turner, Nick and Anil, Cem and Denison, Carson and Röttger, Paul and others},
    journal={Transformer Circuits Thread},
    year={2023},
    month={oct},
    url={https://transformer-circuits.pub/2023/monosemantic-features/index.html}
}

@misc{cunningham-et-al-2023-sae,
    title={{Sparse Autoencoders Find Highly Interpretable Features in Language Models}}, 
    author={Hoagy Cunningham and Aidan Ewart and Logan Riggs and Robert Huben and Lee Sharkey},
    year={2023},
    eprint={2309.08600},
    archivePrefix={arXiv},
    primaryClass={cs.LG},
    url={https://arxiv.org/abs/2309.08600}, 
}

@inproceedings{rajamanoharan-et-al-2024-gated-sae,
    title={Improving Sparse Decomposition of Language Model Activations with Gated Sparse Autoencoders},
    author={Senthooran Rajamanoharan and Arthur Conmy and Lewis Smith and Tom Lieberum and Vikrant Varma and Janos Kramar and Rohin Shah and Neel Nanda},
    booktitle={The Thirty-eighth Annual Conference on Neural Information Processing Systems},
    year={2024},
    url={https://openreview.net/forum?id=zLBlin2zvW}
}

@inproceedings{gao-et-al-2025-topk-sae,
    title={{Scaling and Evaluating Sparse Autoencoders}},
    author={Leo Gao and Tom Dupre la Tour and Henk Tillman and Gabriel Goh and Rajan Troll and Alec Radford and Ilya Sutskever and Jan Leike and Jeffrey Wu},
    booktitle={The Thirteenth International Conference on Learning Representations},
    year={2025},
    url={https://openreview.net/forum?id=tcsZt9ZNKD}
}

@article{bussmann-et-al-2024-batchtopk-sae,
    title = {{BatchTopK Sparse Autoencoders}},
    author = {Bussmann, Bart and Leask, Patrick and Nanda, Neel},
    journal = {CoRR},
    volume = {abs/2412.06410},
    year = {2024},
    doi = {10.48550/arXiv.2412.06410},
    url = {https://arxiv.org/abs/2412.06410},
    eprint = {2412.06410},
    archivePrefix = {arXiv},
    primaryClass = {cs.LG}
}

@inproceedings{ghilardi-et-al-2025-group-sae,
    title = {{Group-SAE: Efficient Training of Sparse Autoencoders for Large Language Models via Layer Groups}},
    author = "Ghilardi, Davide  and
      Belotti, Federico  and
      Molinari, Marco  and
      Ma, Tao  and
      Palmonari, Matteo",
    editor = "Christodoulopoulos, Christos  and
      Chakraborty, Tanmoy  and
      Rose, Carolyn  and
      Peng, Violet",
    booktitle = "Proceedings of the 2025 Conference on Empirical Methods in Natural Language Processing",
    month = nov,
    year = "2025",
    address = "Suzhou, China",
    publisher = "Association for Computational Linguistics",
    url = "https://aclanthology.org/2025.emnlp-main.942/",
    doi = "10.18653/v1/2025.emnlp-main.942",
    pages = "18657--18677",
    ISBN = "979-8-89176-332-6"
}

@inproceedings{braun-et-al-2024-end-to-end-sae,
    author = {Braun, Dan and Taylor, Jordan and Goldowsky-Dill, Nicholas and Sharkey, Lee},
    booktitle = {Advances in Neural Information Processing Systems},
    doi = {10.52202/079017-3408},
    editor = {A. Globerson and L. Mackey and D. Belgrave and A. Fan and U. Paquet and J. Tomczak and C. Zhang},
    pages = {107286--107325},
    publisher = {Curran Associates, Inc.},
    title = {{Identifying Functionally Important Features with End-to-End Sparse Dictionary Learning}},
    url = {https://proceedings.neurips.cc/paper_files/paper/2024/file/c212c1b88395ab68d5e1671c17883ec6-Paper-Conference.pdf},
    volume = {37},
    year = {2024}
}

@inproceedings{chanin-et-al-2025-absorption,
    title = {{A is for Absorption: Studying Feature Splitting and Absorption in Sparse Autoencoders}},
    author = {Chanin, David and Wilken-Smith, James and Dulka, Tom{\'a}{\v{s}} and Bhatnagar, Hardik and Golechha, Satvik and Bloom, Joseph Isaac},
    booktitle = {Advances in Neural Information Processing Systems},
    year = {2025},
    url = {https://openreview.net/forum?id=R73ybUciQF}
}

@Article{pan-et-al-2023-tgdk,
    author =	{Pan, Jeff Z. and Razniewski, Simon and Kalo, Jan-Christoph and Singhania, Sneha and Chen, Jiaoyan and Dietze, Stefan and Jabeen, Hajira and Omeliyanenko, Janna and Zhang, Wen and Lissandrini, Matteo and Biswas, Russa and de Melo, Gerard and Bonifati, Angela and Vakaj, Edlira and Dragoni, Mauro and Graux, Damien},
    title =	{{Large Language Models and Knowledge Graphs: Opportunities and Challenges}},
    journal =	{Transactions on Graph Data and Knowledge},
    pages =	{2:1--2:38},
    year =	{2023},
    volume =	{1},
    number =	{1},
    publisher =	{Schloss Dagstuhl -- Leibniz-Zentrum f{\"u}r Informatik},
    address =	{Dagstuhl, Germany},
    URL =		{https://drops.dagstuhl.de/entities/document/10.4230/TGDK.1.1.2},
    URN =		{urn:nbn:de:0030-drops-194766},
    doi =		{10.4230/TGDK.1.1.2}
}

@inproceedings{petroni-et-al-2019-lama,
    title = "Language Models as Knowledge Bases?",
    author = {Petroni, Fabio  and
      Rockt{\"a}schel, Tim  and
      Riedel, Sebastian  and
      Lewis, Patrick  and
      Bakhtin, Anton  and
      Wu, Yuxiang  and
      Miller, Alexander},
    editor = "Inui, Kentaro  and
      Jiang, Jing  and
      Ng, Vincent  and
      Wan, Xiaojun",
    booktitle = "Proceedings of the 2019 Conference on Empirical Methods in Natural Language Processing and the 9th International Joint Conference on Natural Language Processing (EMNLP-IJCNLP)",
    month = nov,
    year = "2019",
    address = "Hong Kong, China",
    publisher = "Association for Computational Linguistics",
    url = "https://aclanthology.org/D19-1250/",
    doi = "10.18653/v1/D19-1250",
    pages = "2463--2473"
}

@inproceedings{giglou-et-al-2023-llms4ol,
author="Babaei Giglou, Hamed
and D'Souza, Jennifer
and Auer, S{\"o}ren",
editor="Payne, Terry R.
and Presutti, Valentina
and Qi, Guilin
and Poveda-Villal{\'o}n, Mar{\'i}a
and Stoilos, Giorgos
and Hollink, Laura
and Kaoudi, Zoi
and Cheng, Gong
and Li, Juanzi",
title="LLMs4OL: Large Language Models for Ontology Learning",
booktitle="The Semantic Web -- ISWC 2023",
year="2023",
publisher="Springer Nature Switzerland",
address="Cham",
pages="408--427",
isbn="978-3-031-47240-4"
}

@article{tsaneva-et-al-2025,
    title = {Knowledge graph validation by integrating LLMs and human-in-the-loop},
    journal = {Information Processing \& Management},
    volume = {62},
    number = {5},
    pages = {104145},
    year = {2025},
    issn = {0306-4573},
    doi = {https://doi.org/10.1016/j.ipm.2025.104145},
    url = {https://www.sciencedirect.com/science/article/pii/S030645732500086X},
    author = {Stefani Tsaneva and Danilo Dessì and Francesco Osborne and Marta Sabou},
}

@inproceedings{wu-et-al-2023-plm,
    title = {{Do PLMs Know and Understand Ontological Knowledge?}},
    author = "Wu, Weiqi  and
      Jiang, Chengyue  and
      Jiang, Yong  and
      Xie, Pengjun  and
      Tu, Kewei",
    editor = "Rogers, Anna  and
      Boyd-Graber, Jordan  and
      Okazaki, Naoaki",
    booktitle = "Proceedings of the 61st Annual Meeting of the Association for Computational Linguistics (Volume 1: Long Papers)",
    month = jul,
    year = "2023",
    address = "Toronto, Canada",
    publisher = "Association for Computational Linguistics",
    url = "https://aclanthology.org/2023.acl-long.173/",
    doi = "10.18653/v1/2023.acl-long.173",
    pages = "3080--3101",
}

@article{huang-et-al-2025-longtail,
    title = {Prompting large language models with knowledge graphs for question answering involving long-tail facts},
    journal = {Knowledge-Based Systems},
    volume = {324},
    pages = {113648},
    year = {2025},
    issn = {0950-7051},
    doi = {https://doi.org/10.1016/j.knosys.2025.113648},
    url = {https://www.sciencedirect.com/science/article/pii/S095070512500694X},
    author = {Wenyu Huang and Guancheng Zhou and Mirella Lapata and Pavlos Vougiouklis and Sebastien Montella and Jeff Z. Pan}
}

@article{he-et-al-2022-bertmap, 
    title={BERTMap: A BERT-Based Ontology Alignment System}, 
    volume={36}, 
    url={https://ojs.aaai.org/index.php/AAAI/article/view/20510}, 
    DOI={10.1609/aaai.v36i5.20510}, 
    abstractNote={Ontology alignment (a.k.a ontology matching (OM)) plays a critical role in knowledge integration. Owing to the success of machine learning in many domains, it has been applied in OM. However, the existing methods, which often adopt ad-hoc feature engineering or non-contextual word embeddings, have not yet outperformed rule-based systems especially in an unsupervised setting. In this paper, we propose a novel OM system named BERTMap which can support both unsupervised and semi-supervised settings. It first predicts mappings using a classifier based on fine-tuning the contextual embedding model BERT on text semantics corpora extracted from ontologies, and then refines the mappings through extension and repair by utilizing the ontology structure and logic. Our evaluation with three alignment tasks on biomedical ontologies demonstrates that BERTMap can often perform better than the leading OM systems LogMap and AML.}, 
    number={5}, 
    journal={Proceedings of the AAAI Conference on Artificial Intelligence}, 
    author={He, Yuan and Chen, Jiaoyan and Antonyrajah, Denvar and Horrocks, Ian}, 
    year={2022}, 
    month={Jun.}, 
    pages={5684-5691} 
}

@inproceedings{de-cao-et-al-2020-dbm,
    title = {{How do Decisions Emerge across Layers in Neural Models? Interpretation with Differentiable Masking}},
    author = "De Cao, Nicola  and
      Schlichtkrull, Michael Sejr  and
      Aziz, Wilker  and
      Titov, Ivan",
    editor = "Webber, Bonnie  and
      Cohn, Trevor  and
      He, Yulan  and
      Liu, Yang",
    booktitle = "Proceedings of the 2020 Conference on Empirical Methods in Natural Language Processing (EMNLP)",
    month = nov,
    year = "2020",
    address = "Online",
    publisher = "Association for Computational Linguistics",
    url = "https://aclanthology.org/2020.emnlp-main.262/",
    doi = "10.18653/v1/2020.emnlp-main.262",
    pages = "3243--3255",
}

@InProceedings{geiger-et-al-2024-das,
    title = 	 {{Finding Alignments Between Interpretable Causal Variables and Distributed Neural Representations}},
    author =       {Geiger, Atticus and Wu, Zhengxuan and Potts, Christopher and Icard, Thomas and Goodman, Noah},
    booktitle = 	 {Proceedings of the Third Conference on Causal Learning and Reasoning},
    pages = 	 {160--187},
    year = 	 {2024},
    editor = 	 {Locatello, Francesco and Didelez, Vanessa},
    volume = 	 {236},
    series = 	 {Proceedings of Machine Learning Research},
    month = 	 {01--03 Apr},
    publisher =    {PMLR},
    url = 	 {https://proceedings.mlr.press/v236/geiger24a.html},
}

@inproceedings{wu-et-al-2023-boundless-das,
    author = {Wu, Zhengxuan and Geiger, Atticus and Icard, Thomas and Potts, Christopher and Goodman, Noah D.},
    title = {Interpretability at scale: identifying causal mechanisms in alpaca},
    year = {2023},
    publisher = {Curran Associates Inc.},
    address = {Red Hook, NY, USA},
    booktitle = {Proceedings of the 37th International Conference on Neural Information Processing Systems},
    articleno = {3418},
    numpages = {22},
    location = {New Orleans, LA, USA},
    series = {NIPS '23}
}

@inproceedings{jacovi-goldberg-2020-faithfulness,
    title = {{Towards Faithfully Interpretable NLP Systems: How Should We Define and Evaluate Faithfulness?}},
    author = "Jacovi, Alon  and
      Goldberg, Yoav",
    editor = "Jurafsky, Dan  and
      Chai, Joyce  and
      Schluter, Natalie  and
      Tetreault, Joel",
    booktitle = "Proceedings of the 58th Annual Meeting of the Association for Computational Linguistics",
    month = jul,
    year = "2020",
    address = "Online",
    publisher = "Association for Computational Linguistics",
    url = "https://aclanthology.org/2020.acl-main.386/",
    doi = "10.18653/v1/2020.acl-main.386",
    pages = "4198--4205",
}

@inproceedings{wiegreffe-pinter-2019,
    title = {{Attention is not not Explanation}},
    author = "Wiegreffe, Sarah  and
      Pinter, Yuval",
    editor = "Inui, Kentaro  and
      Jiang, Jing  and
      Ng, Vincent  and
      Wan, Xiaojun",
    booktitle = "Proceedings of the 2019 Conference on Empirical Methods in Natural Language Processing and the 9th International Joint Conference on Natural Language Processing (EMNLP-IJCNLP)",
    month = nov,
    year = "2019",
    address = "Hong Kong, China",
    publisher = "Association for Computational Linguistics",
    url = "https://aclanthology.org/D19-1002/",
    doi = "10.18653/v1/D19-1002",
    pages = "11--20",
}

@misc{qwen3,
    title={{Qwen3 Technical Report}}, 
    author={An Yang and Anfeng Li and Baosong Yang and Beichen Zhang and Binyuan Hui and Bo Zheng and Bowen Yu and Chang Gao and Chengen Huang and Chenxu Lv and others},
    year={2025},
    eprint={2505.09388},
    archivePrefix={arXiv},
    primaryClass={cs.CL},
    url={https://arxiv.org/abs/2505.09388}, 
}

@inproceedings{lo-et-al-2024-end-to-end,
    author = {Lo, Andy and Jiang, Albert Q. and Li, Wenda and Jamnik, Mateja},
    booktitle = {Advances in Neural Information Processing Systems},
    doi = {10.52202/079017-2767},
    editor = {A. Globerson and L. Mackey and D. Belgrave and A. Fan and U. Paquet and J. Tomczak and C. Zhang},
    pages = {87184--87225},
    publisher = {Curran Associates, Inc.},
    title = {End-to-End Ontology Learning with Large Language Models},
    url = {https://proceedings.neurips.cc/paper_files/paper/2024/file/9e89f068a62f6858c661a8abecf5bb0a-Paper-Conference.pdf},
    volume = {37},
    year = {2024}
}

@InProceedings{zhang-et-al-2025-ontochat,
    author="Zhang, Bohui
    and Carriero, Valentina Anita
    and Schreiberhuber, Katrin
    and Tsaneva, Stefani
    and Gonz{\'a}lez, Luc{\'i}a S{\'a}nchez
    and Kim, Jongmo
    and de Berardinis, Jacopo",
    editor="Mero{\~{n}}o Pe{\~{n}}uela, Albert
    and Corcho, Oscar
    and Groth, Paul
    and Simperl, Elena
    and Tamma, Valentina
    and Nuzzolese, Andrea Giovanni
    and Poveda-Villal{\'o}n, Maria
    and Sabou, Marta
    and Presutti, Valentina
    and Celino, Irene
    and Revenko, Artem
    and Raad, Joe
    and Sartini, Bruno
    and Lisena, Pasquale",
    title={{OntoChat: A Framework for Conversational Ontology Engineering Using Language Models}},
    booktitle="The Semantic Web: ESWC 2024 Satellite Events",
    year="2025",
    publisher="Springer Nature Switzerland",
    address="Cham",
    pages="102--121",
    isbn="978-3-031-78952-6"
}

@article{geiger-et-al-2025-foundation,
    author  = {Atticus Geiger and Duligur Ibeling and Amir Zur and Maheep Chaudhary and Sonakshi Chauhan and Jing Huang and Aryaman Arora and Zhengxuan Wu and Noah Goodman and Christopher Potts and Thomas Icard},
    title   = {{Causal Abstraction: A Theoretical Foundation for Mechanistic Interpretability}},
    journal = {Journal of Machine Learning Research},
    year    = {2025},
    volume  = {26},
    number  = {83},
    pages   = {1--64},
    url     = {http://jmlr.org/papers/v26/23-0058.html}
}

@inproceedings{geiger-et-al-2021-causal,
    author = {Geiger, Atticus and Lu, Hanson and Icard, Thomas and Potts, Christopher},
    title = {{Causal Abstractions of Neural Networks}},
    year = {2021},
    isbn = {9781713845393},
    publisher = {Curran Associates Inc.},
    address = {Red Hook, NY, USA},
    booktitle = {Proceedings of the 35th International Conference on Neural Information Processing Systems},
    articleno = {733},
    numpages = {13},
    series = {NIPS '21}
}

@inproceedings{geiger-et-al-2022-iit,
    title = {{Inducing Causal Structure for Interpretable Neural Networks}},
    author = {Geiger, Atticus and Wu, Zhengxuan and Lu, Hanson and Rozner, Josh and Kreiss, Elisa and Icard, Thomas and Goodman, Noah and Potts, Christopher},
    booktitle = {Proceedings of the 39th International Conference on Machine Learning},
    pages = {7324--7338},
    year = {2022},
    editor = {Chaudhuri, Kamalika and Jegelka, Stefanie and Song, Le and Szepesvari, Csaba and Niu, Gang and Sabato, Sivan},
    volume = {162},
    series = {Proceedings of Machine Learning Research},
    month = {17--23 Jul},
    publisher = {PMLR},
    url = {https://proceedings.mlr.press/v162/geiger22a.html}
}

@inproceedings{fellbaum-2005-wordnet,
    address = {Oxford},
    author = {Fellbaum, Christiane},
    booktitle = {Encyclopedia of Language and Linguistics},
    description = {About WordNet - WordNet - About WordNet},
    edition = {Second Edition},
    editor = {Brown, Keith},
    pages = {665--670},
    publisher = {Elsevier},
    title = {WordNet and wordnets},
    url = {http://wordnet.princeton.edu/},
    year = 2005
}

@inproceedings{vrandecic-2012-wikidata,
    author = {Vrande\v{c}i\'{c}, Denny},
    title = {Wikidata: a new platform for collaborative data collection},
    year = {2012},
    isbn = {9781450312301},
    publisher = {Association for Computing Machinery},
    address = {New York, NY, USA},
    url = {https://doi.org/10.1145/2187980.2188242},
    doi = {10.1145/2187980.2188242},
    booktitle = {Proceedings of the 21st International Conference on World Wide Web},
    pages = {1063–1064},
    numpages = {2},
    location = {Lyon, France},
    series = {WWW '12 Companion}
}

@article{guha-et-al-2016-schemaorg,
    author = {Guha, R. V. and Brickley, Dan and Macbeth, Steve},
    title = {Schema.org: evolution of structured data on the web},
    year = {2016},
    issue_date = {February 2016},
    publisher = {Association for Computing Machinery},
    address = {New York, NY, USA},
    volume = {59},
    number = {2},
    issn = {0001-0782},
    url = {https://doi.org/10.1145/2844544},
    doi = {10.1145/2844544},
    journal = {Commun. ACM},
    month = jan,
    pages = {44–51},
    numpages = {8}
}

@inproceedings{he-et-al-2023-ontolama,
    title = "Language Model Analysis for Ontology Subsumption Inference",
    author = "He, Yuan  and
      Chen, Jiaoyan  and
      Jimenez-Ruiz, Ernesto  and
      Dong, Hang  and
      Horrocks, Ian",
    editor = "Rogers, Anna  and
      Boyd-Graber, Jordan  and
      Okazaki, Naoaki",
    booktitle = "Findings of the Association for Computational Linguistics: ACL 2023",
    month = jul,
    year = "2023",
    address = "Toronto, Canada",
    publisher = "Association for Computational Linguistics",
    url = "https://aclanthology.org/2023.findings-acl.213/",
    doi = "10.18653/v1/2023.findings-acl.213",
    pages = "3439--3453"
}

@inproceedings{zhang-et-al-2025-schema,
    title = "Schema Generation for Large Knowledge Graphs Using Large Language Models",
    author = "Zhang, Bohui  and
      He, Yuan  and
      Pintscher, Lydia  and
      Mero{\~n}o-Pe{\~n}uela, Albert  and
      Simperl, Elena",
    editor = "Christodoulopoulos, Christos  and
      Chakraborty, Tanmoy  and
      Rose, Carolyn  and
      Peng, Violet",
    booktitle = "Findings of the Association for Computational Linguistics: EMNLP 2025",
    month = nov,
    year = "2025",
    address = "Suzhou, China",
    publisher = "Association for Computational Linguistics",
    url = "https://aclanthology.org/2025.findings-emnlp.671/",
    doi = "10.18653/v1/2025.findings-emnlp.671",
    pages = "12561--12580",
    ISBN = "979-8-89176-335-7"
}

@misc{wikidump,
  author = {{Wikimedia Foundation}},
  title = {Wikimedia Downloads},
  year = {2022},
  howpublished = {\url{https://dumps.wikimedia.org}}
}

@inproceedings{meng-et-al-2022-rome,
    author = {Meng, Kevin and Bau, David and Andonian, Alex and Belinkov, Yonatan},
    title = {{Locating and editing factual associations in GPT}},
    year = {2022},
    isbn = {9781713871088},
    publisher = {Curran Associates Inc.},
    address = {Red Hook, NY, USA},
    booktitle = {Proceedings of the 36th International Conference on Neural Information Processing Systems},
    articleno = {1262},
    numpages = {14},
    location = {New Orleans, LA, USA},
    series = {NIPS '22}
}

@inproceedings{dai-et-al-2022-knowledge,
    title = {{Knowledge Neurons in Pretrained Transformers}},
    author = "Dai, Damai  and
      Dong, Li  and
      Hao, Yaru  and
      Sui, Zhifang  and
      Chang, Baobao  and
      Wei, Furu",
    editor = "Muresan, Smaranda  and
      Nakov, Preslav  and
      Villavicencio, Aline",
    booktitle = "Proceedings of the 60th Annual Meeting of the Association for Computational Linguistics (Volume 1: Long Papers)",
    month = may,
    year = "2022",
    address = "Dublin, Ireland",
    publisher = "Association for Computational Linguistics",
    url = "https://aclanthology.org/2022.acl-long.581/",
    doi = "10.18653/v1/2022.acl-long.581",
    pages = "8493--8502"
}

@inproceedings{geva-et-al-2022-transformer,
    title = "Transformer Feed-Forward Layers Build Predictions by Promoting Concepts in the Vocabulary Space",
    author = "Geva, Mor  and
      Caciularu, Avi  and
      Wang, Kevin  and
      Goldberg, Yoav",
    editor = "Goldberg, Yoav  and
      Kozareva, Zornitsa  and
      Zhang, Yue",
    booktitle = "Proceedings of the 2022 Conference on Empirical Methods in Natural Language Processing",
    month = dec,
    year = "2022",
    address = "Abu Dhabi, United Arab Emirates",
    publisher = "Association for Computational Linguistics",
    url = "https://aclanthology.org/2022.emnlp-main.3/",
    doi = "10.18653/v1/2022.emnlp-main.3",
    pages = "30--45"
}

@inproceedings{sun-et-al-2024-head-to-tail,
    title = {{Head-to-Tail: How Knowledgeable are Large Language Models (LLMs)? A.K.A. Will LLMs Replace Knowledge Graphs?}},
    author = "Sun, Kai  and
      Xu, Yifan  and
      Zha, Hanwen  and
      Liu, Yue  and
      Dong, Xin Luna",
    editor = "Duh, Kevin  and
      Gomez, Helena  and
      Bethard, Steven",
    booktitle = "Proceedings of the 2024 Conference of the North American Chapter of the Association for Computational Linguistics: Human Language Technologies (Volume 1: Long Papers)",
    month = jun,
    year = "2024",
    address = "Mexico City, Mexico",
    publisher = "Association for Computational Linguistics",
    url = "https://aclanthology.org/2024.naacl-long.18/",
    doi = "10.18653/v1/2024.naacl-long.18",
    pages = "311--325"
}

\newpage






\appendix

\section{Dataset Details}\label{appendix:dataset}

\subsection{Dataset construction}\label{app:dataset-construction}
Our dataset construction follows two design goals. First, the selected subjects should cover a broad range of ontology domains and popularity levels, following the head-to-tail motivation in factual-knowledge evaluation~\cite{sun-et-al-2024-head-to-tail}. Second, the selected subjects should not be so obscure that the underlying LLM fails on most target-relation or auxiliary-attribute queries, since the final benchmark retains only subject-template pairs for which the unintervened model predicts an accepted answer. We therefore sample a set of domains and select common but diverse ontology classes within each domain. For the \texttt{instance of} subset, we sample entities from each selected class by considering both their popularity level and the amount of structured information available on Wikidata, using the number of associated Wikidata statements as a practical proxy. For the \texttt{subclass of} subset, we similarly select parent classes whose subclasses cover a range of popularity levels, using the number of subclasses as a proxy. This sampling strategy aims to maximize ontology coverage while avoiding overly rare subjects that would be mostly removed by behavioral filtering. We retain at least ten candidate entities or subclasses for each selected class before filtering. To reduce ambiguity in factual evaluation, we avoid highly time-sensitive subjects and attributes whose values are expected to change frequently. Ground-truth attribute values are queried from Wikidata using SPARQL through the Wikidata Query Service. For attributes with corresponding values available in Wikipedia infoboxes, we cross-validate the Wikidata values against those infobox values. Conflicting values, missing values, and cases without aligned infobox evidence are manually checked against reliable web references.

The class-specific attributes are defined at different levels for these two subsets. The first is at the instance level, where attribute values are specified for each instance. The second is at the class level, where attributes are more abstract, such as the general study object for an academic discipline.

\paragraph{Data appendix and release.}
The anonymous code-and-data appendix contains the class CSV files, attribute-value dictionaries, prompt-template files, Wikipedia-derived context prompts, and construction scripts needed to reconstruct the benchmark instances. Upon publication, the code and data will be released under a research-permissive license.

\section{Implementation Details}

The single-site SAE feature-selection pipeline and the DBM/MDBM and DAS/MDAS
intervention implementations are adapted from the released RAVEL
codebase~\cite{huang-et-al-2024-ravel}.\footnote{\url{https://github.com/explanare/ravel}}
We modify them for ontology-specific prompts, labels, and splits, and add the
multi-layer DAS/MDAS extension described below. 
All intervention training and evaluation jobs used a single NVIDIA A100 GPU. The submitted code appendix specifies the software environment.
Table~\ref{tab:experimental-settings} lists model-level settings, input lengths, batch sizes, seeds, and realized training-pair counts.

\subsection{Method-specific Implementation}

\paragraph{SAE.}
For each site, we train a single-layer ReLU encoder with a linear decoder on
unlabeled target-relation training activations ending at the subject position.
We evaluate latent expansion ratios $d_{\mathrm{lat}}/d\in\{2,4\}$, giving
5,120/10,240 latent dimensions for Qwen3-4B and 8,192/16,384 for Qwen3-8B. The
autoencoder uses mean-squared reconstruction error plus mean absolute latent
activation with coefficient $\beta=0.5$, optimized with AdamW at initial
learning rate $10^{-4}$, weight decay $10^{-4}$, and cosine decay to $0.5$ of
the initial rate. 
We then fit an $\ell_1$-regularized linear SVM with $C\in\{0.1,1,10,1000\}$ on latent codes to instantiate $F_\tau$ as a selected latent feature set. During intervention, only these selected latent dimensions are replaced with source values before decoding.

\paragraph{DBM and MDBM.}
DBM and MDBM learn one mask logit per residual-stream coordinate, giving
ambient mask widths of 2,560 for Qwen3-4B and 4,096 for Qwen3-8B. The logits
are initialized to zero, the temperature is linearly annealed from $10^{-2}$ to
$10^{-7}$, and optimization uses AdamW with constant learning rate $10^{-4}$
and zero weight decay. DBM uses cause-only answer-token supervision with an
$\ell_1$ coefficient $\lambda=10^{-3}$ on the relaxed mask, while MDBM mixes
Cause examples with auxiliary examples and uses $\lambda=0$. Evaluation
thresholds the relaxed mask at $0.5$ and exchanges only the selected
original-basis coordinates. Thus, the \emph{Dimension} column in the main paper denotes
ambient mask width rather than effective feature size. The effective DBM/MDBM
sizes are 1,631/1,684 and 2,211/2,580 for Qwen3-4B and Qwen3-8B on
\texttt{instance of}, and 758/1,467 and 1,586/2,043 on \texttt{subclass of}.

\paragraph{DAS and MDAS.}
DAS and MDAS learn a row-orthonormal map $W\in\mathbb{R}^{k\times d}$,
implemented as a bias-free linear map with PyTorch orthogonal parametrization
to maintain $WW^\top=I_k$. Single-layer experiments use
$k\in\{1024,2048\}$, while multi-layer ranks are specified below. The language
model is frozen, and only the subspace parameters are optimized with AdamW at
constant learning rate $10^{-4}$, zero weight decay, and no additional
regularization. Training uses autoregressive cross-entropy on answer tokens:
DAS uses Cause examples whose target is the accepted answer corresponding to
$c_\tau(s_{\mathrm{source}})$, whereas MDAS mixes these Cause examples with
auxiliary examples targeted to $v_A(s_{\mathrm{base}})$.

\subsection{Multi-layer Realization}

Let $S=(\ell_1,\ldots,\ell_m)$ be an intervention window with
$\ell_1<\cdots<\ell_m$, and let $z_\ell(x)$ denote the block-output
residual-stream activation at the subject position in layer $\ell$.  We first
run one clean source forward pass and cache
\[
s_j=z_{\ell_j}(x_{\mathrm{source}}),
\qquad j=1,\ldots,m.
\]
No intervention is applied during this pass, so every $s_j$ is a clean source
activation rather than the output of an earlier source-side intervention.  We
then run the base input once.  Writing
$b_1=z_{\ell_1}(x_{\mathrm{base}})$ and letting
$\Phi_{\ell_{j-1}\rightarrow\ell_j}$ denote the frozen Transformer
computation between two intervention sites, the base execution follows
\[
\begin{aligned}
b_j
&=
\Phi_{\ell_{j-1}\rightarrow\ell_j}(\tilde b_{j-1}),
\\
\tilde b_j
&=
\Gamma_{\vartheta_j}^{(\ell_j)}(b_j,s_j),
\qquad j=2,\ldots,m .
\end{aligned}
\]
with $\tilde b_1=\Gamma_{\vartheta_1}^{(\ell_1)}(b_1,s_1)$.  Thus, the hooks are invoked
in ascending layer order: the base activation encountered at $\ell_j$
already reflects all interventions at lower layers.  The layer-specific
parameters $\{\vartheta_j\}_{j=1}^m$ are optimized jointly under the same
autoregressive objective.

We evaluate contiguous two- and four-layer windows.  For DAS and MDAS, each
layer has an independent rank-$k$ orthogonal map, so the total intervention
dimension is $mk$.  The two-layer sweep uses $k=512$ (1,024 dimensions in
total), while the four-layer sweep uses $k\in\{256,512\}$ (1,024 or 2,048
dimensions in total).  All reported multi-layer results use block-output
interventions at the subject position.


\begin{table*}[t]
\centering
\small
\setlength{\tabcolsep}{5pt}
\renewcommand{\arraystretch}{1.08}
\begin{tabular}{lll}
\toprule
\textbf{Setting} & \textbf{Qwen3-4B} & \textbf{Qwen3-8B} \\
\midrule
Layers / hidden size & 36 / 2,560 & 36 / 4,096 \\
Prompt / answer length & 48 / at most 3 tokens & 48 / at most 3 tokens \\
Site / position & block output / last subject subtoken & block output / last subject subtoken \\
Supervised train / eval batch & 128 / 128 & 128 / 128 \\
Epochs & 6 & 6 \\
Random seed & 42 & 42 \\
\texttt{instance of} Cause / Iso pairs & 54,184 / 68,936 & 55,888 / 70,872 \\
\texttt{subclass of} Cause / Iso pairs & 44,864 / 21,448 & 43,552 / 31,720 \\
\bottomrule
\end{tabular}
\caption{Experimental settings. Training pair counts are the examples passed to the supervised methods after dataset construction.}
\label{tab:experimental-settings}
\end{table*}

%
%

\section{Additional Results}

\subsection{LLM Performance on Ontology Engineering Tasks}
\label{appendix:llm-performance}

Table~\ref{table:llm-performance} summarizes the LLMs' micro- and class-macro accuracy on target axiom prediction and auxiliary attribute prediction under each axiom type.

\begin{table*}[t]
\centering
\small
\setlength{\tabcolsep}{4pt}
\renewcommand{\arraystretch}{1.05}
\begin{tabular}{lcccc cccc}
\toprule
& \multicolumn{4}{c}{\textbf{\texttt{instance of}}} & \multicolumn{4}{c}{\textbf{\texttt{subclass of}}} \\
\cmidrule(lr){2-5} \cmidrule(lr){6-9}
\textbf{Model} & \textbf{\makecell{Target\\Micro}} & \textbf{\makecell{Target\\Macro}} & \textbf{\makecell{Attr.\\Micro}} & \textbf{\makecell{Attr.\\Macro}} & \textbf{\makecell{Target\\Micro}} & \textbf{\makecell{Target\\Macro}} & \textbf{\makecell{Attr.\\Micro}} & \textbf{\makecell{Attr.\\Macro}} \\
\midrule
Qwen3-4B & 92.4 & 92.1 & 59.7 & 58.7 & 89.2 & 87.4 & 69.2 & 68.8 \\
Qwen3-8B & 94.7 & 94.3 & 64.9 & 63.8 & 93.0 & 91.8 & 69.1 & 68.3 \\
\bottomrule
\end{tabular}
\caption{Base LLM performance on ontology-oriented knowledge engineering queries. \textbf{Target Micro/Macro} denotes micro- and class-macro accuracy on target axiom queries; \textbf{Attr. Micro/Macro} reports the corresponding accuracy on auxiliary attribute queries.}
\label{table:llm-performance}
\end{table*}

\subsection{Class-Level Intervention Analysis}
\label{appendix:class-level-analysis}

\begin{figure*}[t]
  \centering
  \captionsetup[subfigure]{font=small,skip=2pt}
  \begin{subfigure}[t]{0.48\textwidth}
    \centering
    \includegraphics[width=\linewidth]{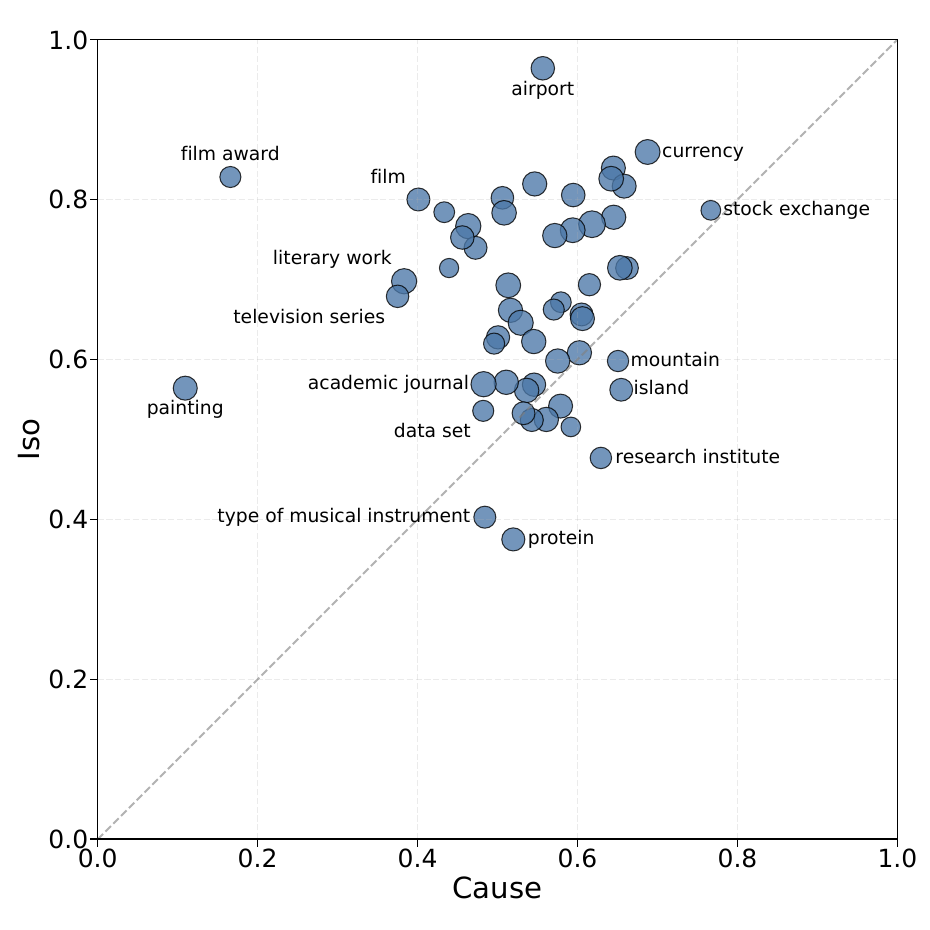}
    \caption{Qwen3-8B base}
  \end{subfigure}
  \hfill
  \begin{subfigure}[t]{0.48\textwidth}
    \centering
    \includegraphics[width=\linewidth]{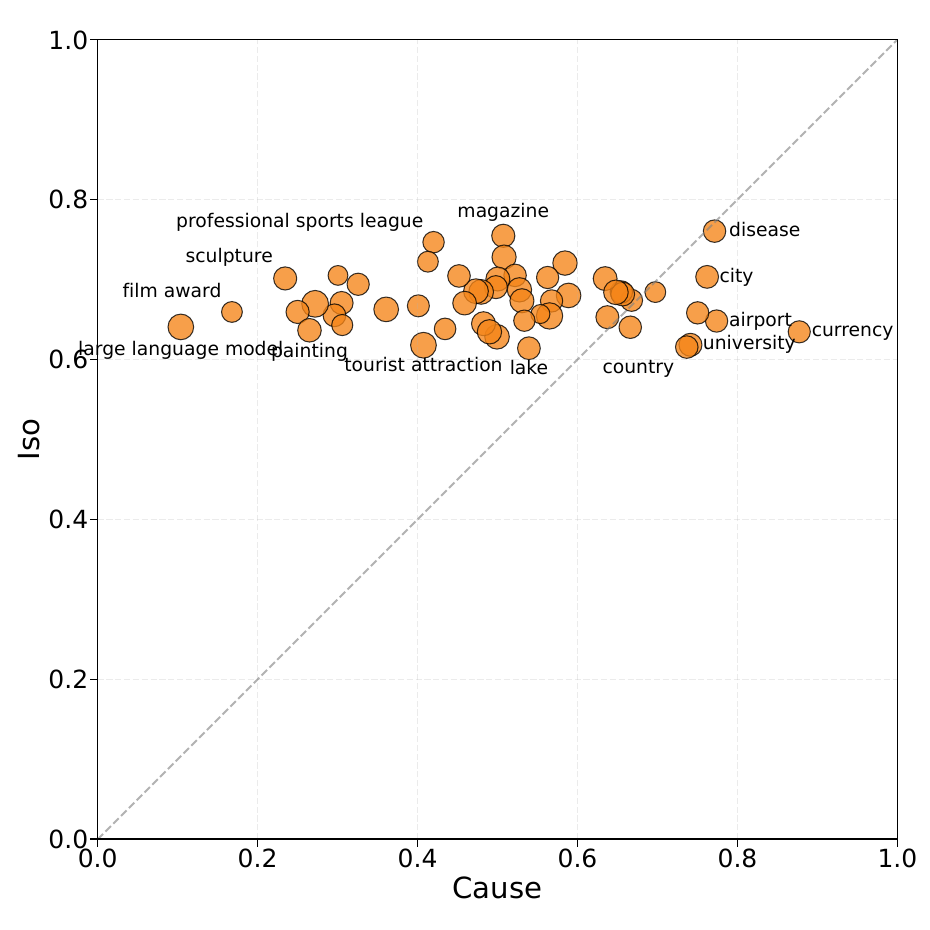}
    \caption{Qwen3-8B source}
  \end{subfigure}

  \vspace{0.1em}
  \begin{subfigure}[t]{0.48\textwidth}
    \centering
    \includegraphics[width=\linewidth]{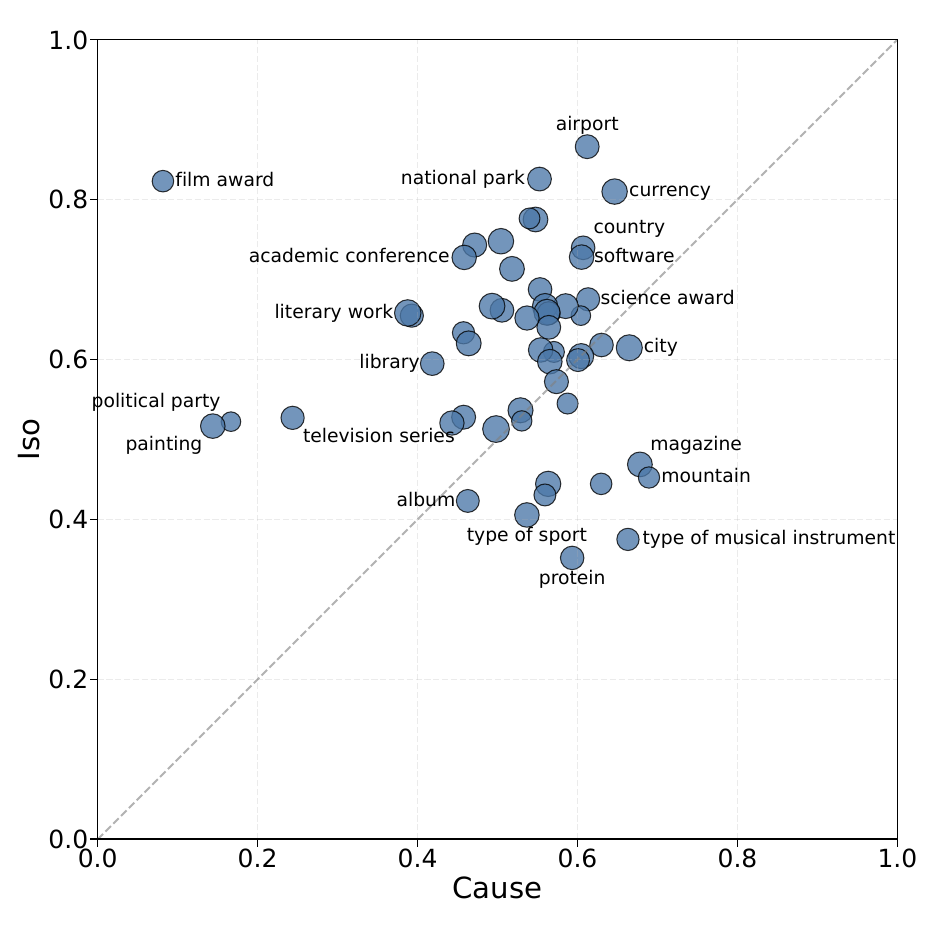}
    \caption{Qwen3-4B base}
  \end{subfigure}
  \hfill
  \begin{subfigure}[t]{0.48\textwidth}
    \centering
    \includegraphics[width=\linewidth]{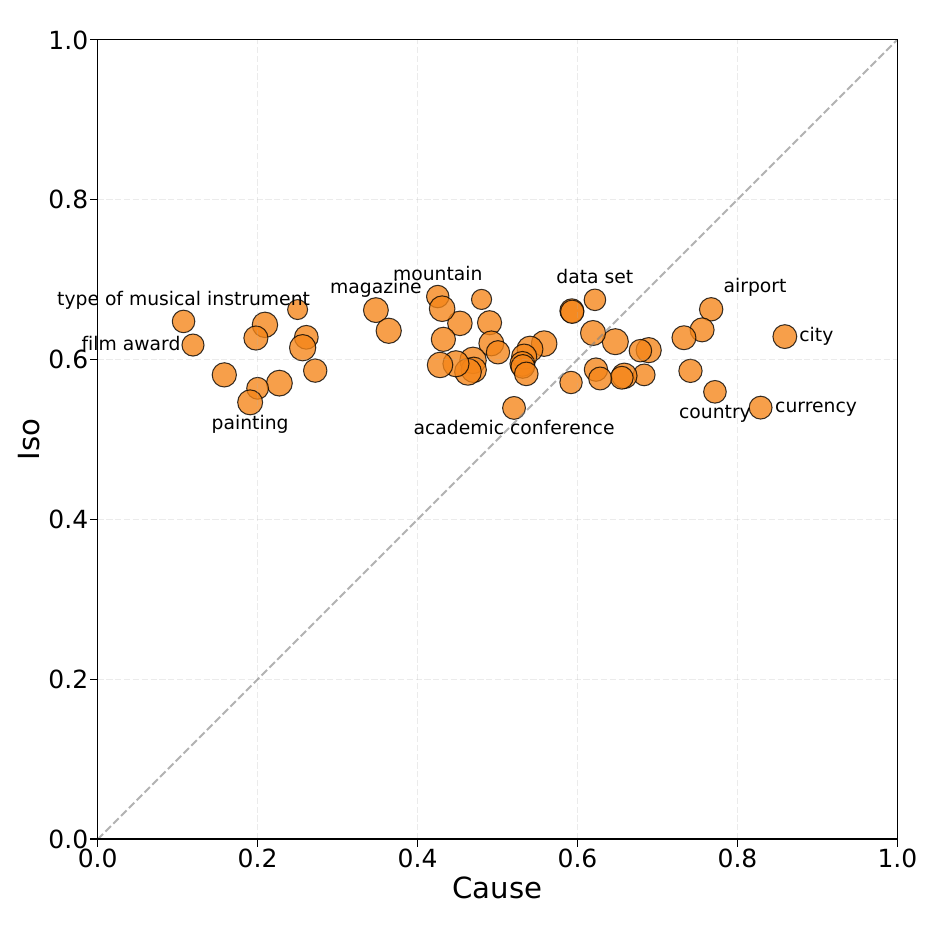}
    \caption{Qwen3-4B source}
  \end{subfigure}

  \caption{Class-level Cause--Iso scatter plots for the best MDAS instance-of settings. Each row shows base classes on the left and source classes on the right.}
  \label{fig:appendix-class-level-instance-scatter}
\end{figure*}

\begin{figure*}[t]
  \centering
  \captionsetup[subfigure]{font=small,skip=2pt}
  \begin{subfigure}[t]{0.48\textwidth}
    \centering
    \includegraphics[width=\linewidth]{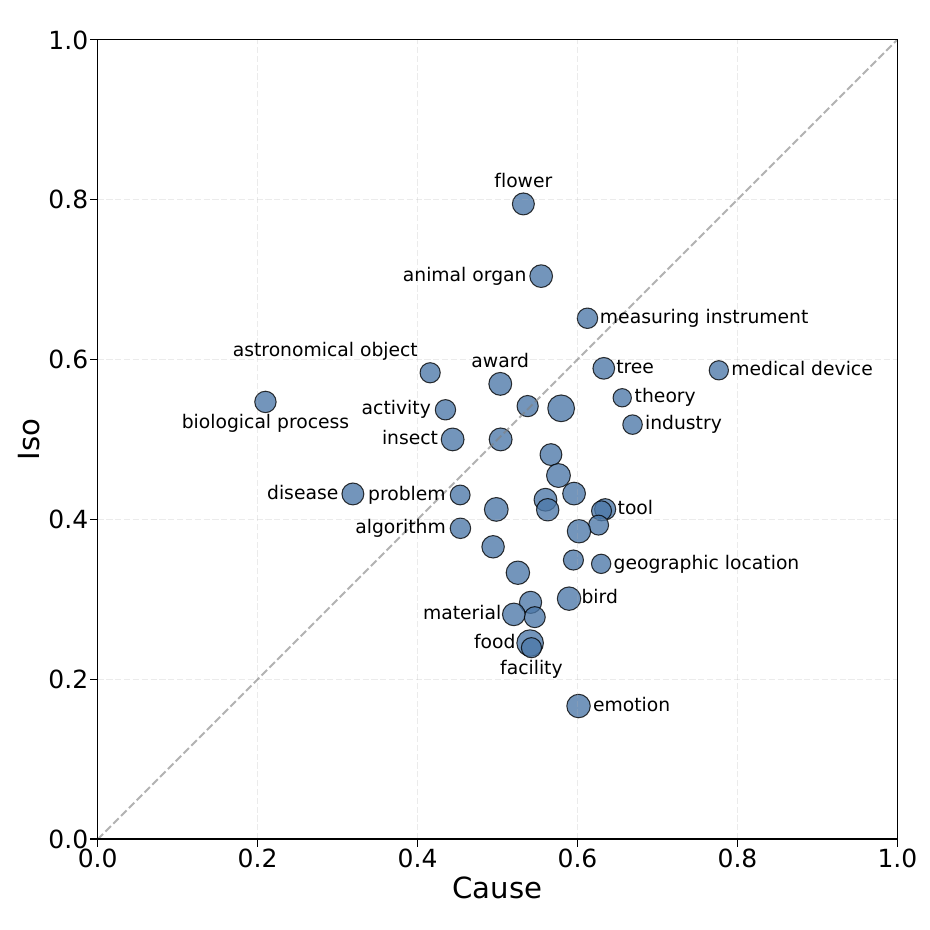}
    \caption{Qwen3-8B base}
  \end{subfigure}
  \hfill
  \begin{subfigure}[t]{0.48\textwidth}
    \centering
    \includegraphics[width=\linewidth]{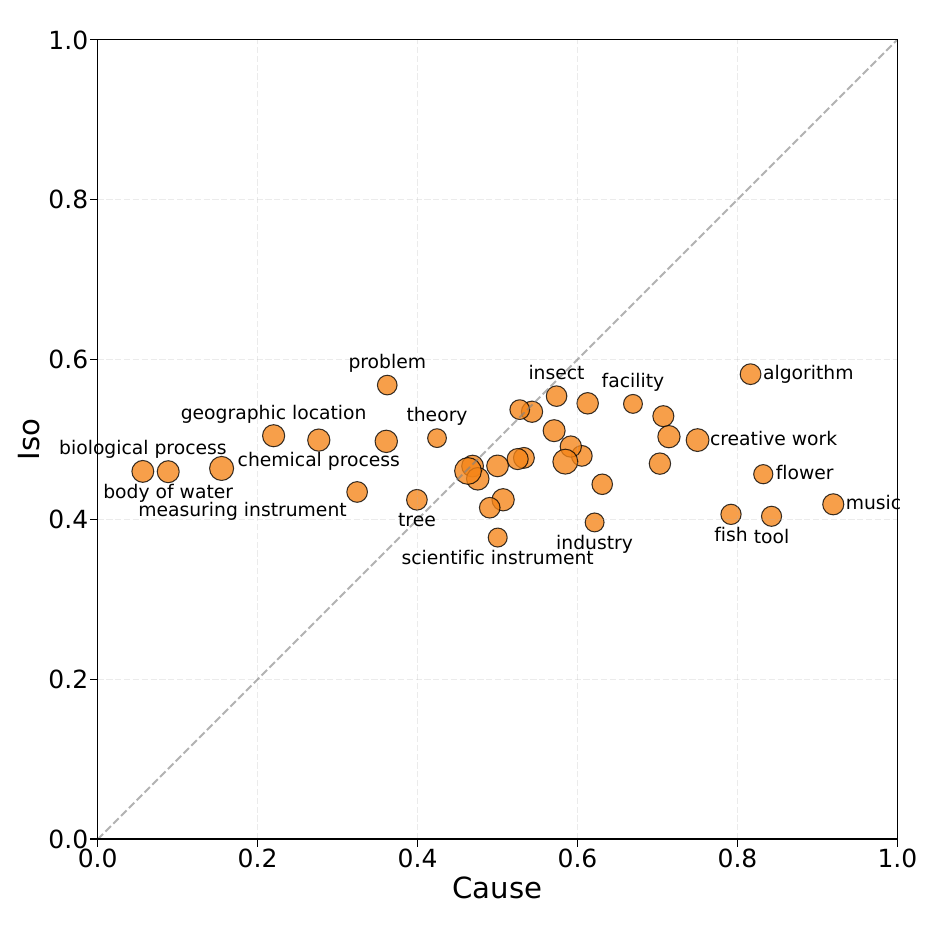}
    \caption{Qwen3-8B source}
  \end{subfigure}

  \vspace{0.1em}
  \begin{subfigure}[t]{0.48\textwidth}
    \centering
    \includegraphics[width=\linewidth]{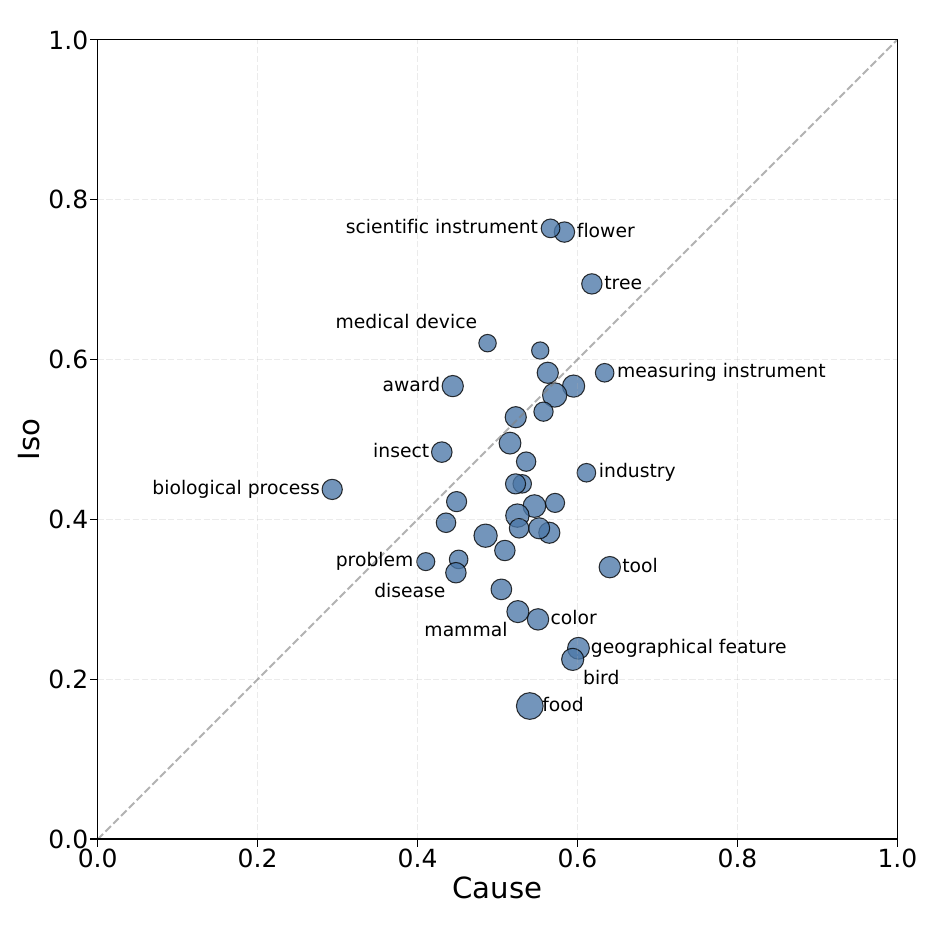}
    \caption{Qwen3-4B base}
  \end{subfigure}
  \hfill
  \begin{subfigure}[t]{0.48\textwidth}
    \centering
    \includegraphics[width=\linewidth]{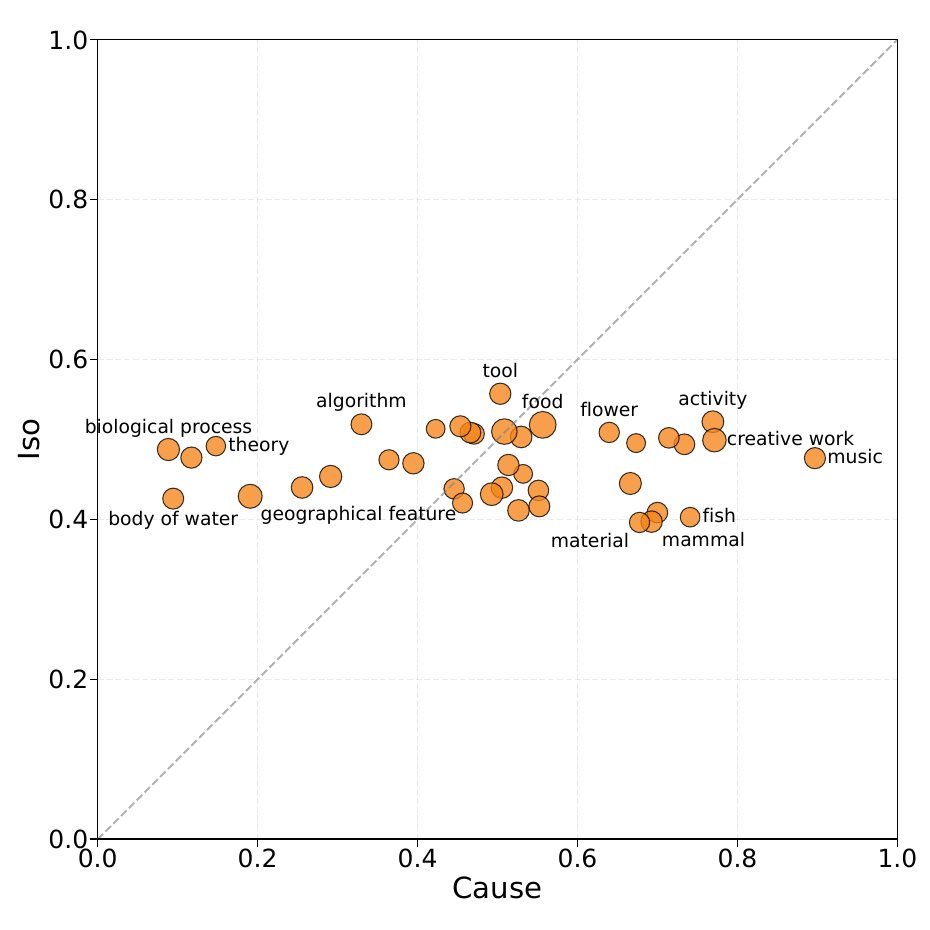}
    \caption{Qwen3-4B source}
  \end{subfigure}

  \caption{Class-level Cause--Iso scatter plots for the best MDAS subclass-of settings. Each row shows base classes on the left and source classes on the right.}
  \label{fig:appendix-class-level-subclass-scatter}
\end{figure*}

\begin{figure*}[t]
  \centering
  \captionsetup[subfigure]{font=small,skip=2pt}
  \begin{subfigure}[t]{0.48\textwidth}
    \centering
    \includegraphics[width=\linewidth]{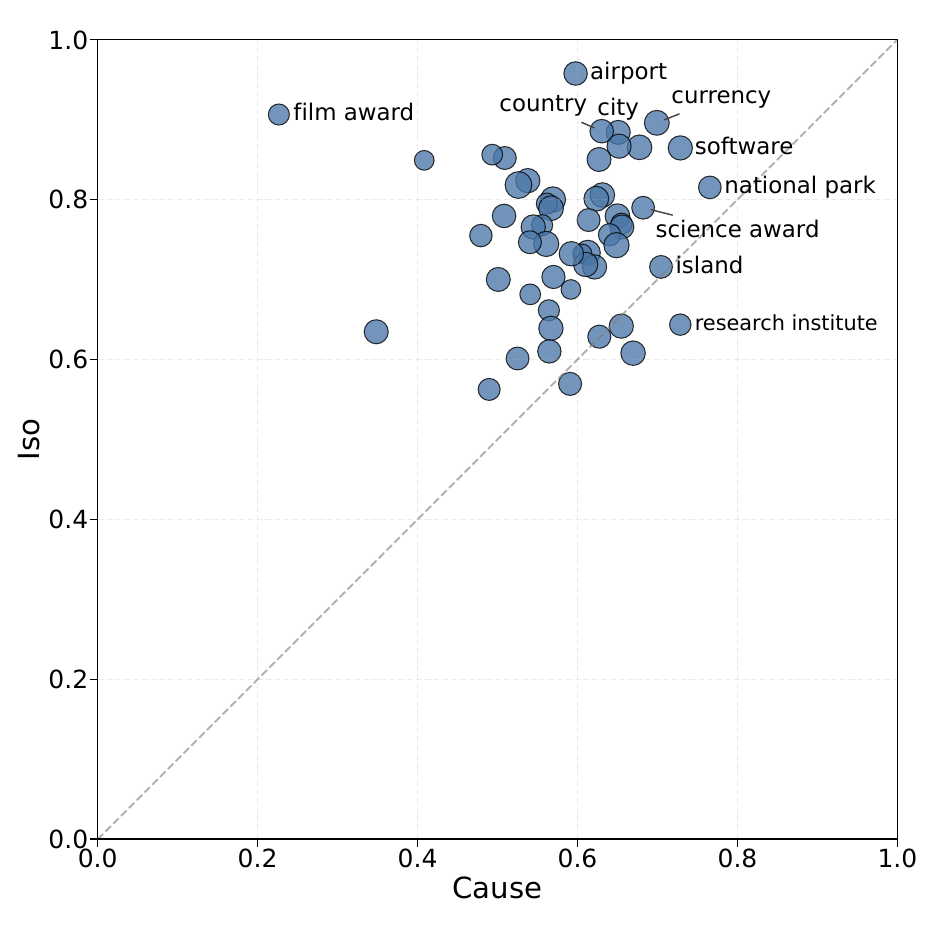}
    \caption{\texttt{instance of}, base}
  \end{subfigure}
  \hfill
  \begin{subfigure}[t]{0.48\textwidth}
    \centering
    \includegraphics[width=\linewidth]{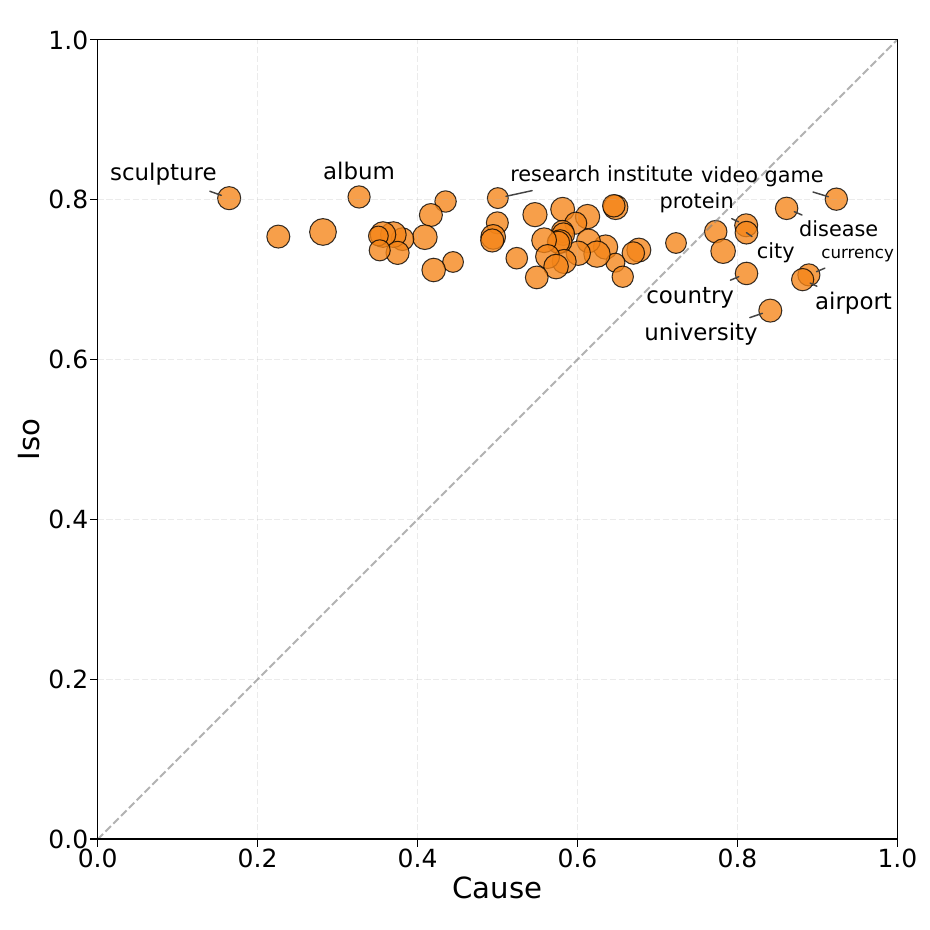}
    \caption{\texttt{instance of}, source}
  \end{subfigure}

  \vspace{0.1em}
  \begin{subfigure}[t]{0.48\textwidth}
    \centering
    \includegraphics[width=\linewidth]{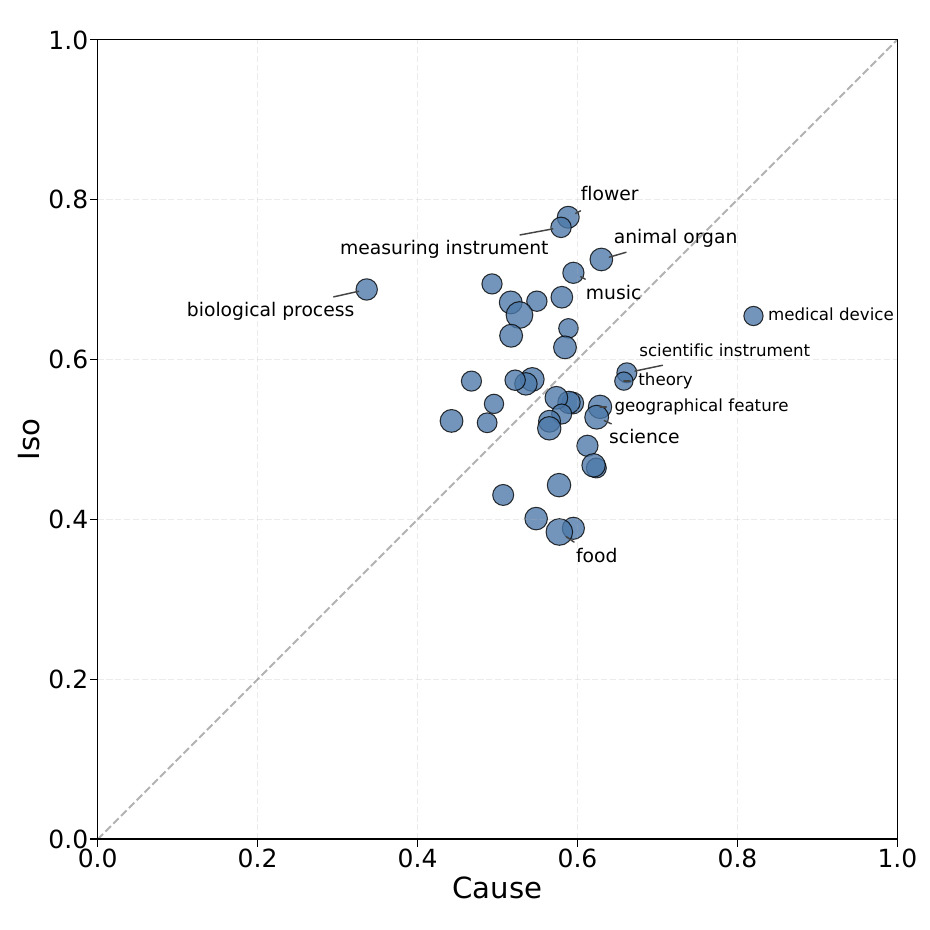}
    \caption{\texttt{subclass of}, base}
  \end{subfigure}
  \hfill
  \begin{subfigure}[t]{0.48\textwidth}
    \centering
    \includegraphics[width=\linewidth]{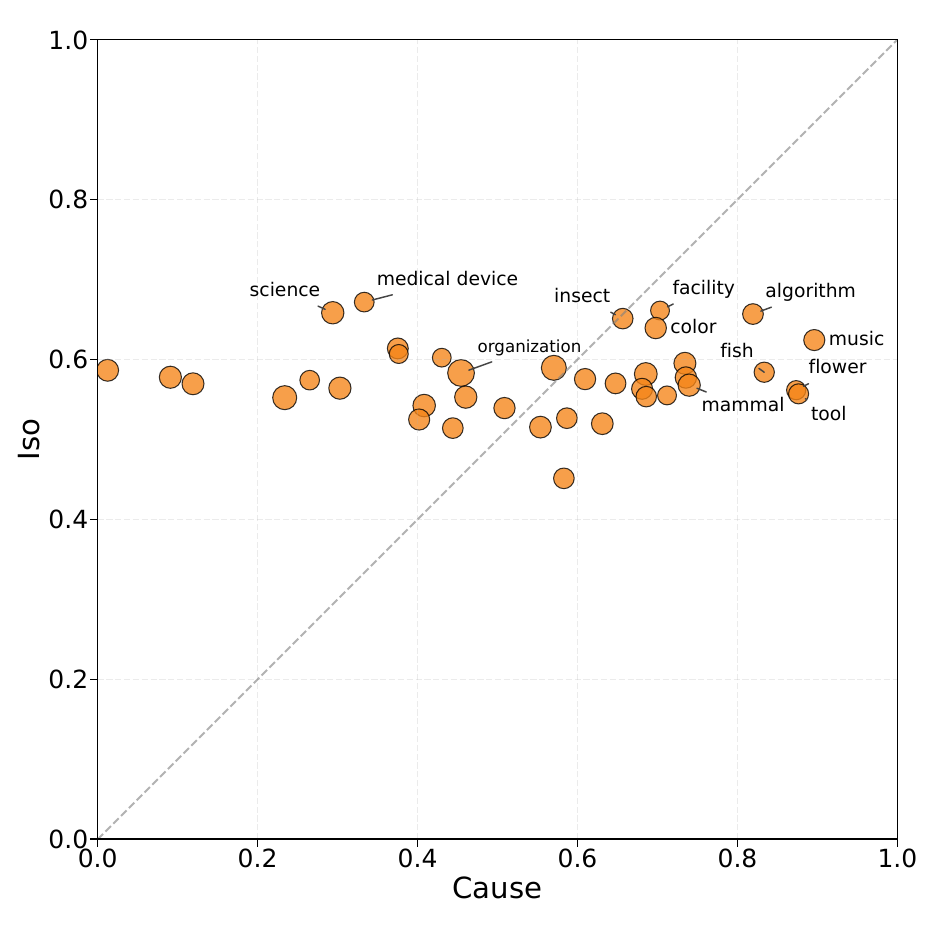}
    \caption{\texttt{subclass of}, source}
  \end{subfigure}

  \caption{Class-level Cause--Iso scatter plots for the best Qwen3-8B multi-layer MDAS settings. Each row corresponds to one axiom type, with base classes on the left and source classes on the right.}
  \label{fig:appendix-class-level-parallel-scatter}
\end{figure*}

Figures~\ref{fig:appendix-class-level-instance-scatter} and~\ref{fig:appendix-class-level-subclass-scatter} break down the four MDAS class-evaluation runs by ontology class. We use the same role terminology as in the interchange-intervention formulation: the base class is the class of the subject whose computation receives the intervention, whereas the source class is the class of the subject that supplies the intervened representation. In the base-role panels, each point aggregates examples whose base subject belongs to that class. In the source-role panels, each point aggregates examples whose source subject belongs to that class. All scores use the Entity split and task-macro aggregation: each prompt-specific evaluation condition is scored first, and the resulting condition-level scores are then averaged with equal weight.

\paragraph{Class-level heterogeneity is systematic.}
The scatter plots show substantial class-level heterogeneity behind the aggregate scores reported in the paper. Table~\ref{tab:class-level-role-summary} reports Cause, Iso, and Disentangle after first computing scores for each ontology class and then averaging over classes. For \texttt{instance of}, Iso is consistently higher than Cause under this class-level averaging. This means that the main bottleneck is not preserving auxiliary attributes after intervention, but changing the target ontology answer toward the source subject's answer. For \texttt{subclass of}, the pattern is different: Cause is comparable to or higher than Iso, especially for classes in the base role. This supports the paper's aggregate finding that subsumption prediction is more entangled with surrounding semantic and attribute information than entity typing. 

\begin{table}[t]
\centering
\small
\setlength{\tabcolsep}{3pt}
\renewcommand{\arraystretch}{1.05}
\begin{tabular}{llcccc}
\toprule
\textbf{Model} & \textbf{Role} & \textbf{Cause} & \textbf{Iso} & \textbf{Disent.} & $\boldsymbol{\Delta}$ \\
\midrule
\rowcolor{black!8}\multicolumn{6}{c}{\textbf{\texttt{instance of}}} \\
\midrule
Qwen3-8B & Base & 0.539 & 0.674 & 0.606 & +0.135 \\
Qwen3-8B & Source   & 0.507 & 0.674 & 0.590 & +0.167 \\
Qwen3-4B & Base & 0.521 & 0.613 & 0.567 & +0.091 \\
Qwen3-4B & Source   & 0.493 & 0.612 & 0.552 & +0.119 \\
\midrule
\rowcolor{black!8}\multicolumn{6}{c}{\textbf{\texttt{subclass of}}} \\
\midrule
Qwen3-8B & Base & 0.545 & 0.446 & 0.496 & -0.100 \\
Qwen3-8B & Source   & 0.532 & 0.477 & 0.504 & -0.055 \\
Qwen3-4B & Base & 0.527 & 0.447 & 0.487 & -0.080 \\
Qwen3-4B & Source   & 0.498 & 0.468 & 0.483 & -0.030 \\
\bottomrule
\end{tabular}
\caption{Mean scores over ontology classes in the base and source roles. $\Delta$ reports $\mathrm{Iso}-\mathrm{Cause}$, with positive values indicating that changing the target answer is harder than preserving auxiliary answers and negative values indicating the reverse.}
\label{tab:class-level-role-summary}
\end{table}

\paragraph{The class ranking is stable across model scale.}
For each axiom type and role, we rank ontology classes by their class-level Disentangle scores. Table~\ref{tab:class-level-rank-stability} reports Spearman rank correlations over these class rankings. The first two columns compare Qwen3-8B and Qwen3-4B within the same role, asking whether the same classes are easy or difficult across model sizes. The last two columns compare base-role and source-role rankings within the same model, asking whether classes that work well as intervention bases also work well as intervention sources. The high cross-model correlations show that class-level variation is not merely model-specific noise. The much lower base--source correlations show that class difficulty is role-dependent.

\begin{table*}[t]
\centering
\small
\setlength{\tabcolsep}{2pt}
\renewcommand{\arraystretch}{1.05}
\begin{tabular}{lcccc}
\toprule
\textbf{Axiom} &
\textbf{\makecell{Base\\cross-model $\rho$}} &
\textbf{\makecell{Source\\cross-model $\rho$}} &
\textbf{\makecell{8B \\base--source $\rho$}} &
\textbf{\makecell{4B \\base--source $\rho$}} \\
\midrule
\texttt{instance of} & 0.831 & 0.810 & 0.273 & 0.267 \\
\texttt{subclass of} & 0.752 & 0.671 & -0.082 & 0.059 \\
\bottomrule
\end{tabular}
\caption{Rank stability of class-level Disentangle scores. Each value is Spearman's $\rho$ over ontology classes. Base and Source cross-model $\rho$ compare Qwen3-8B and Qwen3-4B class rankings within the same role. The 8B and 4B base--source $\rho$ columns compare base-role and source-role rankings within the same model.}
\label{tab:class-level-rank-stability}
\end{table*}

\begin{figure*}[t]
  \centering
  \captionsetup[subfigure]{font=small,skip=2pt}
  \begin{subfigure}[t]{0.48\textwidth}
    \centering
    \includegraphics[width=\linewidth]{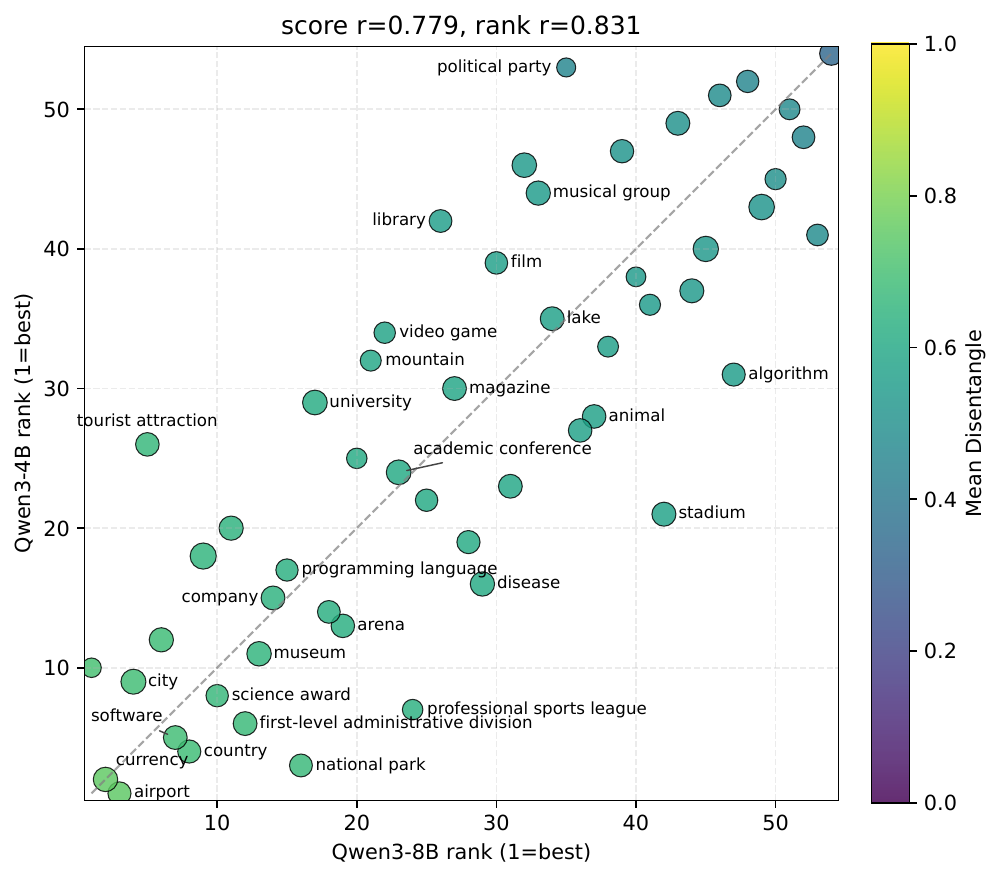}
    \caption{\texttt{instance of}, base}
  \end{subfigure}
  \hfill
  \begin{subfigure}[t]{0.48\textwidth}
    \centering
    \includegraphics[width=\linewidth]{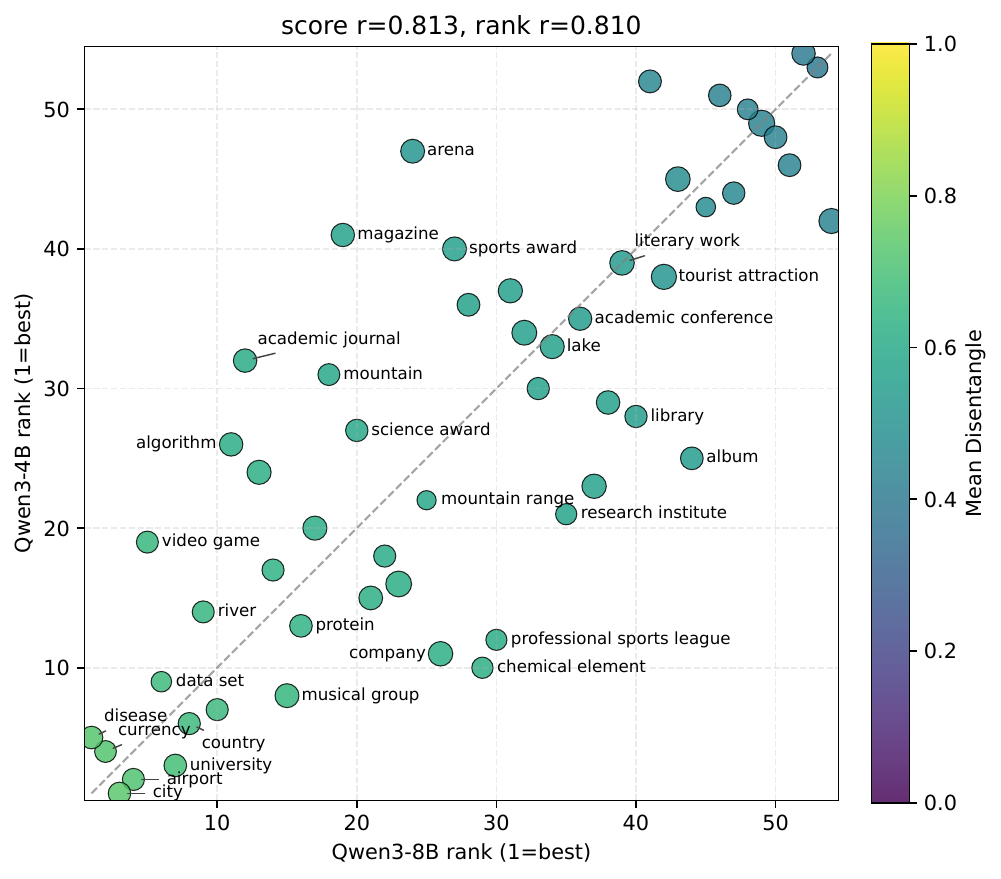}
    \caption{\texttt{instance of}, source}
  \end{subfigure}

  \vspace{0.2em}
  \begin{subfigure}[t]{0.48\textwidth}
    \centering
    \includegraphics[width=\linewidth]{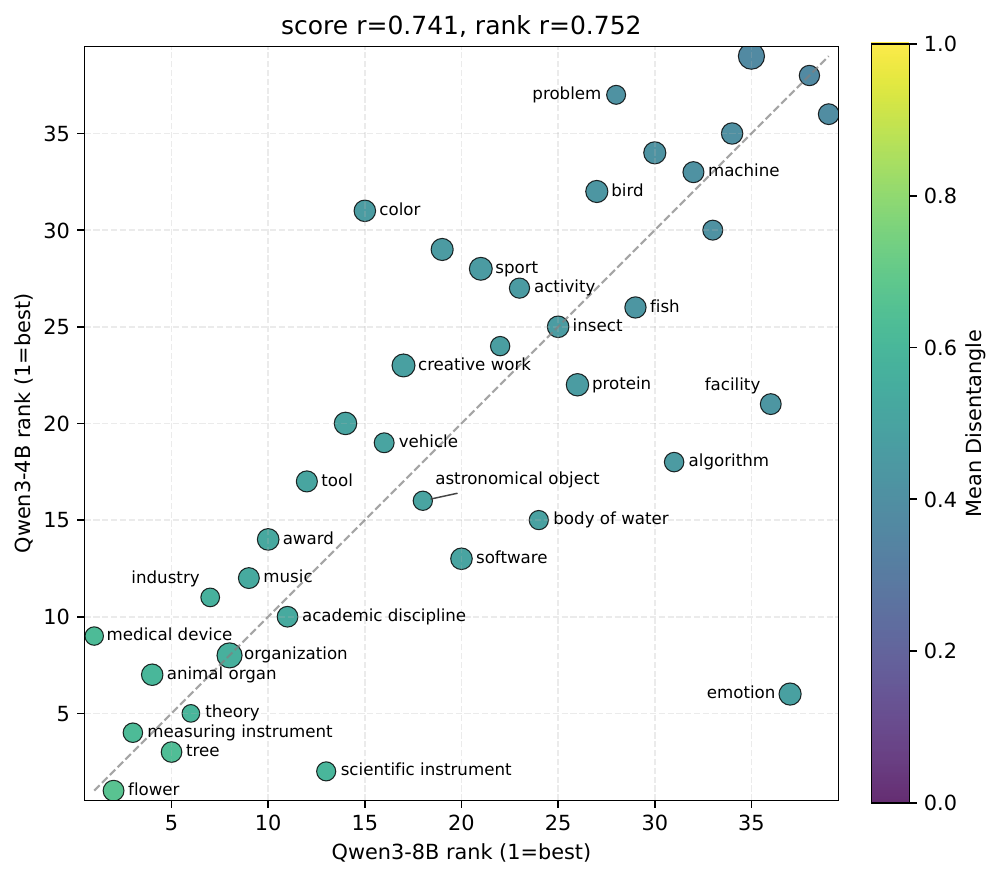}
    \caption{\texttt{subclass of}, base}
  \end{subfigure}
  \hfill
  \begin{subfigure}[t]{0.48\textwidth}
    \centering
    \includegraphics[width=\linewidth]{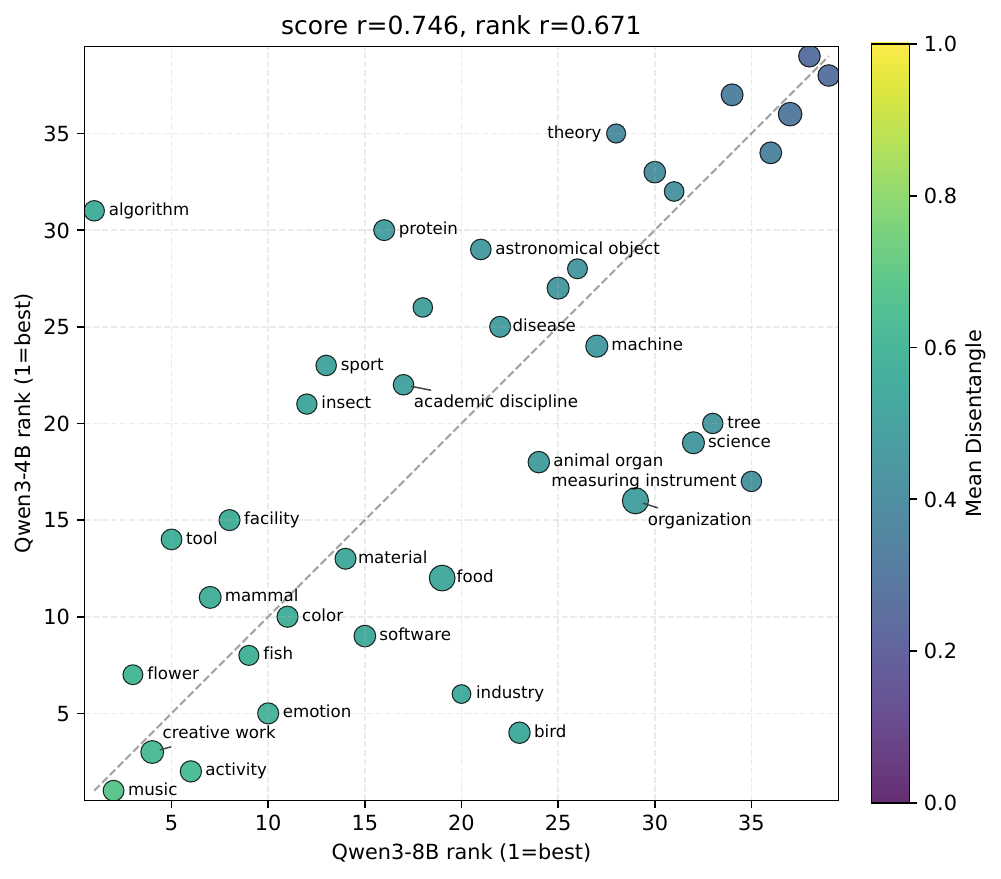}
    \caption{\texttt{subclass of}, source}
  \end{subfigure}

  \caption{Cross-model class-rank correlation for task-macro Disentangle scores. Each point denotes one high-level class, with coordinates given by its Qwen3-8B and Qwen3-4B ranks for the same axiom and role. \emph{Score} $r$ is the Pearson correlation between the two models' class-level Disentangle scores; \emph{rank} $r$ is the Pearson correlation between their ranks, equivalently Spearman's $\rho$. Color gives the arithmetic mean of the two Disentangle scores. Point area increases with class-level evaluation support, using the larger support count across the two runs.}
  \label{fig:appendix-class-level-rank-correlation}
\end{figure*}

Figure~\ref{fig:appendix-class-level-rank-correlation} visualizes these rank correlations. Points close to the diagonal correspond to classes whose relative intervention difficulty is similar in Qwen3-8B and Qwen3-4B, while off-diagonal points identify classes whose difficulty changes substantially with model scale.

The last two columns of Table~\ref{tab:class-level-rank-stability} show a second, more subtle pattern: base and source rankings are only weakly related. Thus, being easy to rewrite in the base role is not the same as being effective in the source role for rewriting other subjects. This role asymmetry is important mechanistically. It suggests that class-level ontology representations do not behave like a single undirected semantic distance metric. Instead, a class can perform well in the base role, the source role, both, or neither. For example, in \texttt{instance of}, \texttt{stock exchange} is strong in the base role but weak in the source role, whereas \texttt{disease}, \texttt{data set}, and \texttt{algorithm} are stronger in the source role. In \texttt{subclass of}, \texttt{measuring instrument}, \texttt{theory}, and \texttt{scientific instrument} are stronger in the base role, while \texttt{algorithm}, \texttt{music}, \texttt{activity}, and \texttt{creative work} are stronger in the source role.

\begin{figure*}[t]
  \centering
  \captionsetup[subfigure]{font=small,skip=2pt}
  \begin{subfigure}[t]{0.48\textwidth}
    \centering
    \includegraphics[width=\linewidth]{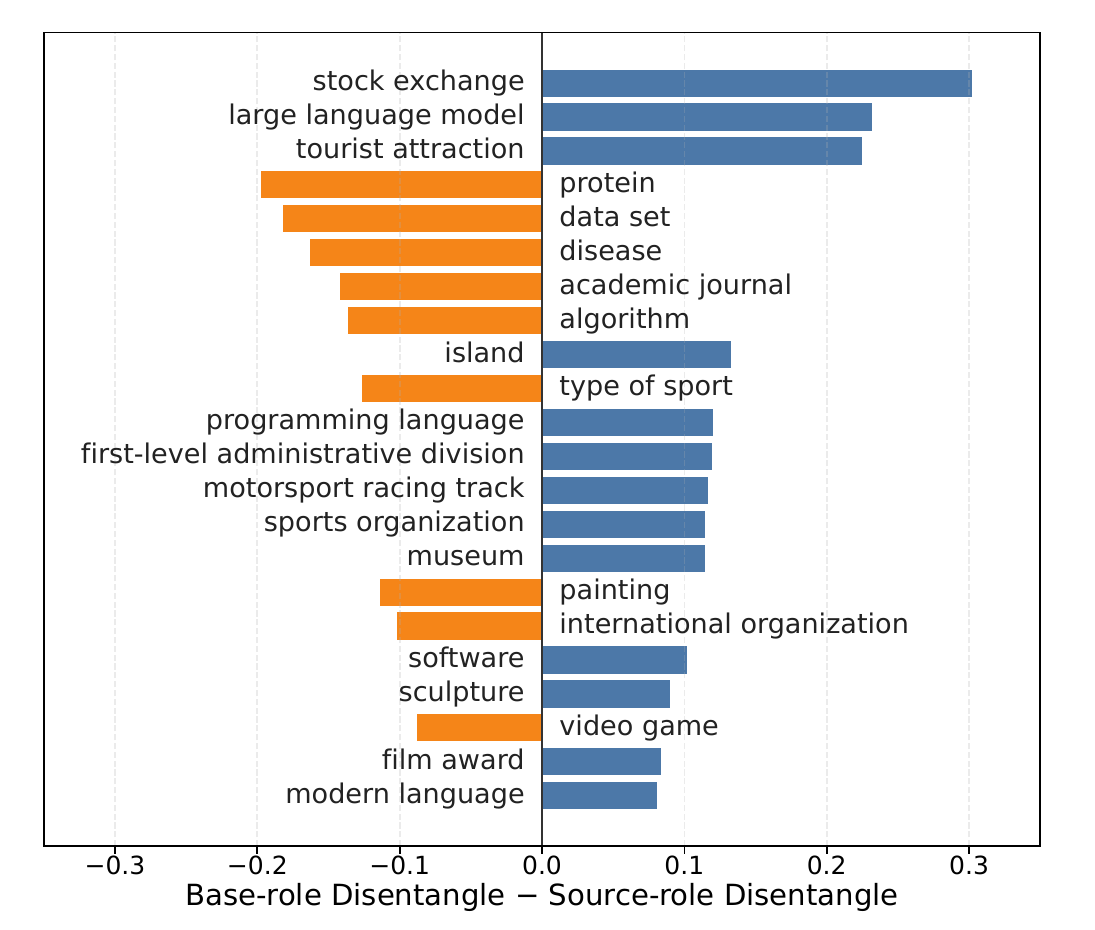}
    \caption{Qwen3-8B, \texttt{instance of}}
  \end{subfigure}%
  \hfill
  \begin{subfigure}[t]{0.48\textwidth}
    \centering
    \includegraphics[width=\linewidth]{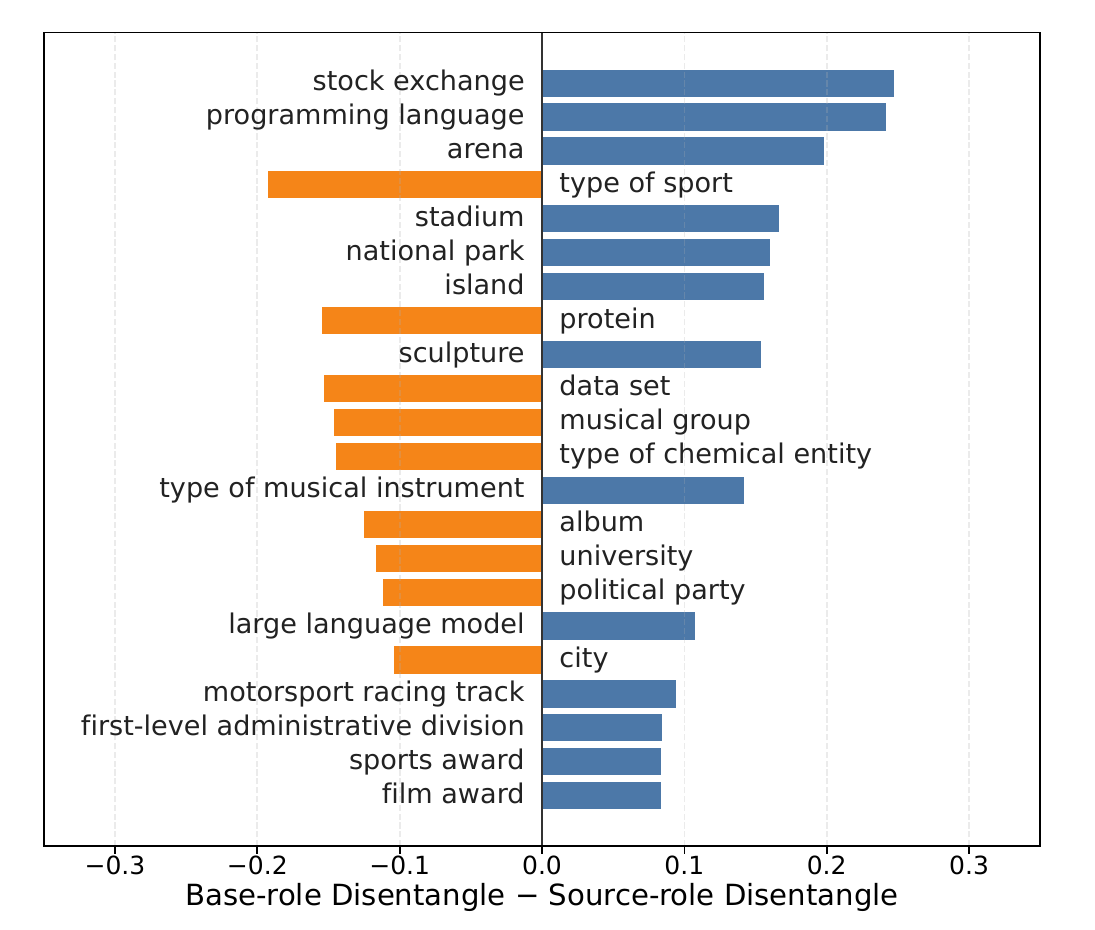}
    \caption{Qwen3-4B, \texttt{instance of}}
  \end{subfigure}

  \vspace{0.4em}
  \begin{subfigure}[t]{0.48\textwidth}
    \centering
    \includegraphics[width=\linewidth]{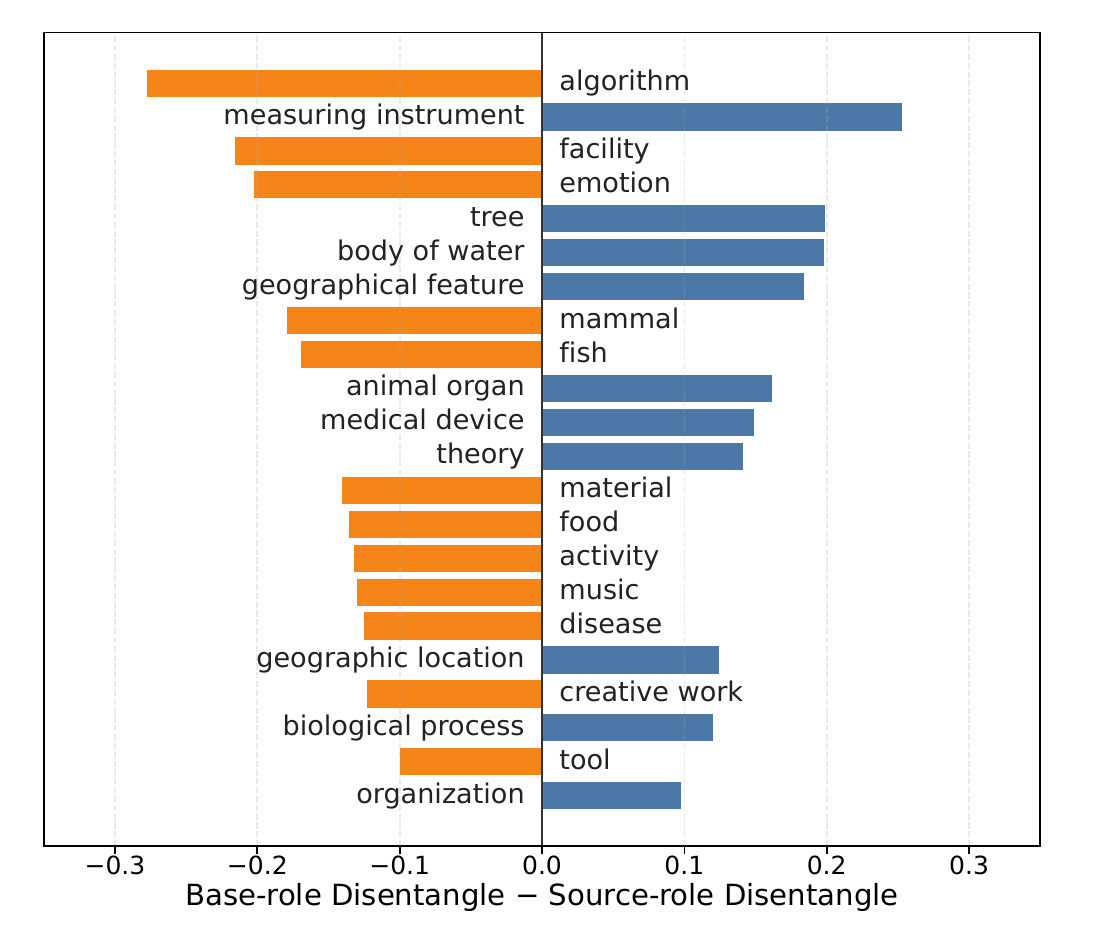}
    \caption{Qwen3-8B, \texttt{subclass of}}
  \end{subfigure}%
  \hfill
  \begin{subfigure}[t]{0.48\textwidth}
    \centering
    \includegraphics[width=\linewidth]{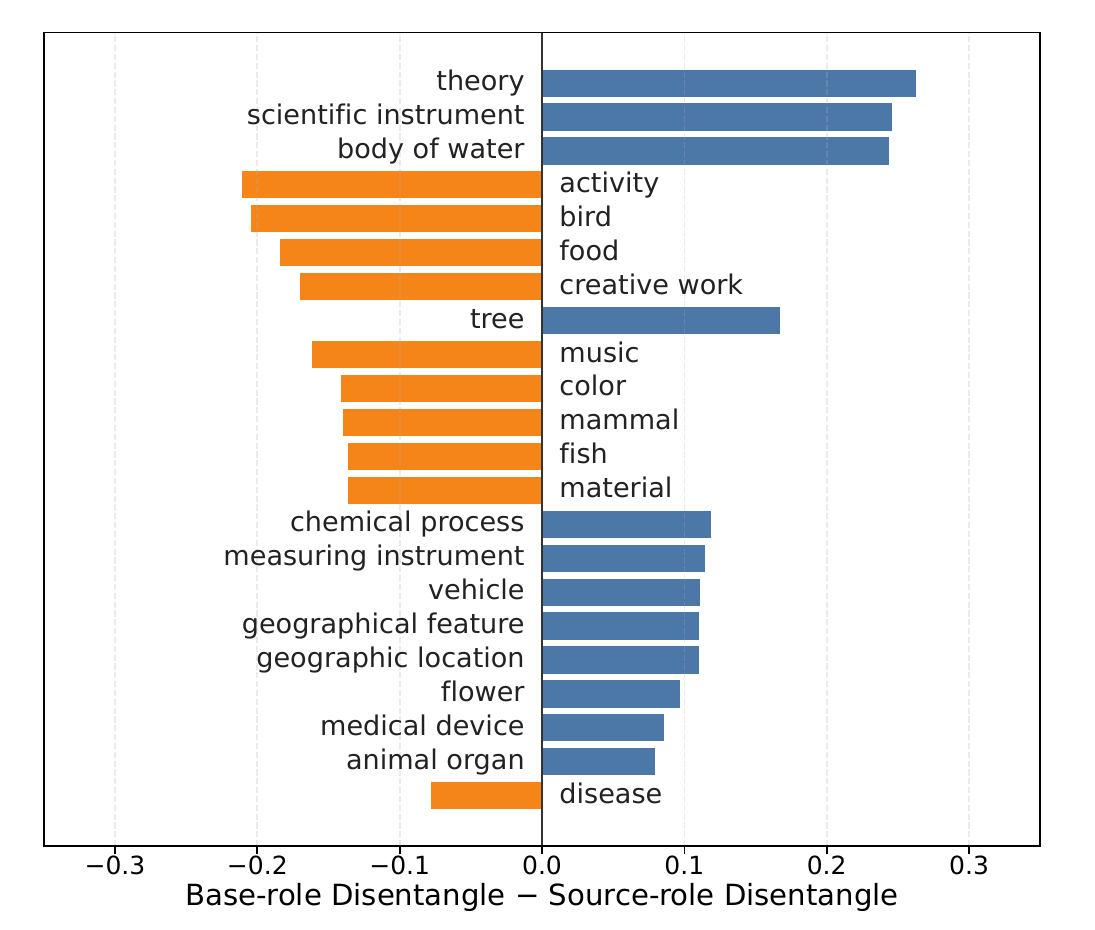}
    \caption{Qwen3-4B, \texttt{subclass of}}
  \end{subfigure}
  
  \caption{Largest base--source class role gaps for the best MDAS settings. Rows correspond to axiom types and columns to model sizes. The horizontal axis reports base-role Disentangle minus source-role Disentangle. Colors denote stronger base-role performance \raisebox{-0.15ex}{\textcolor[HTML]{4C78A8}{\rule{1.2em}{0.7em}}} or stronger source-role performance \raisebox{-0.15ex}{\textcolor[HTML]{F58518}{\rule{1.2em}{0.7em}}}.}
  \label{fig:appendix-class-level-role-gap}
\end{figure*}

Figure~\ref{fig:appendix-class-level-role-gap} shows the largest base--source role gaps directly. Positive values mark classes that are more successful in the base role; negative values mark classes that are more successful in the source role.

\begin{figure*}[t]
  \centering
  \captionsetup[subfigure]{font=small,skip=2pt}
  \begin{subfigure}[t]{0.48\textwidth}
    \centering
    \includegraphics[width=\linewidth]{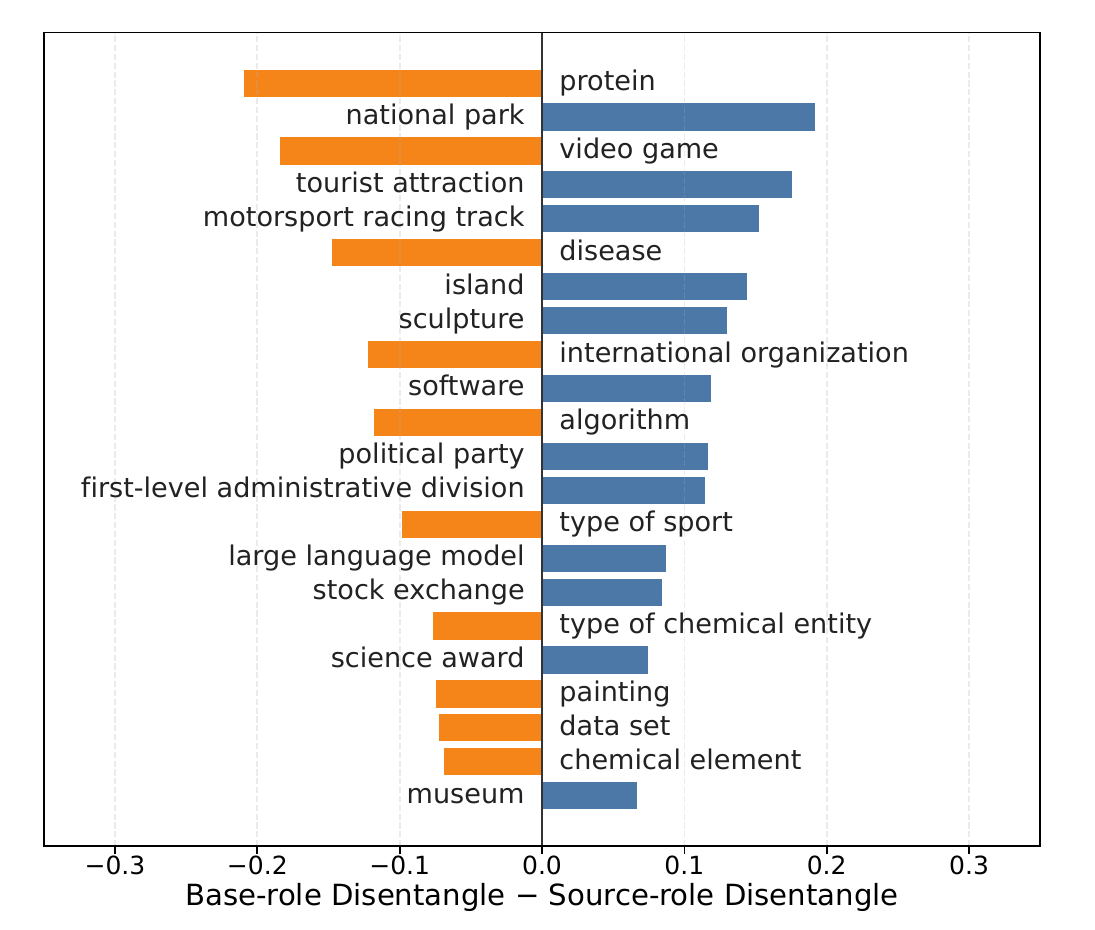}
    \caption{Qwen3-8B, \texttt{instance of}}
  \end{subfigure}%
  \hfill
  \begin{subfigure}[t]{0.48\textwidth}
    \centering
    \includegraphics[width=\linewidth]{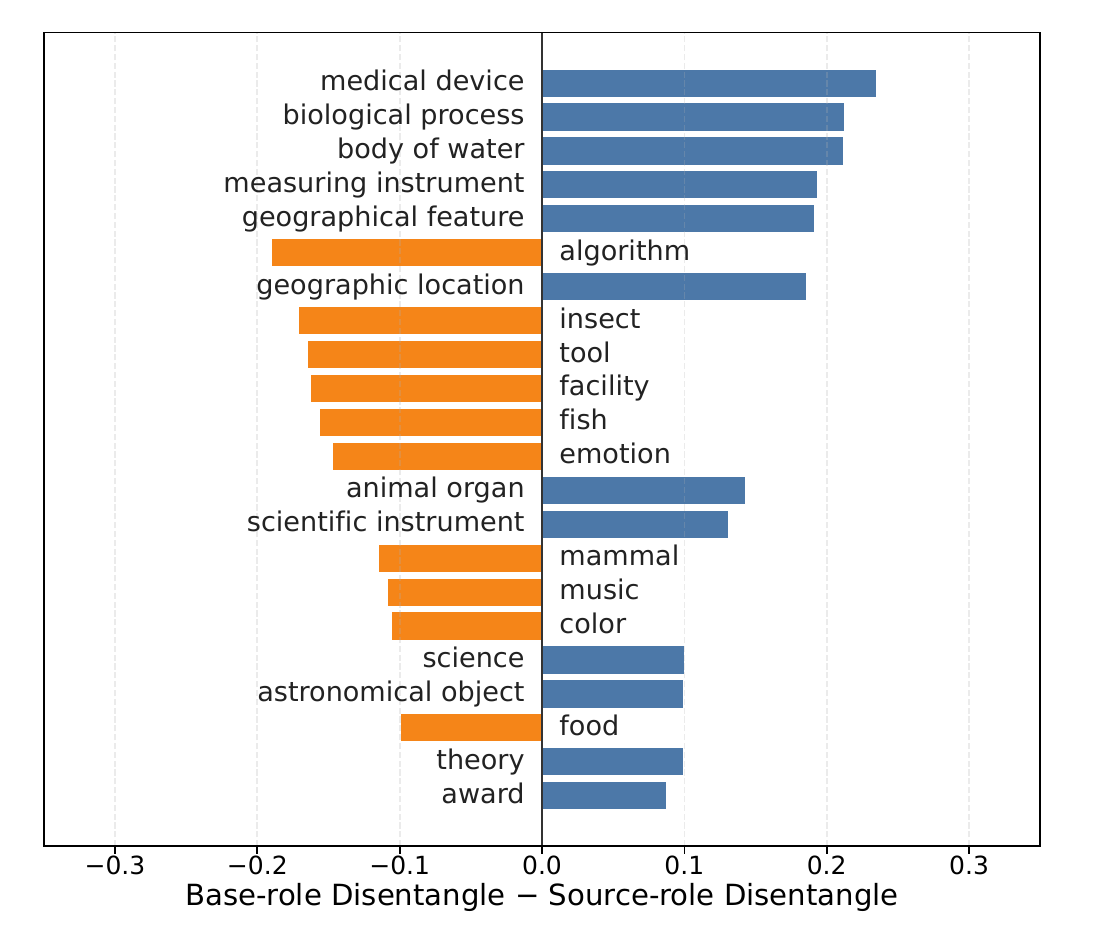}
    \caption{Qwen3-8B, \texttt{subclass of}}
  \end{subfigure}

  \caption{Largest base--source class role gaps for the best Qwen3-8B multi-layer MDAS settings. The horizontal axis reports base-role Disentangle minus source-role Disentangle. Colors denote stronger base-role performance \raisebox{-0.15ex}{\textcolor[HTML]{4C78A8}{\rule{1.2em}{0.7em}}} or stronger source-role performance \raisebox{-0.15ex}{\textcolor[HTML]{F58518}{\rule{1.2em}{0.7em}}}.}
  \label{fig:appendix-class-level-role-gap-parallel}
\end{figure*}

\paragraph{High-performing classes tend to have compact semantic signatures.}
Table~\ref{tab:class-level-consistent-classes} lists recurring high-scoring classes across Qwen3-8B and Qwen3-4B. For \texttt{instance of}, stable high-scoring classes include concrete, attribute-regular categories such as \texttt{airport}, \texttt{city}, \texttt{country}, \texttt{currency}, \texttt{software}, and \texttt{university}. These classes often have distinctive attribute profiles: airports have location and code attributes; currencies have issuers, jurisdictions, and symbols; cities and countries have administrative and demographic attributes. Such compact semantic signatures make the target class easier to change without disrupting unrelated attributes.
For \texttt{subclass of}, stable high-scoring classes in the base role include \texttt{flower}, \texttt{tree}, \texttt{measuring instrument}, and \texttt{medical device}, while stable high-scoring classes in the source role include \texttt{activity}, \texttt{creative work}, \texttt{emotion}, and \texttt{music}. The lower-scoring subclass classes often correspond to broad or internally diverse superclasses such as \texttt{food}, \texttt{biological process}, \texttt{body of water}, \texttt{geographical feature}, and \texttt{material}. These categories are not simply harder labels; they also have more diffuse attribute neighborhoods, making it harder to intervene on the taxonomic relation while preserving auxiliary semantic facts.

\begin{table*}[t]
\centering
\small
\setlength{\tabcolsep}{5pt}
\renewcommand{\arraystretch}{1.08}
\begin{tabular}{lll}
\toprule
\textbf{Axiom} & \textbf{Role} & \textbf{Recurring high-scoring classes} \\
\midrule
\texttt{instance of} & Base &
airport; city; country; currency; science award; software; stock exchange \\
\texttt{instance of} & Source &
airport; city; country; currency; data set; disease; type of sport; university \\
\texttt{subclass of} & Base &
animal organ; flower; measuring instrument; medical device; organization; theory; tree \\
\texttt{subclass of} & Source &
activity; creative work; emotion; fish; flower; music \\
\bottomrule
\end{tabular}
\caption{Representative recurring high-scoring classes from the class-level Disentangle rankings. Rankings are computed separately for each axiom type and base/source role on the Entity split, and listed classes are high-scoring cases for both Qwen3-8B and Qwen3-4B within the same role.}
\label{tab:class-level-consistent-classes}
\end{table*}

\paragraph{Pair-level effects are directionally asymmetric beyond isolated extremes.}
For an unordered class pair $\{A,B\}$, define the reciprocal gap
\[
  \Delta_{A,B}
  =
  \left|
    D(A \leftarrow B) - D(B \leftarrow A)
  \right|,
\]
where $D(A \leftarrow B)$ is the task-macro Disentangle score when a source subject whose target ontology class is $B$ is used to intervene on a base subject whose target ontology class is $A$. We retain only pairs with at least five Cause and five Iso examples in both directions.

\begin{figure*}[t]
  \centering
  \includegraphics[width=0.9\textwidth]{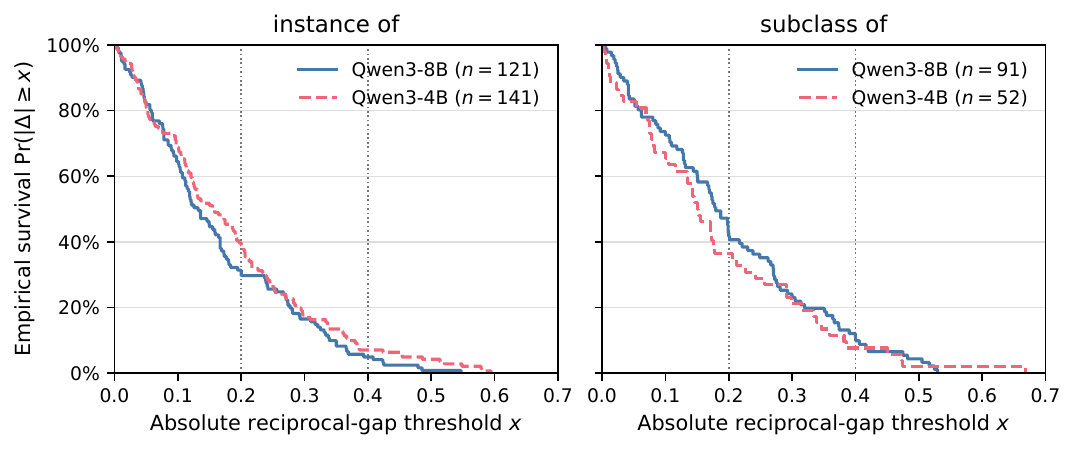}
  \caption{Empirical survival distributions of absolute reciprocal Disentangle gaps for the best MDAS settings. At horizontal position $x$, each curve reports the fraction of eligible unordered class pairs satisfying $\Delta_{A,B}\geq x$. Curves are pair-macro: each eligible reciprocal pair contributes once, without weighting by its evaluation support.}
  \label{fig:appendix-class-level-pair-asymmetry-distribution}
\end{figure*}

Figure~\ref{fig:appendix-class-level-pair-asymmetry-distribution} and Table~\ref{tab:class-level-pair-asymmetry-distribution} summarize the resulting reciprocal gaps. The median gap ranges from 0.132 to 0.179, and 29.8--41.8\% of eligible pairs differ by at least 0.2. Thus, directional asymmetry is not limited to a few extreme examples, although very large gaps remain less common: only 5.0--12.1\% exceed 0.4. The largest eligible examples illustrate the scale of this tail. In Qwen3-8B \texttt{instance of}, \texttt{animal} \(\leftarrow\) \texttt{academic journal} scores 0.742 versus 0.194 in the reverse direction; in Qwen3-4B \texttt{subclass of}, \texttt{body of water} \(\leftarrow\) \texttt{food} scores 0.711 versus 0.042.

The ordering across model sizes is not monotonic: Qwen3-4B has the larger median for \texttt{instance of}, whereas Qwen3-8B has the larger median and heavier tail for \texttt{subclass of}. More fundamentally, the same unordered class pair can change substantially when its base and source roles are exchanged. This suggests that intervention success is not determined only by undirected class similarity, but also by the ordered base--source assignment. Because reciprocal pairs sharing a class are statistically dependent, these distributions are interpreted descriptively rather than as independent samples for pairwise significance testing.

\begin{table*}[t]
\centering
\small
\setlength{\tabcolsep}{5pt}
\renewcommand{\arraystretch}{1.08}
\begin{tabular}{llrcccc}
\toprule
\textbf{Axiom} & \textbf{Model} & $\boldsymbol{N}$ &
\textbf{Mean} & \textbf{Median [IQR]} &
$\boldsymbol{\Pr(\Delta \geq 0.2)}$ &
$\boldsymbol{\Pr(\Delta \geq 0.4)}$ \\
\midrule
\multirow{2}{*}{\texttt{instance of}}
& Qwen3-8B & 121 & 0.166 & 0.132 [0.076, 0.256] & 29.8\% & 5.0\% \\
& Qwen3-4B & 141 & 0.184 & 0.158 [0.069, 0.253] & 39.0\% & 7.1\% \\
\midrule
\multirow{2}{*}{\texttt{subclass of}}
& Qwen3-8B & 91 & 0.204 & 0.179 [0.090, 0.287] & 41.8\% & 12.1\% \\
& Qwen3-4B & 52 & 0.185 & 0.152 [0.075, 0.291] & 36.5\% & 7.7\% \\
\bottomrule
\end{tabular}
\caption{Distribution of absolute reciprocal Disentangle gaps $\Delta_{A,B}$ over eligible unordered class pairs for the best MDAS setting in each model--axiom combination. $N$ is the number of eligible pairs, IQR denotes the interquartile range, and probability columns report the fraction of eligible pairs exceeding each threshold. Eligibility requires at least five Cause and five Iso examples in each direction.}
\label{tab:class-level-pair-asymmetry-distribution}
\end{table*}

Overall, the class-level analysis refines the aggregate findings in the main paper. The main results show that ontology-oriented predictions are causally accessible through learned intervention targets, while their separation from class-specific auxiliary information remains partial. The class-level results show that this partial separation is unevenly distributed across the ontology: MDAS performs better for classes with more regular attribute profiles, and worse for broader or more heterogeneous classes whose target labels are more tightly coupled with auxiliary semantics. The role- and pair-level analyses further show that intervention difficulty depends on the ordered base--source assignment, not only on the class considered in isolation. Together, these results suggest that ontology-level variables are mechanistically accessible in LLMs, but their separability from class-specific information depends on both class structure and intervention direction.

\subsection{Qwen3-4B Single-Layer Results}
\label{appendix:qwen3-4b-single-layer}

\begin{table*}[t]
\centering
\small
\setlength{\tabcolsep}{2pt}
\renewcommand{\arraystretch}{1.05}
\begin{tabular}{llccccc ccc}
\toprule
\multirow{2}{*}{\textbf{Axiom}} &
\multirow{2}{*}{\textbf{Method}} &
\multirow{2}{*}{\textbf{Dimension}} &
\multirow{2}{*}{\textbf{Layer}} &
\multicolumn{3}{c}{\textbf{Entity}} &
\multicolumn{3}{c}{\textbf{Context}} \\
\cmidrule(r){5-7}\cmidrule(l){8-10}
& & & & \textbf{Cause} & \textbf{Iso} & \textbf{Disentangle}
& \textbf{Cause} & \textbf{Iso} & \textbf{Disentangle} \\
\midrule
\multirow{5}{*}{\texttt{instance of}}
& SAE  & 5120 & 14 & 0.408 & 0.375 & 0.391 & 0.350 & 0.369 & 0.359 \\
& DBM  & 2560 & 10 & 0.458 & 0.354 & 0.406 & 0.400 & 0.350 & 0.375 \\
& MDBM & 2560 & 10 & 0.404 & 0.436 & 0.420 & 0.340 & 0.430 & 0.385 \\
& DAS  & 1024 & 22 & 0.516 & 0.537 & 0.527 & 0.489 & 0.509 & 0.499 \\
& MDAS & 2048 & 16 & 0.520 & 0.592 & \textbf{0.556} & 0.471 & 0.639 & \textbf{0.555} \\
\midrule
\multirow{5}{*}{\texttt{subclass of}}
& SAE  & 10240 & 13 & 0.562 & 0.257 & 0.409 & 0.633 & 0.252 & 0.442 \\
& DBM  & 2560  & 12 & 0.525 & 0.333 & 0.429 & 0.549 & 0.351 & 0.450 \\
& MDBM & 2560  & 12 & 0.526 & 0.360 & 0.443 & 0.536 & 0.370 & 0.453 \\
& DAS  & 2048  & 16 & 0.632 & 0.289 & 0.460 & 0.683 & 0.302 & 0.492 \\
& MDAS & 2048  & 20 & 0.511 & 0.510 & \textbf{0.511} & 0.540 & 0.591 & \textbf{0.566} \\
\bottomrule
\end{tabular}
\caption{Qwen3-4B single-layer intervention results under the same reporting rule as the Qwen3-8B comparison. For each axiom type and method, we report the layer/configuration with the strongest Entity Disentangle score and evaluate the same configuration on Context. Dimension denotes subspace rank for DAS/MDAS, ambient mask dimension for DBM/MDBM, and latent width for SAE.}
\label{tab:single-layer-best-qwen3-4b}
\end{table*}

Table~\ref{tab:single-layer-best-qwen3-4b} reports the corresponding single-layer results for Qwen3-4B. The results preserve the complete method ordering observed for Qwen3-8B on both Entity and Context: MDAS $>$ DAS $>$ MDBM $>$ DBM $>$ SAE. 

\subsection{Layer-Wise Intervention Profiles}
\label{appendix:layer-wise-profiles}

\begin{figure*}[t]
  \centering
  \captionsetup[subfigure]{font=small,skip=2pt}
  \begin{subfigure}[t]{0.42\textwidth}
    \centering
    \includegraphics[width=\linewidth]{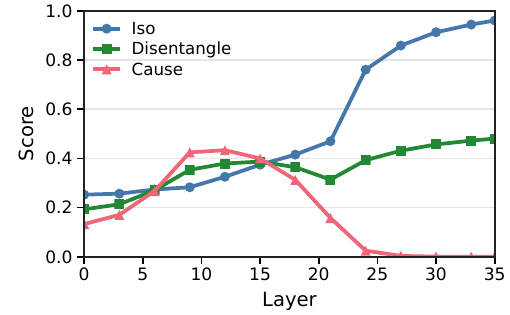}
    \caption{\texttt{instance of}, Qwen3-4B, $2d$, $C=1000$}
  \end{subfigure}
  \begin{subfigure}[t]{0.42\textwidth}
    \centering
    \includegraphics[width=\linewidth]{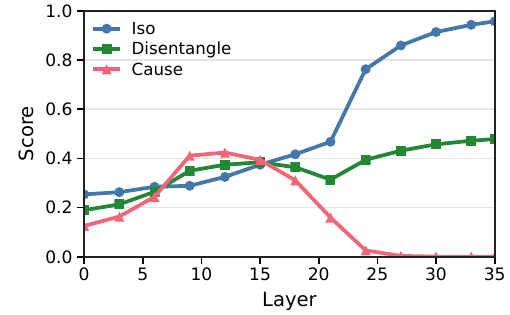}
    \caption{\texttt{instance of}, Qwen3-4B, $4d$, $C=1000$}
  \end{subfigure}
  \vspace{2pt}
  \begin{subfigure}[t]{0.42\textwidth}
    \centering
    \includegraphics[width=\linewidth]{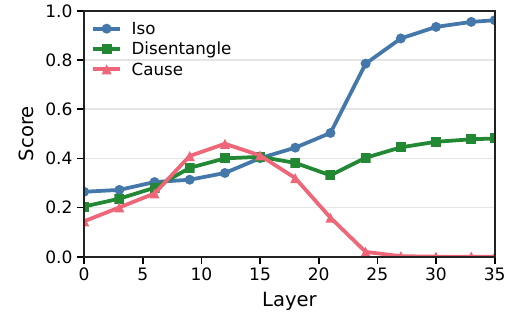}
    \caption{\texttt{instance of}, Qwen3-8B, $2d$, $C=1000$}
  \end{subfigure}
  \begin{subfigure}[t]{0.42\textwidth}
    \centering
    \includegraphics[width=\linewidth]{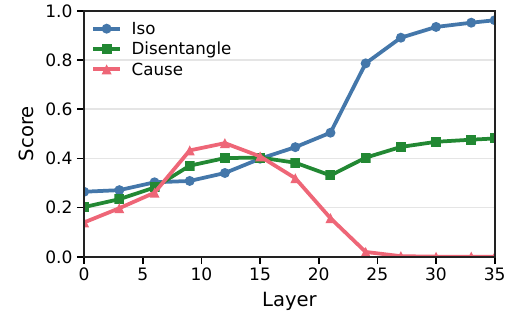}
    \caption{\texttt{instance of}, Qwen3-8B, $4d$, $C=1000$}
  \end{subfigure}
  \begin{subfigure}[t]{0.42\textwidth}
    \centering
    \includegraphics[width=\linewidth]{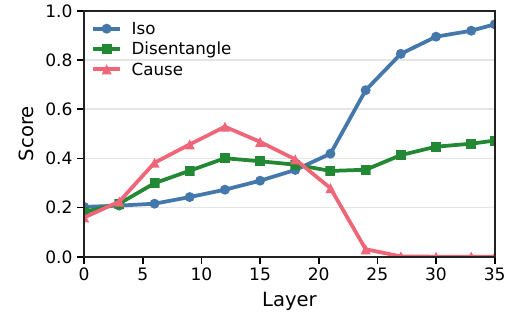}
    \caption{\texttt{subclass of}, Qwen3-4B, $2d$, $C=10$}
  \end{subfigure}
  \begin{subfigure}[t]{0.42\textwidth}
    \centering
    \includegraphics[width=\linewidth]{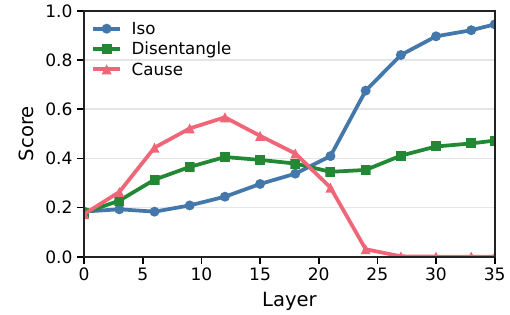}
    \caption{\texttt{subclass of}, Qwen3-4B, $4d$, $C=1000$}
  \end{subfigure}
  \vspace{2pt}
  \begin{subfigure}[t]{0.42\textwidth}
    \centering
    \includegraphics[width=\linewidth]{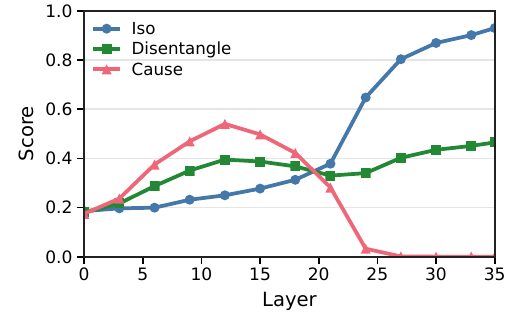}
    \caption{\texttt{subclass of}, Qwen3-8B, $2d$, $C=10$}
  \end{subfigure}
  \begin{subfigure}[t]{0.42\textwidth}
    \centering
    \includegraphics[width=\linewidth]{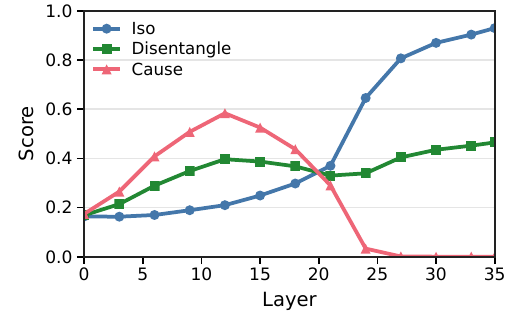}
    \caption{\texttt{subclass of}, Qwen3-8B, $4d$, $C=1000$}
  \end{subfigure}
  \caption{SAE single-layer profiles across model sizes, axiom types, and latent widths.}
  \label{fig:appendix-sae-profiles}
\end{figure*}

\begin{figure*}[t]
  \centering
  \captionsetup[subfigure]{font=small,skip=2pt}
  \begin{subfigure}[t]{0.42\textwidth}
    \centering
    \includegraphics[width=\linewidth]{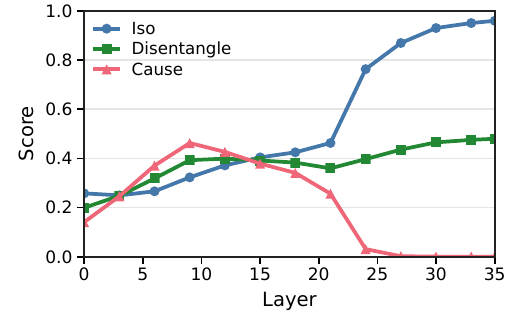}
    \caption{\texttt{instance of}, Qwen3-4B, DBM}
  \end{subfigure}
  \begin{subfigure}[t]{0.42\textwidth}
    \centering
    \includegraphics[width=\linewidth]{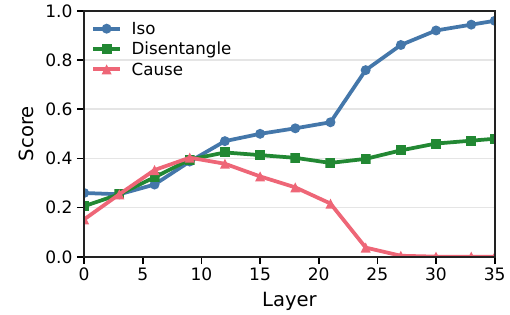}
    \caption{\texttt{instance of}, Qwen3-4B, MDBM}
  \end{subfigure}
  \vspace{2pt}
  \begin{subfigure}[t]{0.42\textwidth}
    \centering
    \includegraphics[width=\linewidth]{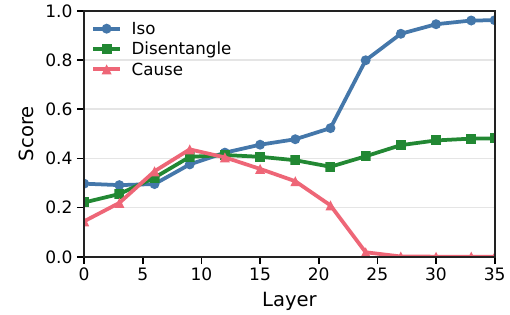}
    \caption{\texttt{instance of}, Qwen3-8B, DBM}
  \end{subfigure}
  \begin{subfigure}[t]{0.42\textwidth}
    \centering
    \includegraphics[width=\linewidth]{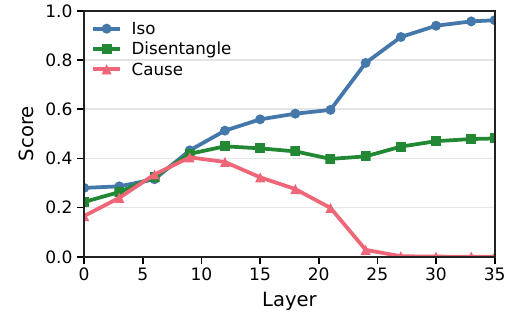}
    \caption{\texttt{instance of}, Qwen3-8B, MDBM}
  \end{subfigure}
  \begin{subfigure}[t]{0.42\textwidth}
    \centering
    \includegraphics[width=\linewidth]{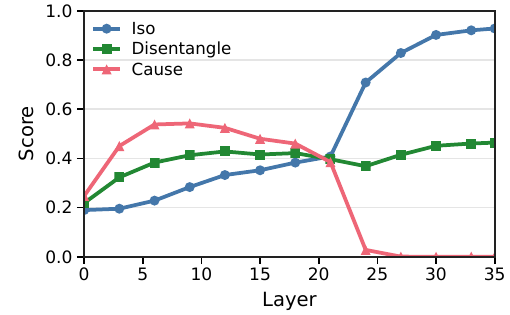}
    \caption{\texttt{subclass of}, Qwen3-4B, DBM}
  \end{subfigure}
  \begin{subfigure}[t]{0.42\textwidth}
    \centering
    \includegraphics[width=\linewidth]{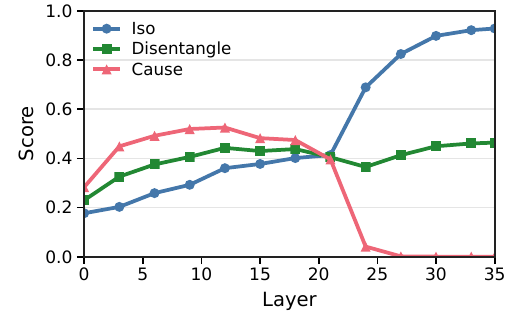}
    \caption{\texttt{subclass of}, Qwen3-4B, MDBM}
  \end{subfigure}
  \vspace{2pt}
  \begin{subfigure}[t]{0.42\textwidth}
    \centering
    \includegraphics[width=\linewidth]{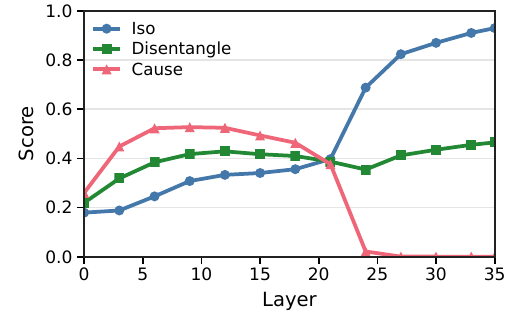}
    \caption{\texttt{subclass of}, Qwen3-8B, DBM}
  \end{subfigure}
  \begin{subfigure}[t]{0.42\textwidth}
    \centering
    \includegraphics[width=\linewidth]{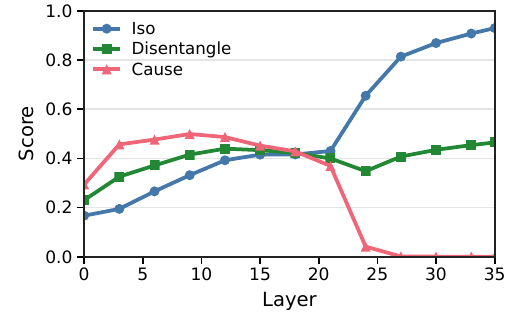}
    \caption{\texttt{subclass of}, Qwen3-8B, MDBM}
  \end{subfigure}
  \caption{Fixed-layer DBM/MDBM intervention profiles across model sizes and axiom types.}
  \label{fig:appendix-fixed-dbm-profiles}
\end{figure*}

\begin{figure*}[t]
  \centering
  \captionsetup[subfigure]{font=small,skip=2pt}
  \begin{subfigure}[t]{0.42\textwidth}
    \centering
    \includegraphics[width=\linewidth]{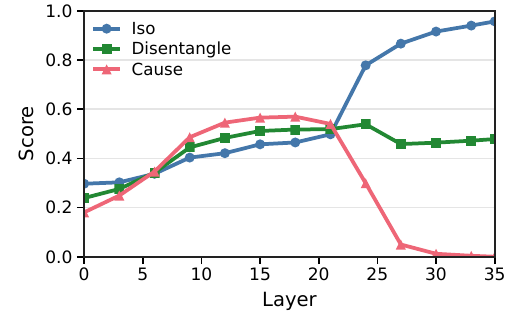}
    \caption{\texttt{instance of}, Qwen3-4B, DAS}
  \end{subfigure}
  \begin{subfigure}[t]{0.42\textwidth}
    \centering
    \includegraphics[width=\linewidth]{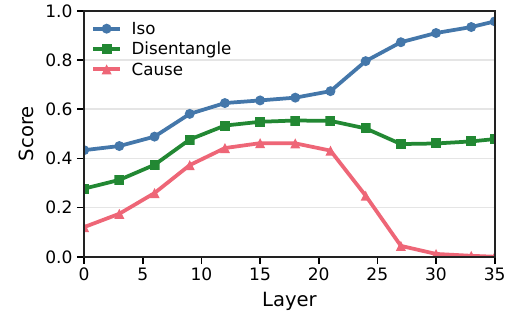}
    \caption{\texttt{instance of}, Qwen3-4B, MDAS}
  \end{subfigure}
  \vspace{2pt}
  \begin{subfigure}[t]{0.42\textwidth}
    \centering
    \includegraphics[width=\linewidth]{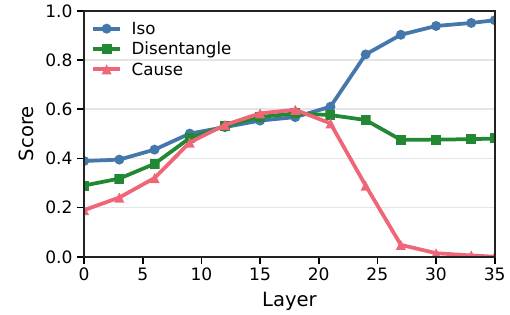}
    \caption{\texttt{instance of}, Qwen3-8B, DAS}
  \end{subfigure}
  \begin{subfigure}[t]{0.42\textwidth}
    \centering
    \includegraphics[width=\linewidth]{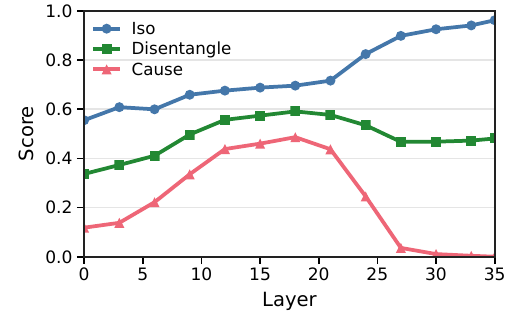}
    \caption{\texttt{instance of}, Qwen3-8B, MDAS}
  \end{subfigure}
  \vspace{2pt}
  \begin{subfigure}[t]{0.42\textwidth}
    \centering
    \includegraphics[width=\linewidth]{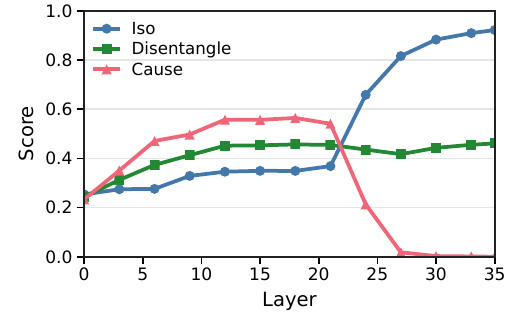}
    \caption{\texttt{subclass of}, Qwen3-4B, DAS}
  \end{subfigure}
  \begin{subfigure}[t]{0.42\textwidth}
    \centering
    \includegraphics[width=\linewidth]{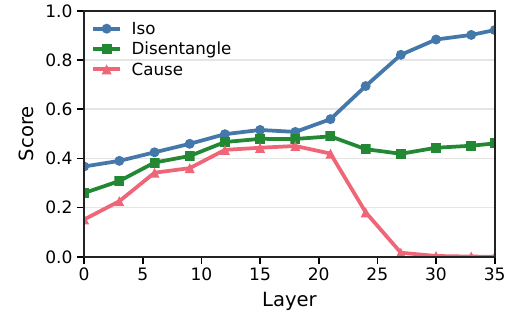}
    \caption{\texttt{subclass of}, Qwen3-4B, MDAS}
  \end{subfigure}
  \vspace{2pt}
  \begin{subfigure}[t]{0.42\textwidth}
    \centering
    \includegraphics[width=\linewidth]{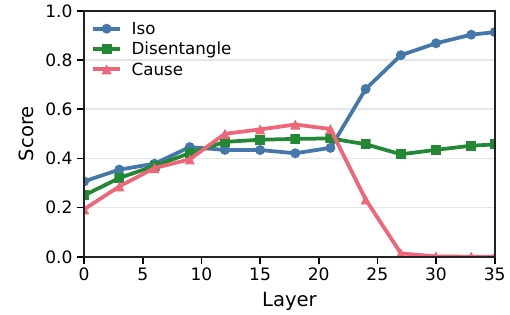}
    \caption{\texttt{subclass of}, Qwen3-8B, DAS}
  \end{subfigure}
  \begin{subfigure}[t]{0.42\textwidth}
    \centering
    \includegraphics[width=\linewidth]{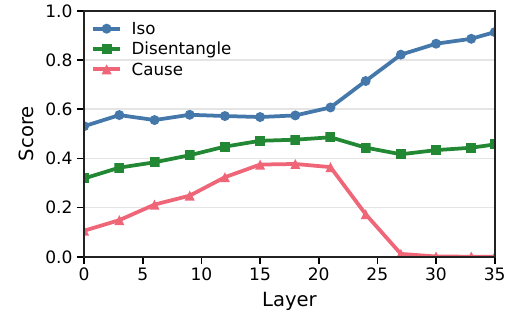}
    \caption{\texttt{subclass of}, Qwen3-8B, MDAS}
  \end{subfigure}
  \caption{Fixed-layer DAS/MDAS intervention profiles across model sizes and axiom types with 1024-dimensional subspaces.}
  \label{fig:appendix-fixed-das-1024-profiles}
\end{figure*}

\begin{figure*}[t]
  \centering
  \captionsetup[subfigure]{font=small,skip=2pt}
  \begin{subfigure}[t]{0.42\textwidth}
    \centering
    \includegraphics[width=\linewidth]{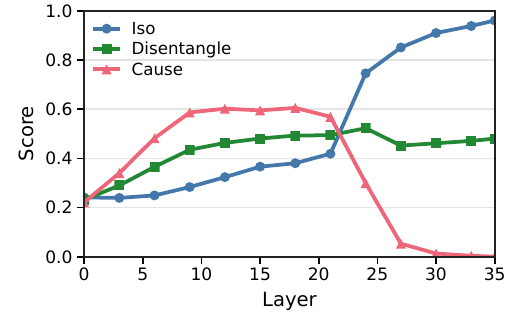}
    \caption{\texttt{instance of}, Qwen3-4B, DAS}
  \end{subfigure}
  \begin{subfigure}[t]{0.42\textwidth}
    \centering
    \includegraphics[width=\linewidth]{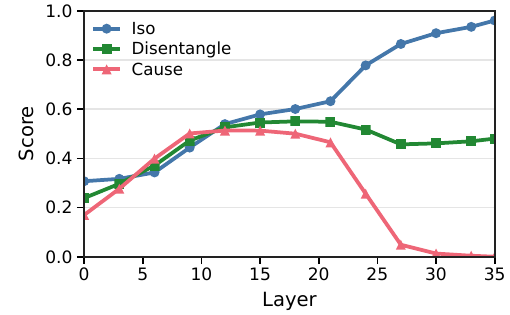}
    \caption{\texttt{instance of}, Qwen3-4B, MDAS}
  \end{subfigure}
  \vspace{2pt}
  \begin{subfigure}[t]{0.42\textwidth}
    \centering
    \includegraphics[width=\linewidth]{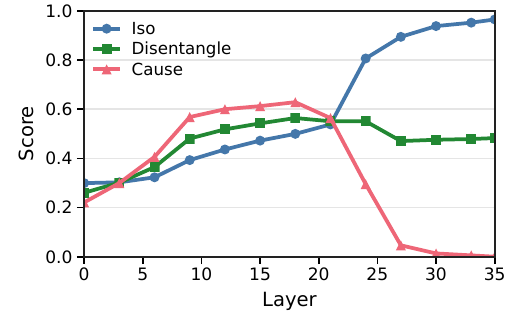}
    \caption{\texttt{instance of}, Qwen3-8B, DAS}
  \end{subfigure}
  \begin{subfigure}[t]{0.42\textwidth}
    \centering
    \includegraphics[width=\linewidth]{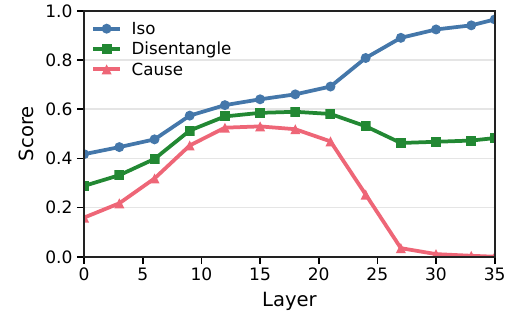}
    \caption{\texttt{instance of}, Qwen3-8B, MDAS}
  \end{subfigure}
  \vspace{2pt}
  \begin{subfigure}[t]{0.42\textwidth}
    \centering
    \includegraphics[width=\linewidth]{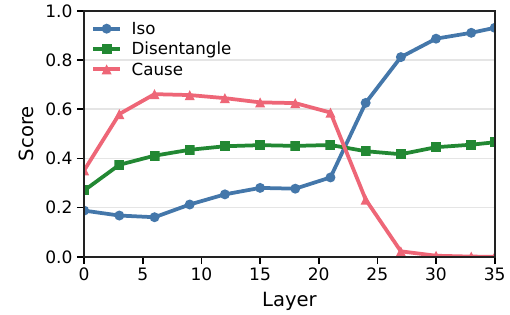}
    \caption{\texttt{subclass of}, Qwen3-4B, DAS}
  \end{subfigure}
  \begin{subfigure}[t]{0.42\textwidth}
    \centering
    \includegraphics[width=\linewidth]{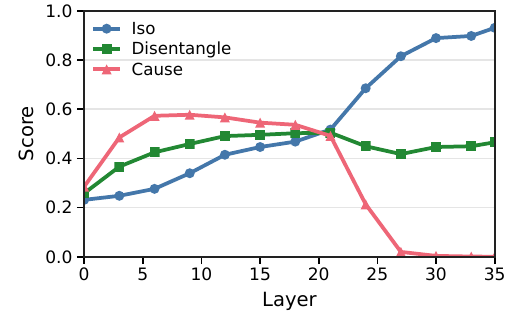}
    \caption{\texttt{subclass of}, Qwen3-4B, MDAS}
  \end{subfigure}
  \vspace{2pt}
  \begin{subfigure}[t]{0.42\textwidth}
    \centering
    \includegraphics[width=\linewidth]{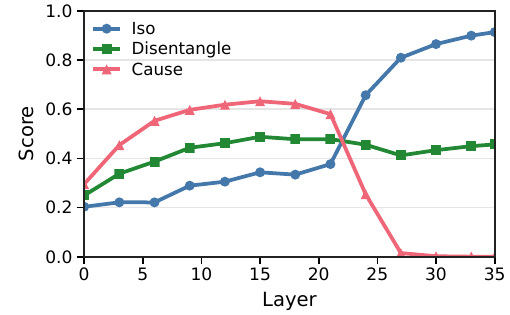}
    \caption{\texttt{subclass of}, Qwen3-8B, DAS}
  \end{subfigure}
  \begin{subfigure}[t]{0.42\textwidth}
    \centering
    \includegraphics[width=\linewidth]{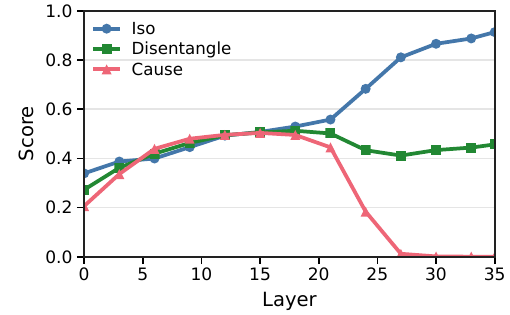}
    \caption{\texttt{subclass of}, Qwen3-8B, MDAS}
  \end{subfigure}
  \caption{Fixed-layer DAS/MDAS intervention profiles across model sizes and axiom types with 2048-dimensional subspaces.}
  \label{fig:appendix-fixed-das-2048-profiles}
\end{figure*}

\begin{figure*}[p]
  \centering
  \captionsetup[subfigure]{font=small,skip=2pt}
  \begin{subfigure}[t]{0.45\textwidth}
    \centering
    \includegraphics[width=\linewidth]{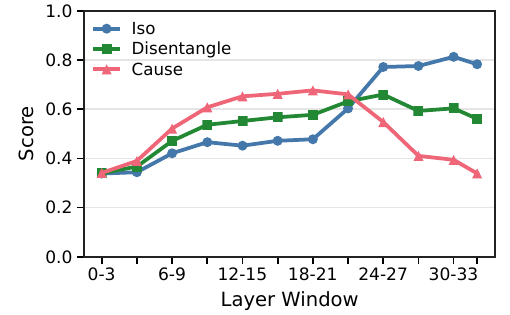}
    \caption{DAS, $256\times4$}
  \end{subfigure}%
  \begin{subfigure}[t]{0.45\textwidth}
    \centering
    \includegraphics[width=\linewidth]{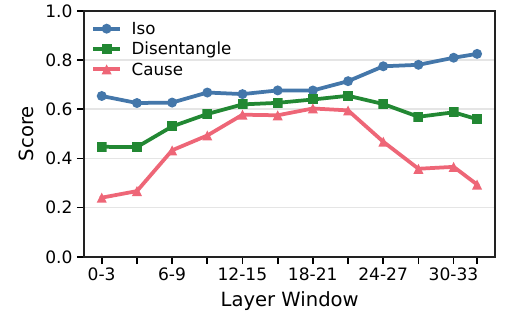}
    \caption{MDAS, $256\times4$}
  \end{subfigure}
  \vspace{2pt}
  \begin{subfigure}[t]{0.45\textwidth}
    \centering
    \includegraphics[width=\linewidth]{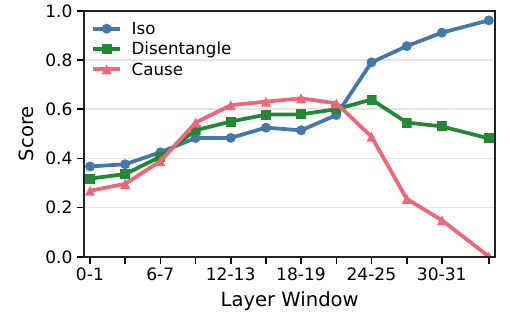}
    \caption{DAS, $512\times2$}
  \end{subfigure}%
  \begin{subfigure}[t]{0.45\textwidth}
    \centering
    \includegraphics[width=\linewidth]{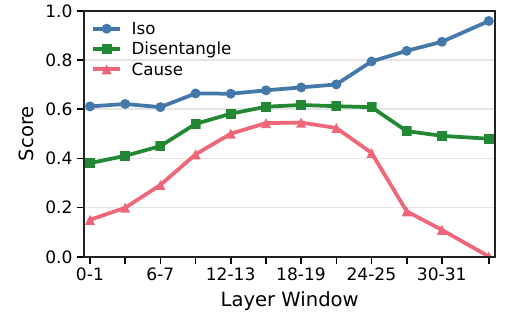}
    \caption{MDAS, $512\times2$}
  \end{subfigure}
  \vspace{2pt}
  \begin{subfigure}[t]{0.45\textwidth}
    \centering
    \includegraphics[width=\linewidth]{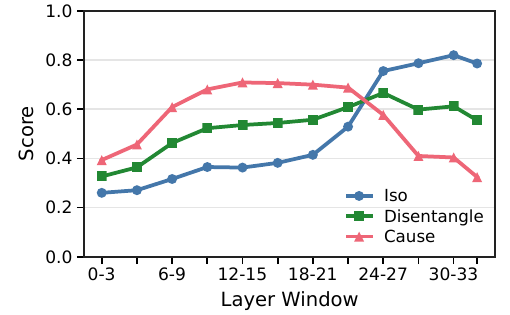}
    \caption{DAS, $512\times4$}
  \end{subfigure}%
  \begin{subfigure}[t]{0.45\textwidth}
    \centering
    \includegraphics[width=\linewidth]{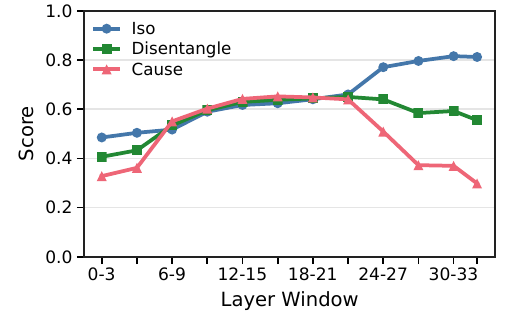}
    \caption{MDAS, $512\times4$}
  \end{subfigure}
  \caption{Fixed-window multi-layer intervention profiles for Qwen3-8B on \texttt{instance of}. Each panel reports Cause, Iso, and Disentangle on 12 contiguous windows. The $a\times b$ notation denotes $a$ intervened dimensions per layer over a $b$-layer window.}
  \label{fig:appendix-parallel-fixed-instanceof-qwen3-8b}
\end{figure*}

\begin{figure*}[p]
  \centering
  \captionsetup[subfigure]{font=small,skip=2pt}
  \begin{subfigure}[t]{0.45\textwidth}
    \centering
    \includegraphics[width=\linewidth]{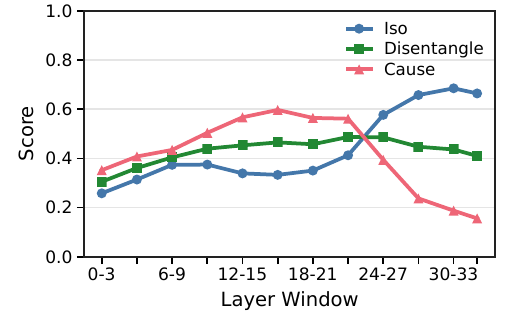}
    \caption{DAS, $256\times4$}
  \end{subfigure}%
  \begin{subfigure}[t]{0.45\textwidth}
    \centering
    \includegraphics[width=\linewidth]{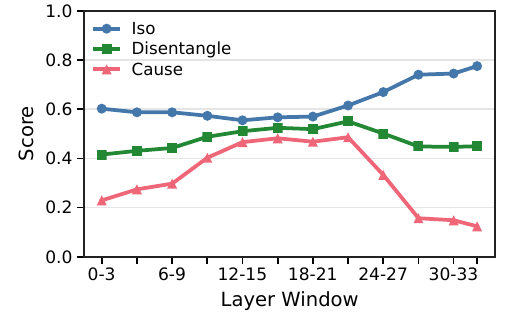}
    \caption{MDAS, $256\times4$}
  \end{subfigure}
  \vspace{2pt}
  \begin{subfigure}[t]{0.45\textwidth}
    \centering
    \includegraphics[width=\linewidth]{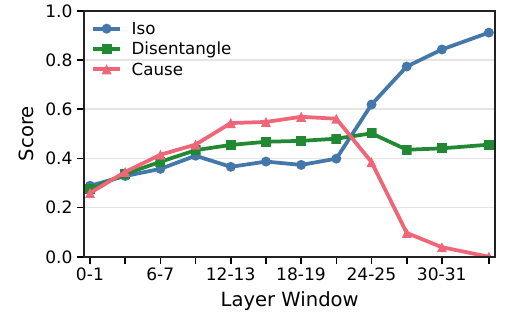}
    \caption{DAS, $512\times2$}
  \end{subfigure}%
  \begin{subfigure}[t]{0.45\textwidth}
    \centering
    \includegraphics[width=\linewidth]{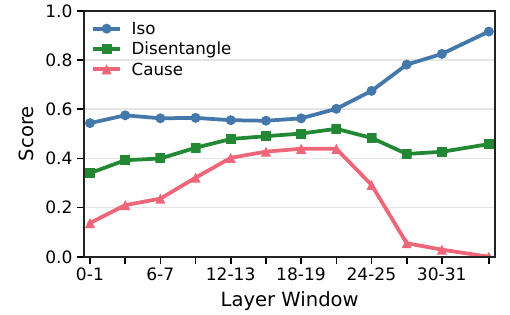}
    \caption{MDAS, $512\times2$}
  \end{subfigure}
  \vspace{2pt}
  \begin{subfigure}[t]{0.45\textwidth}
    \centering
    \includegraphics[width=\linewidth]{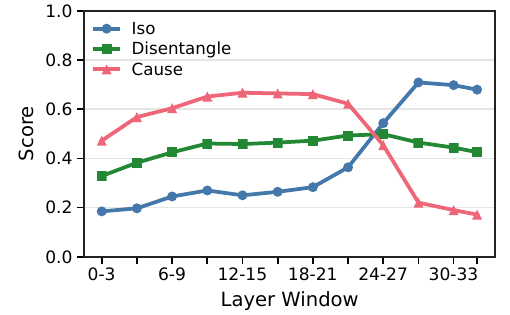}
    \caption{DAS, $512\times4$}
  \end{subfigure}%
  \begin{subfigure}[t]{0.45\textwidth}
    \centering
    \includegraphics[width=\linewidth]{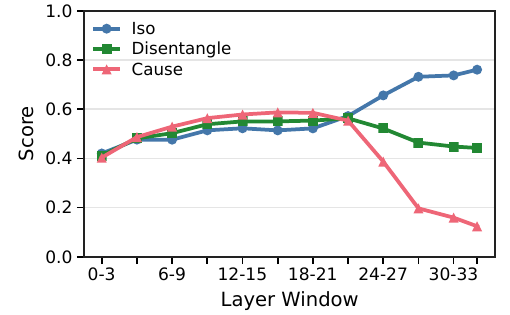}
    \caption{MDAS, $512\times4$}
  \end{subfigure}
  \caption{Fixed-window multi-layer intervention profiles for Qwen3-8B on \texttt{subclass of}. Each panel reports Cause, Iso, and Disentangle on 12 contiguous windows. The $a\times b$ notation denotes $a$ intervened dimensions per layer over a $b$-layer window.}
  \label{fig:appendix-parallel-fixed-subclassof-qwen3-8b}
\end{figure*}

Figure~\ref{fig:appendix-sae-profiles} shows the SAE layer-wise profiles. Figure~\ref{fig:appendix-fixed-dbm-profiles} shows the single-layer DBM/MDBM profiles. Figures~\ref{fig:appendix-fixed-das-1024-profiles} and~\ref{fig:appendix-fixed-das-2048-profiles} show the single-layer DAS/MDAS profiles grouped by subspace rank. Figures~\ref{fig:appendix-parallel-fixed-instanceof-qwen3-8b} and~\ref{fig:appendix-parallel-fixed-subclassof-qwen3-8b} report Qwen3-8B multi-layer DAS/MDAS fixed-window profiles. For single-layer profiles, we report 12 of the 36 layers. For multi-layer profiles, we report 12 evenly spaced contiguous windows spanning the first to last transformer layer.

\subsection{Highest-Cause Single-Layer Configurations}
\label{appendix:highest-cause}

Table~\ref{tab:highest-cause-single-layer} reports the highest-Cause single-layer configuration for each model, axiom type, and method.

\begin{table*}[t]
\centering
\small
\setlength{\tabcolsep}{4pt}
\renewcommand{\arraystretch}{1.05}
\begin{tabular}{lllccccc}
\toprule
\textbf{Axiom} & \textbf{Model} & \textbf{Method} & \textbf{Dim.} & \textbf{Layer} & \textbf{Cause} & \textbf{Iso} & \textbf{Disentangle} \\
\midrule
\multirow{10}{*}{\texttt{instance of}} & \multirow{5}{*}{Qwen3-4B} & SAE  & 5120 & 10 & 0.458 & 0.292 & 0.375 \\
& & DBM  & 2560 & 9  & 0.463 & 0.323 & 0.393 \\
& & MDBM & 2560 & 10 & 0.404 & 0.436 & 0.420 \\
& & DAS  & 2048 & 14 & \textbf{0.615} & 0.346 & 0.481 \\
& & MDAS & 2048 & 13 & 0.524 & 0.559 & 0.541 \\
\cmidrule(lr){2-8}
& \multirow{5}{*}{Qwen3-8B} & SAE  & 16384 & 13 & 0.472 & 0.350 & 0.411 \\
& & DBM  & 4096 & 10 & 0.441 & 0.409 & 0.425 \\
& & MDBM & 4096 & 10 & 0.418 & 0.469 & 0.443 \\
& & DAS  & 2048 & 13 & \textbf{0.633} & 0.425 & 0.529 \\
& & MDAS & 2048 & 13 & 0.536 & 0.624 & 0.580 \\
\midrule
\multirow{10}{*}{\texttt{subclass of}} & \multirow{5}{*}{Qwen3-4B} & SAE  & 10240 & 12 & 0.567 & 0.244 & 0.406 \\
& & DBM  & 2560 & 8  & 0.553 & 0.257 & 0.405 \\
& & MDBM & 2560 & 10 & 0.531 & 0.318 & 0.425 \\
& & DAS  & 2048 & 8  & \textbf{0.668} & 0.182 & 0.425 \\
& & MDAS & 2048 & 8  & 0.597 & 0.313 & 0.455 \\
\cmidrule(lr){2-8}
& \multirow{5}{*}{Qwen3-8B} & SAE  & 16384 & 12 & 0.584 & 0.210 & 0.397 \\
& & DBM  & 4096 & 8  & 0.530 & 0.282 & 0.406 \\
& & MDBM & 4096 & 10 & 0.504 & 0.362 & 0.433 \\
& & DAS  & 2048 & 13 & \textbf{0.636} & 0.308 & 0.472 \\
& & MDAS & 2048 & 13 & 0.520 & 0.500 & 0.510 \\
\bottomrule
\end{tabular}
\caption{Highest-Cause single-layer intervention results on the Entity split, selecting the configuration with the highest Cause score for each model, axiom type, and method.}
\label{tab:highest-cause-single-layer}
\end{table*}

\clearpage
\onecolumn


\begin{longtable}{
  p{\dimexpr0.23\textwidth-2\tabcolsep\relax}
  p{\dimexpr0.10\textwidth-2\tabcolsep\relax}
  p{\dimexpr0.66\textwidth-2\tabcolsep\relax}
}
\toprule
\textbf{Class} & \textbf{\# Entity} & \textbf{Attributes} \\
\midrule
\endfirsthead

\toprule
\textbf{Class} & \textbf{\# Entity} & \textbf{Attributes} \\
\midrule
\endhead

\midrule
\multicolumn{3}{r}{Continued on next page} \\
\endfoot

\endlastfoot

academic conference & 30 & inception, main subject, event interval, organizer \\
academic journal & 27 & main subject, country of origin, publisher, ISSN, indexed in bibliographic review, inception, editor \\
airport & 15 & country, located in the administrative territorial entity, date of official opening, elevation above sea level, IATA airport code, ICAO airport code, airline hub \\
album & 15 & form of creative work, genre, performer, publication date, record label, producer, number of parts of this work, tracklist \\
algorithm & 20 & inception, named after, discoverer or inventor, computes solution to, best-case time complexity, average time complexity, best-case space complexity, worst-case time complexity, worst-case space complexity, average space complexity \\
animal & 20 & parent taxon, taxon rank, color, endemic to, main food source, gestation period, diel cycle \\
arena & 20 & sport, country, located in the administrative territorial entity, date of official opening, occupant, maximum capacity \\
chemical element & 30 & atomic number, time of discovery or invention, discoverer or inventor, element symbol, phase of matter \\
city & 59 & country, located in or next to body of water, population, official language, head of government, continent \\
company & 30 & country, headquarters location, inception, industry, chief executive officer, stock exchange \\
country & 30 & capital, currency, continent, shares border with, anthem, official language \\
currency & 14 & country, central bank/issuer, start time, unit symbol, applies to jurisdiction, ISO 4217 code \\
data set & 10 & inception, creator, copyright license, main subject, has use \\
disease & 20 & symptoms and signs, drug or therapy used for treatment, has cause, medical examination, disease transmission process, possible treatment \\
film & 25 & director, genre, country of origin, narrative location, production company, publication date \\
film award & 10 & inception, country, start time, conferred by, location \\
first-level administrative division & 20 & country, official language, population, shares border with, located in or next to body of water, capital \\
international organization & 20 & inception, chairperson, official language, headquarters location, operating area, position held by head of the organization, field of work \\
island & 15 & continent, country, located in the administrative territorial entity, located in or next to body of water, highest point \\
lake & 13 & inflows, outflows, drainage basin, basin country, country, located in the administrative territorial entity \\
large language model & 20 & inception, publication date, operator, developer, copyright license, context window \\
library & 15 & country, location, located in the administrative territorial entity, inception, director / manager \\
literary work & 23 & genre, language of work or name, author, publication date, publisher, country of origin \\
magazine & 20 & country of origin, inception, language of work or name, publisher, main subject, editor, publication interval \\
modern language & 20 & country, indigenous to, writing system, has grammatical case, ISO 639-1 code \\
motorsport racing track & 20 & sport, country, located in the administrative territorial entity, significant event, length, inception \\
mountain & 10 & continent, country, elevation above sea level, mountain range, topographic prominence, significant event \\
mountain range & 10 & continent, country, highest point, elevation above sea level, time period \\
museum & 20 & inception, director / manager, located in the administrative territorial entity, country, named after \\
musical group & 20 & genre, location of formation, country of origin, notable work, start of work period, has part(s) \\
musical work & 20 & genre, performer, lyricist, language of work or name, publication date, composer \\
national park & 14 & country, inception, located in the administrative territorial entity, highest point, elevation above sea level, heritage designation \\
painting & 20 & creator, inception, collection, genre, depicts \\
political party & 10 & chairperson, country, inception, political ideology, political alignment, founded by \\
professional sports league & 10 & sport, country, inception, number of participants, event interval \\
programming language & 20 & inception, designed by, developer, programming paradigm \\
protein & 20 & physically interacts with, active ingredient in, molecular function \\
recurring sporting event & 25 & event interval, inception, sport, organizer, number of participants, trophy awarded \\
research institute & 10 & located in the administrative territorial entity, country, inception, member of, field of work, named after, chairperson \\
river & 12 & origin of the watercourse, mouth of the watercourse, tributary, drainage basin, continent, country of origin \\
science award & 15 & country, conferred by, inception, named after, event interval, field of work \\
sculpture & 10 & inception, country, creator, made from material, location, depicts, time period, owned by \\
software & 27 & inception, developer, operating system, publication date, copyright license, platform, programmed in \\
sports award & 20 & conferred by, event interval, inception, sport \\
sports organization & 50 & country, inception, sport, head coach, home venue, league or competition \\
stadium & 20 & country, located in the administrative territorial entity, date of official opening, maximum capacity, occupant, sport, surface played on \\
stock exchange & 10 & country, headquarters location, stock market index, inception, chief executive officer \\
television series & 20 & genre, creator, original broadcaster, country of origin, original language of film or TV show, production company, distributed by, number of episodes, number of seasons \\
tourist attraction & 20 & inception, country, located in the administrative territorial entity, made from material, architectural style \\
type of chemical entity & 30 & chemical formula, has use, melting point, boiling point, has part(s) \\
type of musical instrument & 20 & made from material, has part(s), used by, discoverer or inventor, time of discovery or invention \\
type of sport & 20 & inception, authority, regulated by, country of origin, maximum number of players, minimum number of players \\
university & 30 & country, inception, located in the administrative territorial entity, chairperson, motto text, member of, founded by \\
video game & 25 & genre, platform, developer, publisher, publication date, game mode, distributed by \\
\bottomrule
\caption{Classes in the \texttt{instance of} dataset, their number of entities, and associated attributes.}
\label{table:instanceof-attributes}\\
\end{longtable}

\clearpage
\twocolumn

\begin{table*}[t]
\centering
\small
\setlength{\tabcolsep}{4pt}
\renewcommand{\arraystretch}{1.05}
\begin{tabular}{lll}
\toprule
\textbf{Class} & \textbf{\# Concept} & \textbf{Attributes} \\
\midrule
academic discipline & 17 & field of work, has subclasses, is the study of, part of \\
activity & 65 & has subclasses, location, participant \\
algorithm & 34 & has subclasses, has use \\
animal organ & 17 & body system, function, has part(s) \\
astronomical object & 26 & has part(s), has subclasses, parent astronomical body \\
award & 17 & conferred by, field of work, has subclasses \\
biological process & 20 & has effect, has part(s), has subclasses \\
bird & 17 & main food source, habitat, has subclasses, parent taxon \\
body of water & 15 & has part(s), has subclasses, largest example \\
chemical process & 20 & has part(s), has subclasses, product or material produced \\
color & 21 & has part(s), has subclasses, sRGB color hex triplet \\
creative work & 80 & creator role, form of creative work, has subclasses \\
disease & 29 & anatomical location, has subclasses, symptoms and signs \\
emotion & 36 & has subclasses, opposite of, positive valence \\
facility & 77 & has part(s), has subclasses, has use \\
fish & 18 & main food source, habitat, has subclasses, parent taxon \\
flower & 14 & blooming season, flower color, is pollinated by \\
food & 79 & has subclasses, physical state, vegetable or not \\
geographic location & 23 & has part(s), has subclasses \\
geographical feature & 24 & has part(s), has subclasses, largest example \\
industry & 13 & has subclasses, main offerings, occupation \\
insect & 14 & main food source, habitat, has subclasses, parent taxon \\
machine & 17 & has part(s), has subclasses, has use \\
mammal & 20 & main food source, habitat, has subclasses, parent taxon \\
material & 41 & has part(s), has subclasses, has use \\
measuring instrument & 13 & has use, measurement unit, measures \\
medical device & 14 & anatomical location, has part(s), has use \\
music & 19 & has subclasses, instrumentation \\
organization & 84 & field of work, has subclasses, occupation \\
problem & 11 & field of study, has cause, has effect, has subclasses \\
protein & 29 & has part(s), has subclasses, molecular function \\
science & 27 & has part(s), has subclasses, is the study of \\
scientific instrument & 11 & field of usage, has part(s), has use, observed object \\
software & 41 & has subclasses, has use, notable instance \\
sport & 35 & equipment, has subclasses, regulated by \\
theory & 10 & field of work, has subclasses, main subject \\
tool & 26 & has part(s), has use, made from material \\
tree & 14 & habitat, life expectancy, parent taxon \\
vehicle & 25 & has subclasses, operating medium, powered by \\
\bottomrule
\end{tabular}
\caption{Classes in the \texttt{subclass of} dataset, their number of subclasses, and associated attributes.}
\label{table:subclassof-attributes}
\end{table*}





\end{document}